\documentclass[12pt]{article}
\usepackage{amsmath}
\usepackage{graphicx}
\usepackage{enumerate}
\usepackage{natbib}
\usepackage{url} 

\newcommand{\blind}{1}

\usepackage[a4paper, margin=1in]{geometry} 

\renewcommand{\baselinestretch}{1.95}

\usepackage{latexsym, epsfig, amssymb, amsthm, mathrsfs, bbm, enumitem}
\usepackage{xcolor}
\usepackage[colorlinks,citecolor=blue,linkcolor=blue,urlcolor=blue]{hyperref}
\usepackage{multirow,multicol,booktabs}

\usepackage{algorithm, algpseudocode}
\usepackage{pgfmath} 

\newtheorem{lemma}{Lemma}
\newtheorem{proposition}{Proposition}
\newtheorem{theorem}{Theorem}
\newtheorem{definition}{Definition}
\newtheorem{remark}{Remark}
 
\newtheorem{assumption}{Assumption} 
\renewcommand{\theequation}{
	\arabic{equation}%
}

\newcommand{\ens}{\text{ens}}
\renewcommand{\part}{\text{part}}

\DeclareMathOperator*{\argmax}{arg\,max}
\newcommand{\vv}[1]{{\boldsymbol{#1}}}
\renewcommand{\epsilon}{\varepsilon}
\newcommand{\E}{\mathbb{E}}
\renewcommand{\P}{\mathbb{P}}
\newcommand{\R}{\mathbb{R}}
\newcommand{\N}{\mathbb{N}}

\renewcommand{\hat}[1]{\widehat{#1}}
\renewcommand{\tilde}[1]{\widetilde{#1}}
\allowdisplaybreaks[1]

\begin{document}

\def\spacingset#1{\renewcommand{\baselinestretch}%
{#1}\small\normalsize} \spacingset{0.}



	


\if1\blind
{
\title{\bf Exogenous Randomness Empowering Random Forests
  }
  \author{
    Tianxing Mei\thanks{
    Tianxing Mei is Postdoctoral Scholar, Data Sciences and Operations Department, Marshall School of Business, University of Southern California, Los Angeles, CA 90089 (E-mail: \textit{meit@marshall.usc.edu}).
    },  
    Yingying Fan\thanks{
    Yingying Fan is Centennial Chair in Business Administration and Professor, Data Sciences and Operations Department, Marshall School of Business, University of Southern California, Los Angeles, CA 90089 (E-mail: \textit{fanyingy@marshall.usc.edu}).
    %
    },  
    and
    Jinchi Lv\thanks{
    Jinchi Lv is Kenneth King Stonier Chair in Business Administration and Professor, Data Sciences and Operations Department, Marshall School of Business, University of Southern California, Los Angeles, CA 90089 (E-mail: \textit{jinchilv@marshall.usc.edu}). 
    This work was supported in part by NSF Grants EF-2125142, DMS-2310981, and DMS-2324490.
    }\\
      University of Southern California
    }
  \date{November 11, 2024}
  \maketitle
} \fi

\if0\blind
{
  \bigskip
  \bigskip
  \bigskip
  \begin{center}
    {\LARGE\bf Exogenous Randomness 
  Empowering Random Forests}
\end{center}
  \medskip
} \fi

\bigskip
\begin{abstract}
We offer theoretical and empirical insights into the impact of exogenous randomness on the effectiveness of random forests with tree-building rules independent of training data. We formally introduce the concept of exogenous randomness and identify two types of commonly existing randomness: Type I from feature subsampling, and Type II from tie-breaking in tree-building processes. We develop non-asymptotic expansions for the mean squared error (MSE) for both individual trees and forests and establish sufficient and necessary conditions for their consistency. In the special example of the linear regression model with independent features, our MSE expansions are more explicit, providing more understanding of the random forests' mechanisms. It also allows us to derive an upper bound on the MSE with explicit consistency rates for trees and forests. Guided by our theoretical findings, we conduct simulations to further explore how exogenous randomness enhances random forest performance. Our findings unveil that feature subsampling reduces both the bias and variance of random forests compared to individual trees, serving as an adaptive mechanism to balance bias and variance. Furthermore, our results reveal an intriguing phenomenon: the presence of noise features can act as a ``blessing" in enhancing the performance of random forests thanks to feature subsampling.
\end{abstract}

\noindent%

\spacingset{1.9} 

\section{Introduction} 
\label{new.Sec.intro}
Random forests, a type of ensemble estimator, have gained significant attention due to their appealing empirical performance across various applications over the past two decades. 
At a high level, random forests build individual trees (a type of partitioning estimator) using some recursive tree building rule (a type of domain partitioning rule), and then combine these trees into a single ensemble estimator.   
In Breiman’s random forests algorithm \citep{breiman2001random}, a key innovation is feature subsampling, where a random subset of features is chosen during each split, enhancing diversity among the trees. 
The size of the feature subset is chosen independently of tree construction and is controlled by a parameter 
$\gamma \in (0,1]$ with $\gamma=1$ representing no feature subsampling.   

Despite their success, random forests are often regarded as a ``black-box" model, as the reasons behind their effectiveness remain incompletely understood. 
A primary distinction between random forests, and individual trees and bagged trees is feature subsampling; yet its specific impact on performance needs deeper exploration. 
An important goal of our paper is to address this open question. For this purpose, we simplify other aspects of the random forests algorithm by considering deterministic tree-building rules that are independent of training data. We acknowledge that such simplifications exclude the training-data-dependent tree-building rules, including the sample CART \citep{breiman2001random}. 
However, it also allows us to peel out the effects of feature subsampling and ensemble. 
Additionally, the population CART can serve as a good proxy of the sample CART when the sample size is large and the tree depth is not too large. 

This paper considers nonparametric estimation via partitioning estimator ensembles within a general regression framework (see Section \ref{sec:partition-ensemble}).  
Our study will focus on the individual base learners being the partitioning estimators, including tree estimators as a special case, constructed from some training-data-independent partitioning rules. This includes popularly studied methods such as the population CART random forest \citep{klusowski2020analyzingcart,Chi2022}, 
the Mondrain forest \citep{Mourtada2020,cattaneo2024inferencemondrianrandomforests}, the centered forest \citep{ biau12a, klusowski21b}, and 
the mean/median forests \citep{Scornet2016}. 
To simplify the presentation, we use the term ``random forests" (RF) in a broad sense, equating it with the class of partitioning estimator ensemble methods throughout the paper.

In forming the RF estimators, two sources of randomness can arise: the \textit{endogenous} randomness related to training data, and the \textit{exogenous} randomness that is independent of training data. 
An example of the latter is the randomness introduced by feature subsampling in Breiman's RF.  In this paper, we consider ensemble estimators incorporating feature subsampling when forming their base estimators, and thus, there always exists \textit{exogenous} randomness in our RF. It is worth noting that for some partitioning rules, such as the population CART, there can exist \textit{additional} exogenous randomness beyond the one caused by feature subsampling, such as the randomness related to tie-breaking in choosing the next domain splitting location. We formalize these two types of exogenous randomness in Definition \ref{def:exogenous-randomness}.
This formal introduction is an important contribution of ours, as our study reveals that both types contribute to the success of RF, albeit in \textit{distinct} ways.  

In this paper, we establish nonasymptotic expansions for the mean squared error (MSE) of the individual tree estimator and the corresponding forest estimator. Our study confirms and demonstrates how exogenous randomness enhances the performance of RF. 
In particular, our theoretical results in Section \ref{new.Sec.modsetandpart} show that with exogenous randomness, i) ensemble ensures that RF has both smaller squared bias and variance than its base estimator; ii) the leading order contributions to both the squared bias and variance  are completely different for the forest estimator and the base estimator; 
and iii) forest estimators can be consistent under weaker conditions than their individual base estimators. We provide sufficient and necessary conditions for the consistency of the partitioning estimator and related forest estimator.
These results hold broadly for any partitioning estimator ensembles with training-data-independent partitioning rules and exogenous randomness. 

To gain additional insights into the population CART partitioning rule, we further consider a simplified setting of sparse linear regression with independent features. Two different feature distributions, binary and continuous uniform, are studied. Thanks to the simplified model setting, our general results on the MSE expansion take mathematically more specific forms, providing us additional insights into the effects of exogenous randomness and ensemble. We also establish an explicit (conservative) consistency rate for both the tree and the forest estimators, which clearly shows a bias-variance tradeoff as the tree depth and $\gamma$ vary. We further conduct theory-guided simulations based on our exact MSE expansions. 
Our theoretical and simulation results reveal an interesting yet surprising phenomenon of \textit{blessing} of dimensionality and noise features, which is made possible by feature subsampling. We list our major findings here: iv) our study challenges the conventional belief that when $\gamma=1$, RF is equivalent to a single tree; instead, we demonstrate that with Type II exogenous randomness, RF can achieve reduced variance without increasing bias compared to individual trees; v) when $\gamma < 1$, Type I exogenous randomness and the existence of numerous noise features act together to enable early-stage variance reduction during tree building, potentially coexisting with bias reduction for a while as tree depth grows; they may also help with bias reduction compared to the $\gamma = 1$ case, though this effect can be model-dependent; vi) for each fixed $\gamma$ value, as tree grows deeper, RF gradually shifts its focus from bias reduction to variance reduction (although they can coexist at all tree depths), often resulting in a $U$-shaped MSE curve. Although these additional insights are gained in linear models, we conjecture that they can carry over to broad model settings. We believe that some of the findings also shed light on the sample CART random forests, especially the role that feature subsampling plays in balancing the bias and variance of forest estimators. 

\subsection{Related literature}



Existing efforts to de-blackbox RF broadly fall into two categories.
The first category focuses on theoretically understanding RF’s consistency and predictive performance. 
Due to the complicated training-data-dependent nature of the sample CART, there only exist limited works  addressing the consistency of Breiman's original RF, including works on
the one-split stumps \citep{Buehlmann2002, Banerjee2007},
sparse additive regression models  \citep{Scornet2015,Klusowski2024}, 
binary feature models \citep{syrgkanis20a}, high-dimensional nonlinear models satisfying the sufficient impurity decrease (SID) condition \citep{Chi2022}, and the merged-staircase property (MSP) \citep{tan2024statcomp}. 
To make the problem more accessible,  many studies have instead focused on stylized RF variants, such as
purely random forests \citep{biau12a, klusowski21b}, 
median forests \citep{Scornet2016}, 
Mondrian forests \citep{Mourtada2020,cattaneo2024inferencemondrianrandomforests},
and honest forests  \citep{Wager2018,Athey2019}. These simplified models facilitate understanding consistency and parameter tuning, though they often fail to characterize theoretically the difference between trees and forests, particularly the underlying mechanisms of forests for enhanced performance. This underscores the importance for targeted study to address these questions.

The second category focuses on understanding the role of tuning parameters in the success of RFs, mostly from the empirical perspective. 
 \cite{MentchZhou2020} provided empirical evidence that in a low signal-to-noise ratio (SNR) setting, the feature subsampling rate serves as a regularizing function similar to a shrinkage penalty. 
 \cite{liu2024randomizationreducebiasvariance} used a simulation study to show that even in a high SNR setting, RF can achieve both bias and variance reduction, highlighting the role of feature subsampling in controlling RFs' sensitivity to noise features.  
\cite{Lin2006} and \cite{curth2024randomforestsworkunderstanding} interpreted RFs as adaptive nearest-neighbor estimators. 
Additionally, the impact of other parameters, such as the tree depth,  the forest size, the noise variance, and the maximum number of leaf nodes, have been explored empirically in
\cite{delgado14a,Probst2018,le2023survey,bernard2009influence,zhou2022random}. In contrast, our paper provides \textit{both} theoretical and empirical insights into how exogenous randomness helps enhance the performance of random forests over trees.


\section{Ensemble estimator with exogenous randomness}\label{sec:partition-ensemble}


Consider a training data set consisting of independent and identically distributed (i.i.d.) observations $\vv{Z} = \{(\boldsymbol{X}_i, Y_i)\}_{1\leq i\leq n}$ drawn from a generic population $(\boldsymbol{X}, Y)$, where random vector $\vv{X} = (X_1,\ldots,X_d)^\top$ takes values in a $d$-dimensional product topological space $\mathcal{X}^d = \mathcal{X} \times \cdots \times \mathcal{X}$ and response $Y$ takes values in $\mathcal{Y} = \R$ or subsets on $\N = \{0,1\ldots\}$.  The relationship between  $Y$ and  $\boldsymbol{X}$ is expressed through the nonlinear model 
\begin{equation} \label{eq:model0}
Y = \mu(\boldsymbol{X}) + \epsilon,
\end{equation}
where $\mu: \mathcal{X}^d \to \mathcal{Y}$ represents an unknown function to be estimated, and $\epsilon$ is the random model error satisfying $\mathbb{E}(\epsilon | \boldsymbol{X}) = 0$ and ${\rm Var}(\epsilon | \boldsymbol{X}) = \sigma^2(\boldsymbol{X})$ with $\sigma^2:\mathcal{X}^d \to \mathcal{Y}$ a variance profile function.
The model above covers the more realistic scenarios of heteroscedastic error variance. Our results in this paper are general and apply to both fixed and diverging dimensionality $d$; we will discuss the effects of dimensionality $d$ and the noise features in detail after we present our results and when the context is clearer.

Since random forests belong to the family of partitioning estimator ensemble methods, we introduce the concept of partitioning estimators and their ensemble in this section. Before proceeding, we introduce some necessary notation. 
Let $g$ be a generic integrable function on $\mathcal{X}^d$ and $A$ a generic measurable subset of $\mathcal{X}^d$. 
    We use $\E_{\vv X}(g)$ and $\P_{\vv X}(A)$ to stand for $\E(g(\vv X))$ and $\P(\vv X \in A)$, respectively, and  $I_A(\vv x) = I\{\vv x \in A\}$ to stand for the indicator function of set $A$. Denote by $\P_{\vv{X}}(A|B) = \P(\vv{X} \in A| \vv{X}\in B)$ for any $A\subset B$, $\E_{\vv{X}}(g|A) = \E(g(\vv{X})|\vv{X} \in A)$, and ${\rm Var}_{\vv{X}}(g|A) = {\rm Var}(g(\vv{X})|\vv{X}\in {A})$. 
    By $f(n)\lesssim g(n)$, we mean that there exists a constant $C$ independent of $n$ such that $|f(n)| \leq C g(n)$.
    If $f(n)$ depends on additional quantities $d,\mu,\sigma^2$ and so on, the implicit constant $C$ is assumed to be independent of those parameters unless specified otherwise.  
    Throughout, we use $a_n \asymp b_n$ to denote that there exist positive constants $c$ and $C$ such that $c b_n \leq a_n \leq C b_n$. 
    For any real number $x$, the smallest integer greater than or equal to $x$ is written as $\lceil x\rceil$.

\subsection{Partitioning estimator} \label{new.Sec.partest}


A (finite) partition $P$ of space $\mathcal{X}^d$ consists of a finite number of disjoint subsets with their union being the whole space; that is,
$
  P = \big\{
  P_j\subset \mathcal{X}^d: \bigcup_{j=1}^d P_j = \mathcal{X}^d,~P_j \cap P_j = \varnothing \text{ for } i\neq j\big\}.
$
We use $|P|$ to represent the cardinality of $P$. Denote by $\mathcal{F}_P = \sigma(P)$ the $\sigma$-algebra generated by partition $P$. For any square-integrable function $\mu$, 
we call the conditional expectation $\mu_P = \E_\vv{X}(\mu|\mathcal{F}_P)= \sum_{P_j\in P} \E_{\vv{X}}(\mu|P_j) I_{P_j}$ the projection of $\mu$ on $\mathcal{F}_P$.
It is well known that this projection is a population-version optimal estimator of $\mu$ when $\mu$ is given, with mean squared error (MSE) 
$\E_{\vv{X}}[(\mu - \mu_P)^2]
    = \min_{f\in\mathcal{F}_P}\E_{\vv{X}}[(\mu - f)^2]$.

 In practice, $\mu_P$ is inaccessible to us because the ground truth $\mu$ is generally unknown. With observed data $\vv{Z}$, we can form the partitioning estimator below based on partition $P$  by locally estimating the population means in $\mu_P$ using  sample observations 
\begin{equation}\label{defn:part0}
     \hat{\mu}_{\part}(\vv{x};\vv{Z},P) := \frac{1}{N_{\vv{x}}}\sum_{i=1}^n Y_i I\{\vv{X}_i \in P_{\vv{x}}\},
\end{equation}
where  $\vv{x}\in \mathcal{X}^d$ is a target test point, $P_\vv{x}$ stands for the unique region in $P$ that contains $\vv x$, and $N_{\vv{x}} = \sum_{i=1}^n I\{\vv{X}_i \in P_{\vv{x}}\}$ represents the sample size in $P_{\vv{x}}$. By convention, we define $\frac{0}{0} = 0$ to avoid ambiguity.

\subsection{Ensemble estimator with exogenous randomness} \label{new.Sec.partrule}

Various methods can be used to form partitions $P$.  
A \textit{partitioning rule} is a set of procedural steps, 
typically implemented algorithmically, 
that defines how a finite partition $P$ is formed for the feature space $\mathcal{X}^d$.
These rules determine partition boundaries based on some decision factors, which can carry inherent randomness. 
As a result, the corresponding partitions are typically random as well. 
This randomness usually arises from two main sources: the \textit{endogenous randomness} stemming from the training data $\vv{Z}$ itself, and 
the \textit{exogenous randomness} introduced by the user or inherently by the algorithm, independently of the training data. 
Since we focus on \textit{data-independent} partitioning rules, the resulting randomness in partitioning is determined only by exogenous factors. 

An arguably most well-known example is the  CART partitioning rule for building trees in RF. 
We provide details on the population CART partitioning rule, which will be the main focus of Section \ref{sec:CART}. 
First, we define the \textit{impurity decrement} when splitting a cell (hyperrectangle) $\mathbf{t} = \prod_{j=1}^d t_j \subset \mathcal{X}^d$ along direction $j$ at location $c\in t_j\subset \mathcal X$ into two daughter cells $\mathbf{t}_{j,c;L} = \mathbf{t} \cap \{X_j \leq c\}$ and $\mathbf{t}_{j,c;R} = \mathbf{t} \cap \{X_j >c\}$ as   \begin{equation}\label{eq:imdecrem}
    \begin{split}
        \Delta_{j,c}(\mathbf{t}) &:= \text{Var}_{\vv{X}}(\mu|\mathbf{t})
        - \text{Var}_{\vv{X}}(\mu|\mathbf{t}_{j,c;L}) \P_{\vv{X}}(\mathbf{t}_{j,c;L} | \mathbf{t}) 
        - \text{Var}_{\vv{X}}(\mu|\mathbf{t}_{j,c;R}) \P_{\vv{X}}(\mathbf{t}_{j,c;R}|\mathbf{t}).
        \end{split}
    \end{equation}
 For a parent cell $\mathbf{t}$, 
    a subset of features $\Xi \subset \{1,\ldots,d\}$ is randomly selected with $|\Xi|=\lceil \gamma d\rceil$ for a given $\gamma \in (0,1]$, 
    and
    the population CART finds the optimal pair $(j^*, c^*)$ as
    \begin{align}\label{eq:CART-obj}
        (j^*,c^*) = \argmax_{j\in \Xi, c\in t_j} \Delta_{j,c}(\mathbf{t}).
    \end{align}
  The parent cell $\mathbf{t}$ is then split into two daughter cells $\mathbf{t}_{j^*,c^*;L}$ and $\mathbf{t}_{j^*,c^*;R}$, each of which becomes new parent cell for the next round of splits. Ties are broken randomly in the optimization problem  \eqref{eq:CART-obj}. The entire process starts from the root cell $\mathcal{X}^d$,  proceeds recursively, and stops when a pre-given tree depth $l$ is attained. This results in a so-called (population) CART decision tree whose terminal cells (the cells at depth $l$) form a partition $P$ of space $\mathcal X^d$. A partitioning estimator in the form of \eqref{defn:part0} can be defined, which we will refer to as the CART tree estimator.  

The RF algorithm in  \cite{breiman2001random} forms tree estimators in a similar fashion to the process discussed above, with the difference that the sample CART  partitioning rule is used to grow trees, where the sample CART mimics the population CART by estimating the population moments in \eqref{eq:imdecrem} with sample moments formed by training data $\vv{Z}$. 
  
\begin{definition}[Exogenous randomness in population CART]\label{def:exogenous-randomness}
There are two types of exogenous randomness in  forming a population CART tree estimator: 
   \begin{enumerate}
    \item[i)](Type I). The randomness introduced by feature subsampling.
\item[ii)](Type II). The random tie-breaking in the optimization problem \eqref{eq:CART-obj}. 
\end{enumerate} 
\end{definition}
We remark that although the above definition is for the population CART partitioning rule, the general concept is broadly applicable to other partitioning rules.





To facilitate a rigorous analysis, we formalize the concept of a partitioning rule in mathematical terms. 
Let $\mathcal{P}$ denote the collection of all finite partitions of $\mathcal{X}^d$. 
A partitioning rule $\vv{P}:\mathcal{D} \to \mathcal{P}$
is a mapping from space 
$\mathcal{D}$ to collection 
$\mathcal{P}$ of finite partitions, where $\mathcal{D}$ summarizes all potential decision factors to determine the partition boundaries. 
Let $\Theta:\Omega \to \mathcal{D}$ be a random element mapping from certain measurable space $\Omega$ to the decision space  $\mathcal{D}$, and is assumed to be independent of the training data. 
Then $\vv{P}(\Theta)$ yields a random partition of feature space $\mathcal{X}^d$ driven by exogenous randomness. 
We note that,
in practice, partitioning rules are typically provided in the form of algorithmic procedures, meaning that both rule $\vv{P}(\cdot)$ itself and random element $\Theta$ of exogenous randomness may not have explicit mathematical expressions. For the purpose of theoretical analysis, 
we primarily focus on the aspect that partition 
$\vv{P}(\Theta)$ is the observable outcome of exogenous randomness, where notation
$\vv{P}(\Theta)$ is used to encapsulate the impact of exogenous randomness on the partition structure.
Such a simplification approach is commonly used in theoretical analysis of complex algorithmic procedures and helps gain robust insights across different algorithmic implementations.


Let $\vv{P}(\Theta_{1:B}) = \{\vv{P}(\Theta_1),\ldots,\vv{P}(\Theta_B)\}$ be a set of independent copies of random partition $\vv{P}(\Theta)$, 
where
$\Theta_{1:B} = \{\Theta_1,\ldots,\Theta_B\}$ is an i.i.d. sample from a population $\Theta$ of the exogenous random element. The \textit{ensemble estimator} averages the outputs of $B$ partitioning estimators formed on the \textit{same} input training sample $\vv{Z}$; that is,
\begin{equation}\label{defn:ens0}
    \hat{\mu}_{\ens}(\vv{x};\vv{Z},\vv{P}(\Theta_{1:B})) = \frac{1}{B} \sum_{b=1}^B \hat{\mu}_{\part}(\vv{x};\vv{Z}, \vv{P}(\Theta_b)).
\end{equation}

Let $\hat\mu(\vv x; \vv Z, \vv P(\Theta))$ be either $\hat{\mu}_{\part}$ or $\hat{\mu}_{\ens}$ as in \eqref{defn:part0} and \eqref{defn:ens0}. 
We use the following global mean squared error (GMSE) to evaluate the performance of $\hat\mu(\vv x; \vv Z, \vv P(\Theta))$ 
\begin{equation}\label{defn:MSE}
    \hbox{MSE}(\hat{\mu}) = \E_{\vv{X},\vv{Z},\Theta}\left[(\mu(\vv{X}) - \hat{\mu}(\vv{X};\vv{Z}, \vv P(\Theta))^2\right],
\end{equation}
where $\vv{X}$ is an independent test point that is identically distributed as the training covariate vector.

To facilitate our future presentation, for any partitions $P= \{P_j\}$ and $P'= \{P'_j\}$ with positive cell probabilities, we define \textit{the cross-partition covariance function} as
\begin{equation}\label{eq:modindpcov}
    \text{Cov}(P,P') = \sum_{P_i \in P}\sum_{P'_j \in P'} \frac{\P_{\vv{X}}(P_i\cap P'_j)^2}{\P_{\vv{X}}(P_i)\P_{\vv{X}}(P'_j)} = \sum_{P_i \in P}\sum_{P'_j \in P'}\P_{\vv{X}}(P_i|P'_j)\P_{\vv{X}}(P'_j|P_i).
\end{equation}
When $P = P'$, it reduces to the \textit{the single-partition variance function}
\begin{equation}\label{eq:modindpvar}
    \text{Var}(P) := \text{Cov}(P,P) =  \sum_{P_i\in P} 1 =|P|.
\end{equation}
It is clear that $0 \leq \text{Cov}(P,P')\leq \sqrt{\text{Var}(P)\text{Var}(P')}$ and the last equality holds if and only if $P$ and $P'$ are indistinguishable; see the paragraph before Assumption  \ref{assm:4} for the formal definition of distinguishability. 
The \textit{cross-partition correlation} is then defined as $\text{Corr}(P,P') = \text{Cov}(P,P')/\sqrt{\text{Var}(P)\text{Var}(P')}\in [0,1]$. It measures the correlation between two partitions and plays a key role in understanding the difference between a single partitioning estimator and the corresponding ensemble estimator.  

We conclude this section by re-iterating that our focus on data-independent partitioning rules is motivated by understanding the impact of exogenous randomness and ensemble on the accuracy of forest estimators.
This approach enables us to isolate the ensemble's influence without interference from other algorithmic configurations.


\section{Advantages of ensemble in CART random forests}\label{sec:CART}

In this section, we deviate from our general model in \eqref{eq:model0} and consider the following sparse linear model 
\begin{equation}\label{eq:binary}
y = \beta_1 X_1 + \beta_2 X_2 + \cdots + \beta_s X_s + \epsilon,
\end{equation}
where $\beta_1,\ldots,\beta_s \in \mathbb{R}\backslash \{0\}$ are nonzero regression coefficients, 
$X_1, \ldots, X_d$ are i.i.d. random variables, positive integer $s \leq d$ is the sparsity parameter, 
and the model error $\epsilon$ is independent of covariates with zero mean and a constant variance $\sigma_0^2$.  
Variables $X_1,\ldots,X_s$ are called \textit{informative variables} as they contribute to response $y$; 
when $d>s$, 
the remaining $d-s$ ones are \textit{non-informative} (or noise) variables since they are independent of $y$.  We set $\beta_j = 0$ for $j= s+1,\ldots,d$.
The more general nonlinear regression model  \eqref{eq:model0} will be considered in Section \ref{new.Sec.modsetandpart}. We focus on the simpler model here because it allows us to derive mathematically more explicit results and thus gain more insights.

In Section \ref{new.Sec.linear}, we will theoretically analyze the binary and continuous feature scenarios. We consider both types of feature distributions because the  CART partitioning rule can be significantly simplified in the binary case and is more complicated in the continuous case.  Our theoretical results show an intriguing fact that the MSE expansions of tree and forest estimators are strikingly similar for both binary and continuous features, an evidence supporting that the insights gained in this section may be broadly applicable. 
In Section \ref{new.Sec.simu}, we use theory-guided simulations to gain further insights.

\subsection{Theoretical analysis 
} \label{new.Sec.linear}

\subsubsection{Model with binary features} \label{new.Sec.example1}

In this section, we assume
that $X_1, \ldots, X_d$ are i.i.d. standard Bernoulli random variables $\mathcal{B}(1, 0.5)$, and that the tree depth $l < \lceil \gamma d\rceil$.
Since the covariates are binary, the splitting rule is uniquely determined once the direction for split is selected because any splitting position along that direction  results in the same outcome.
Hence, for any cell $\mathbf{t} = \prod_{j=1}^d t_j \subset \{0,1\}^d$, the impurity decrement in (\ref{eq:imdecrem}) is independent of split location and can be explicitly calculated as
\[
\Delta_j(\mathbf{t}) = 
\begin{cases}
\beta_j^2/4, & \text{if } j \in \{1, \ldots, s\} \text{ and } t_j = \{0, 1\};\\
0, & \text{if } j \in \{s+1, \ldots, d\} \text{ or } t_j = \{0\}, \{1\}.
\end{cases}
\]

Thus, in view of the description around \eqref{eq:imdecrem}, for each parent node $\mathbf{t}$ and a randomly subsampled feature set $\Xi \subset\{1,\ldots,p\}$ with size $\lceil \gamma d\rceil$, 
 \begin{itemize}
     \item If there are indices in $\{1,\ldots,s\}\cap \Xi$ that have not been split yet, we randomly choose one with the maximal $\beta_j^2$ as the optimal $j^*$;
     \item If all indices in $\{1,\ldots,s\}\cap \Xi$ have been split, then we randomly pick one from $\Xi$ that have not been split and set it as $j^*$.
 \end{itemize}


It is seen that each terminal cell $\mathbf{t}$ at depth $l$ can be uniquely determined by the states of whether each coordinate is split or not.  We thus define a stochastic process, named the \textit{Binary CART process}, to describe the states of coordinates in the tree-growing process.

\begin{definition}[Binary CART process]\label{def:binary-cart}
For a terminal cell $\mathbf{t}$ in a depth-$l$ tree, the binary CART process $I_k = (I_{k1}, \ldots, I_{kd})$  records the states of coordinates by tree depth $k$ for $k=0,1,\ldots,l$, where $I_{kj} = 1$ indicates that  $X_j$ has been split by depth $k$, and $I_{kj} = 0$ otherwise. 
\end{definition}
 
 The algorithm for generating the binary CART process is outlined in Section \ref{new.SecC.1}. For a point $\vv{x}_0 = (x_{01},\ldots,x_{0d})^\top$ and a tree with depth $l$, let  $\mathbf{t}_{\vv{x}_0}$ be the terminal cell containing $\vv x_0$. Then we have $\mathbf{t}_{\vv{x}_0} = \prod_{j: I_{lj} = 1} \{x_{0j}\} \times \prod_{j:I_{lj} = 0} \{0,1\}$. That is, the terminal cells of a depth $l$ tree can be completely characterized by their $I_l$'s.

 Using Definition \ref{def:exogenous-randomness}, it is seen that Type II exogenous randomness can occur in two scenarios: (a) when multiple informative features have identical $\beta_j^2$'s, or (b) when no informative feature is available, resulting in a random non-informative feature being chosen as $j^*$. 
Type I exogenous randomness is controlled by  parameter $\gamma$, and Type II exogenous randomness in this example is controlled by multiple parameters $s$, $\beta_j$'s, $d$, and $\gamma$. It will be made clear that both types can contribute to the success of random forests. 
 

Observe that given $I_k$, the distribution of $I_{k+1}$ only depends on the exogenous randomness discussed above. Further, given model \eqref{eq:binary}, the Binary CART process $I_l$ is \textit{Markovian} and parameterized by feature subsample rate $\gamma$. It is also important to note that $I_l$ and $I_l'$ from two separate depth $l$ trees are independent and identically distributed. 

Denote by $\hat{\mu}_{\text{tree}}$ and $\hat{\mu}_{\text{RF}}$ the tree and forest estimates formed with the  CART partitioning rule described above, respectively.
We present below the explicit expansion for  the MSE in \eqref{defn:MSE} of these two estimators using the binary CART process. 

\begin{theorem}\label{new.thm.binary}
Let $I_l$ and $I'_{l}$ be two independent Binary CART processes. 
The global MSE of the random forests estimator is given by
\begin{equation}\label{eq:binaryRF}
    \begin{split}
        \text{MSE}(\hat{\mu}_{\text{RF}}) & = \underbrace{\frac{B - 1}{4B} \E\left[\sum_{j=1}^s \beta_j^2 (1 - \max\{I_{lj},I'_{lj}\})\right]}_{\text{squared bias from tree ensemble}} 
         + \frac{1}{4B}\underbrace{ \E\left[\sum_{j=1}^s \beta_j^2(1- I_{lj})\right]}_{\text{single-tree squared bias}} \\
        & + \underbrace{\frac{B-1}{B}\E\left[\left(\sigma_0^2 + \frac{1}{4} \sum_{j=1}^s \beta_j^2 (1 - \max\{I_{lj},I'_{lj}\})\right)\frac{2^{\sum_{j=1}^d \min\{I_{lj},I'_{lj}\}}}{n}\right]}_{\text{cross-tree covariance}}\\
        & + \frac{1}{B}\underbrace{\E\left[\left(\sigma_0^2 + \frac{1}{4}\sum_{j=1}^s \beta_j^2 (1- I_{lj})\right)\frac{2^{\sum_{j=1}^d I_{lj}}}{n}\right]}_{\text{single-tree variance}} + \mathcal{R}_{RF},
    \end{split}
\end{equation}
where the remainder satisfies
$
\mathcal{R}_{RF} \lesssim \frac{2^l}{n(1 + (n-1) 2^{-l})^{1/2}} + \left(1 - 2^{-l} \right)^n.
$
In particular, when $B =1$, the global MSE of a single tree estimator is
\begin{equation}\label{eq:binarytree}
    \begin{split}
        \text{MSE}(\hat{\mu}_{\text{tree}})  = \underbrace{\frac{1}{4}\E\left[\sum_{j=1}^s \beta_j^2 (1- I_{lj})\right]}_{\text{single-tree squared bias}}  + \underbrace{\E\left[\left(\sigma_0^2 + \frac{1}{4}\sum_{j=1}^s \beta_j^2 (1- I_{lj})\right)\frac{2^{\sum_{j=1}^d I_{lj}}}{n}\right]}_{\text{single-tree variance}}+ \mathcal{R}_{\text{tree}},
    \end{split}
\end{equation}
in which the remainder is bounded by 
$
\mathcal{R}_{\text{tree}} \lesssim \frac{2^l}{n(1 + (n-1) 2^{-l})} + \left(1 - 2^{-l} \right)^n$. 
Furthermore, as $l = l_n \to \infty$ with $2^l/n \to 0$, we have the 
convergence rate result
\begin{equation}\label{eq:binrate}
    \begin{split}
        \max\{\text{MSE}(\hat{\mu}_{RF}), \text{MSE}(\hat{\mu}_{tree})\} \lesssim s \left(1 - \frac{3}{4}\gamma W_{\gamma,d}(s)\right)^{l+1}  + \frac{2^l}{n},
    \end{split}
\end{equation}
where the constant in $\lesssim$ depends only on $\beta_j^2$'s and $\sigma_0^2$, and function
\begin{equation}\label{eq:Wfunc}
  W_{\gamma,d}(x) := \left(1 - \frac{x}{d}\right)\cdots \left(1 - \frac{x}{d - \lceil \gamma d\rceil +1}\right) 
\end{equation}
depends only on feature subsample rate $\gamma \in (0,1]$ and dimensionality $d$.
\end{theorem}

The results in Theorem \ref{new.thm.binary} above are built upon a general Theorem \ref{new.thm2} to be presented in the next section, and its proof is provided in Section \ref{new.example1.derive}. Some interesting insights can be obtained by examining the expressions in \eqref{eq:binaryRF} and \eqref{eq:binarytree}.

\begin{remark}\label{rem:1} 
For simplicity, we assume that $B$ is large enough so that the terms involving $1/B$ are all negligible, and that  $\beta_j^2$'s all have similar magnitude $\beta_0^2>0$. Comparing (\ref{eq:binaryRF}) and (\ref{eq:binarytree}), it is seen that the leading contributions to MSE for trees and forests are completely different. For $\hat{\mu}_{tree}$, the variance and squared bias are driven by a single Markovian process $I_l$. For $\hat{\mu}_{RF}$,  owing to ensemble, MSE is driven by the interaction effects between two independent processes $I_l$ and $I_l'$, and the effects from individual tree estimates are negligible. 

Additionally, ensemble reduces both the squared bias and variance. For the squared bias terms, it is seen from both (\ref{eq:binaryRF}) and (\ref{eq:binarytree}) that they depend only on the $s$ informative variables via the weighted sums of their $\beta_j^2$'s. For a single tree, we have
    \[
    \underbrace{\sum_{j=1}^s \beta_j^2 (1 - I_{lj})}_{\text{squared bias by a single tree}} \asymp \beta_0^2\sum_{j=1}^s (1 - I_{lj}),
    \]
    in which the right-hand side (RHS) is proportional to the number of unsplit informative variables by a single binary CART process $I_l$. For forest, we have
    \begin{align*}
    \underbrace{\sum_{j=1}^s \beta_j^2 (1 - \max\{I_{lj},I'_{lj}\})}_{\text{squared bias by tree ensemble}} 
    &\asymp \beta_0^2\sum_{j=1}^s (1 - \max\{I_{lj},I'_{lj}\}),
    \end{align*}
    where the RHS is proportional to the number of unsplit features shared by  $I_l$ and $I'_l$. By comparing the two results above, the squared bias for $\hat{\mu}_{RF}$ is always smaller than that for $\hat{\mu}_{tree}$, showing that ensemble reduces the squared bias.
    
    For the variance terms, the order of the tree variance  is determined by 
    $2^{\sum_{j=1}^d I_{lj}}/n$, 
    which is driven by the number of splits in $I_l$. The order of the forest variance is determined by $2^{\sum_{j=1}^d \min\{I_{lj},I'_{lj}\}}/n$, which is driven by the number of shared splits between $I_l$ and $I_l'$. Noting that
     $   2^{\sum_{j=1}^d \min\{I_{lj},I'_{lj}\}}\leq 2^{\sum_{j=1}^d I_{lj}}$, 
    the forest estimator admits a smaller variance than the tree estimator. Intuitively, the exogenous randomness makes two individual trees likely split along distinct coordinators (that is, it is likely that $\min\{I_{lj},I'_{lj}\} < I_{lj}$), and hence reduces the variance of forests.

\end{remark}

   \begin{remark}\label{rem:2}
       Since both the squared bias and variance of $\hat{\mu}_{RF}$ are smaller than those of $\hat{\mu}_{tree}$, the MSE of an individual tree dominates that of the forest, and thus, the tree consistency leads to the forest consistency. However, the reverse is not necessarily true,  suggesting that $\hat{\mu}_{RF}$ may be consistent under weaker conditions. The convergence rate in (\ref{eq:binrate}) is conservative for both $\hat{\mu}_{tree}$ and $\hat{\mu}_{RF}$, more so for the latter. 
   \end{remark} 
    
    \begin{remark}\label{rem:3}
       Theorem \ref{new.thm.binary} provides a new perspective on how exogenous randomness helps reduce the pairwise correlation among trees. 
   To understand this, note that 
    using the definitions in (\ref{eq:modindpcov}) and (\ref{eq:modindpvar}), the expectation of the cross-tree covariance of two independent partitions $P$ and $P'$ generated by two 
    separate CART trees  can be written as
    $\E[\text{Cov}(P,P')] = \E\left[2^{\sum_{j=1}^d \min\{I_{lj},I'_{lj}\}}\right]$,
    and the expected variance of the single tree partition $P$ is $\E[\text{Var}(P)] = \E\left[2^{\sum_{j=1}^d I_{lj}}\right]$; the derivations can be found at the end of Section \ref{new.example1.derive}. When $l < \lceil \gamma d\rceil$, we have $\text{Var}(P) \equiv 2^{l}$. Thus, in this case, the expected \textit{cross-tree correlation function} becomes
    \begin{equation}\label{eq:crbinay}
        \E[\text{Corr}(P,P')] 
        = \E\left[2^{\sum_{j=1}^d \min\{I_{lj},I'_{lj}\} - l}\right].
    \end{equation}
    Recall that $\sum_{j=1}^d \min\{I_{lj},I'_{lj}\}$ is the number of shared splits by level $l$ and thus is always no larger than the total splits $l$. The correlation function takes values between $0$ and $1$, with a smaller number of shared splits corresponding to a smaller correlation. In this sense, the exogenous randomness helps increase the variability of splits across different trees and thus reduces $\sum_{j=1}^d \min\{I_{lj},I'_{lj}\}$, which consequently reduces the variance of $\hat{\mu}_{RF}$. 
    \end{remark} 
We note that although the above insights are drawn from this specific example, many carry over to general cases for ensemble estimators and individual partitioning estimators. The general results are summarized in Section \ref{new.Sec.modsetandpart}.

\subsubsection{Model with continuous features}\label{new.Sec.example2}

We now consider the case where $X_1, \cdots, X_d$ are i.i.d. uniform random variables $\mathcal{U}(0, 1)$. Similar to the last subsection, we start by describing the population CART partitioning rule applied to this specific setting. For a cell $\mathbf{t} = \prod_{j=1}^d t_j$ with $t_j = (a_j, b_j)\subset [0,1]$,
the conditional mean of $y$ on $\mathbf{t}$ is given by
$
\E[y|\vv{X} \in \vv{t}] = \sum_{j=1}^s \beta_j \frac{b_j + a_j}{2}
$. 
Then
the impurity decrement in (\ref{eq:imdecrem}) by splitting along $X_j$ is specified as 
\[
\Delta_j(\vv t) = 
\begin{cases}
\beta_j^2(b_j - a_j)^2/12, & \text{if } j \in \{1, \ldots, s\};\\
0, & \text{if } j \in \{s+1, \ldots, d\}.
\end{cases}
\]
We note that for any informative feature $j \in\{ 1,\ldots,s\}$, the optimal splitting position $c^*_j$ along the $j$th coordinate is the midpoint $\frac{b_j+a_j}{2}$ of $t_j$. However, 
in the case of non-informative variables (for $j\in\{s+1,\ldots,d\}$), splitting at any position within $t_j$ yields the same zero impurity decrement; we choose to split at the midpoint for simplicity. This simplifies the tie-breaking process by limiting randomness to split direction only. 

When building trees,
for each parent node $\mathbf{t}$ and a randomly sampled feature subset $\Xi \subset\{1,\ldots,p\}$ with size $\lceil \gamma d\rceil$, 
 if $\Xi \cap \{1,\ldots,s\} \neq \varnothing$, we choose $j^*\in \Xi$ as the one with the maximal impurity decrement (ties are broken randomly); and
 if $\Xi\cap \{1,\ldots,s\} = \varnothing$,  we randomly pick one from $\Xi$ as $j^*$. We note that, unlike the binary case, an index $j\in \{1,\cdots, s\}$ in the uniform case can be split arbitrarily many times, with the impurity decrement always being positive. This difference also makes the continuous feature case more difficult to analyze.

Similar to the binary case, we now define the \textit{Uniform CART process} below. 

\begin{definition}[Uniform CART process]\label{def:unif-CART}
For a terminal cell $\mathbf{t}$ in a depth $l$ tree, the uniform CART process is a vector process $J_k = (J_{k1}, \ldots, J_{kd})$ with $J_{kj}$ being the number of splits along coordinate $X_j$ by tree depth $k$, where $j\in \{1,\ldots,d\}$ and $k\in \{1,\cdots, l\}$. 
\end{definition}

The detailed algorithm for generating $J_k$ is provided in Section \ref{new.SecC.1}. Similar to the binary case, both types of exogenous randomness in Definition \ref{def:exogenous-randomness} can exist here.  The uniform CART process is also \textit{Markovian} and parameterized by  $\gamma$ given model \eqref{eq:binary}, since the conditional distribution of $J_{k+1}$ given $J_k$  is only determined by the exogenous randomness.

Each terminal cell $\mathbf{t}$ can be uniquely determined by its uniform CART process $J_l$.  To understand this, note that  
given $x_0 \in (0,1)$, if interval $(0,1)$ is split equally into $m$ parts, the unique interval containing $x_0$ is $t(x_0,m)  = \left(\frac{K(x_0,m) - 1}{2^{m}}, \frac{K(x_0,m)}{2^{m}}\right]$ with $K(x_0,m) = \lceil x_0 2^{m}\rceil$.
Thus, for a given $\vv{x}_0 = (x_{01},\ldots,x_{0d})^\top$, the terminal cell $\vv{t}_{\vv{x}_0}$ can be expressed as 
\begin{equation}\label{eq:terminalinterval-all}
   \vv{t}(\vv{x}_0,J_l) = \prod_{j=1}^d t(x_{0j},J_{lj}), 
\end{equation}
in which $J_l$ is independent of $\vv{x}_0$.


Same as in the binary case, $J_l$ and $J_l'$ from two separate trees grown by the uniform CART partitioning rule are i.i.d. random vectors.  We are now ready to present the MSE results for individual tree estimator and forest estimator in this setting below. 

\begin{theorem}\label{new.thm.unif}
Let $J_l$ and $J'_{l}$ be two independent Uniform CART processes. 
The global MSE of the related random forests estimator $\hat{\mu}_{RF}$ is 
\begin{equation}\label{eq:unifRF}
    \begin{split}
        \text{MSE}(\hat{\mu}_{RF}) & = \underbrace{\frac{B-1}{12 B} \sum_{i=1}^s \beta_i^2 \E\left[2^{- 2 \max\{J_{li},J'_{li}\}}\right]}_{\text{squared bias by tree ensemble}} + \frac{1}{ B}\underbrace{ \sum_{i=1}^s\frac{a^2}{12}
        \beta_i^2 \E\left[2^{-2 J_{li}}\right]}_{\text{squared bias by a single tree}}\\
        & + \underbrace{\frac{B - 1}{B}\E\left[\left(\sigma_0^2 + \frac{a^2}{12} \sum_{i=1}^s \beta_i^2 2^{-2\max\{J_{li},J'_{li}\}}\right) \frac{2^{\sum_{i=1}^d \min\{J_{li},J'_{li}\}}}{n}\right]}_{\text{cross-tree covariance}} \\
        & + \frac{1}{B}\underbrace{
        \E\left[\left(\sigma_0^2 + \frac{a^2}{12}\sum_{i=1}^s \beta_i^2 2^{-2J_{li}}\right)\frac{2^l}{n}\right]
        }_{\text{single-tree variance}} + \mathcal{R}_{RF}.
    \end{split}
\end{equation}
In particular, when $B=1$, the global MSE for a single tree estimator is 
\begin{equation}\label{eq:uniftree}
    \begin{split}
        \text{MSE}(\hat{\mu}_{\text{tree}})  =\underbrace{ \sum_{i=1}^s \frac{a^2}{12}\beta_i^2\E\left[2^{-2 J_{li}}\right]}_{\text{squared bias by a single tree}} + \underbrace{
        \E\left[\left(\sigma_0^2 + \frac{a^2}{12}\sum_{i=1}^s \beta_i^2 2^{-2J_{li}}\right)\frac{2^l}{n}\right]
        }_{\text{single-tree variance}} + \mathcal{R}_{\text{tree}}.
    \end{split}
\end{equation}
Here, the remainders $\mathcal{R}_{RF}$ and $\mathcal{R}_{\text{tree}}$ are the same as in Theorem \ref{new.thm.binary}. 
Additionally, as $l = l_n \to \infty$ with $2^l/n \to 0$, $\max\{\text{MSE}(\hat{\mu}_{RF}),\text{MSE}(\hat{\mu}_{tree})\}$ has the same upper bound as in \eqref{eq:binrate}.  
\end{theorem}

The proof of Theorem \ref{new.thm.unif} above is also based on Theorem \ref{new.thm2} and is detailed in Section  \ref{new.example2.derive}. It is seen that despite the distinct feature distributions in this example, the MSE expansions exhibit surprisingly similar forms, and all insights presented in the last section remain true. We provide three remarks emphasizing the differences from the binary case.    

\begin{remark} \label{rem:4}
We make the same simplification assumption that $\beta_j^2$'s all have similar magnitude $\beta_0^2$, and that $B$ is large enough so that all terms involving $1/B$ are small enough. Theorem \ref{new.thm.unif} also shows that ensemble reduces both squared bias and variance, and that the leading contributions to MSE for trees and forests are completely different. 

   For the squared bias term, both $\hat{\mu}_{tree}$ and $\hat{\mu}_{RF}$ depend only on the informative features. However, the dependence mechanism here is different from the binary case. To make this clear, we define the \textit{diagonal signal length} along the direction of informative features for a cell $\mathbf{t} = \prod_{i=1}^d t_i$ as $  |\mathbf{t}|_{2,s} = \left(\sum_{i=1}^s |t_i|^2\right)^{1/2}$. 
   For a target point $\vv{x}_0$, recall the terminal cell $\mathbf{t}({\vv{x}_0},J_l)$ in \eqref{eq:terminalinterval-all}. Then we have 
   \begin{align*}
    \underbrace{\sum_{i=1}^s \beta_i^2 2^{-2 J_{li}}}_{\text{squared bias for a single tree}} &\asymp \beta_0^2\sum_{i=1}^s 2^{-2 J_{li}} = \beta_0^2|\mathbf{t}(\vv{x}_0,J_l)|_{2,s}^2,\\
    \underbrace{\sum_{i=1}^s \beta_i^2 2^{-2 \max\{J_{li}, J'_{li}\}}}_{\text{squared bias for forest}} & \asymp \beta_0^2\sum_{i=1}^s 2^{-2 \max\{J_{li}, J'_{li}\}}  = \beta_0^2 |\mathbf{t}(\vv{x}_0,J_l)\cap \mathbf{t}(\vv{x}_0,J'_l)|_{2,s}^2. 
   \end{align*}
    Since the squared bias of $\hat{\mu}_{RF}$ is associated with the intersection of two terminal nodes,  it is naturally smaller than that of $\hat{\mu}_{tree}$. The comparison and interpretation of variance terms are similar to the binary case and omitted here. 
\end{remark}    

\begin{remark}\label{rem:5}
Comparing the above upper bound to \eqref{eq:binrate}, we see a surprising similarity between the results despite the different feature distributions. These upper bounds have an explicit dependence on $\gamma$. However, since these bounds can be conservative and the MSE expansions in \eqref{eq:binaryRF}, \eqref{eq:binarytree}, \eqref{eq:unifRF}, and \eqref{eq:uniftree} are more accurate in characterizing the effect of $\gamma$, we will use simulation studies to simulate the Markovian processes involved in these upper bounds to gain insights on how $\gamma$ affects the MSE.  

\end{remark}
\begin{remark}\label{rem:6}
    Using (\ref{eq:modindpcov}) and (\ref{eq:modindpvar}), the expectation of the cross-tree covariance of partitions $P$ and $P'$ generated by two independent uniform CART processes $J_{lj}$ and $J'_{lj}$ can be calculated as
    $\E[\text{Cov}(P,P')] = \E\left[2^{\sum_{j=1}^d \min\{J_{li},J'_{li}\}}\right]$,
    and the expectation of one single tree variance can be calculated as $\E[\text{Var}(P)] = \E\left[2^{\sum_{i=1}^d J_{li}}\right] \equiv 2^{l}$; see the end of Section \ref{new.example2.derive} for detailed derivations. Thus, the expected {\em cross-tree correlation function} is
    \begin{equation}\label{eq:crunif}
        \E[\text{Corr}(P,P')] = 
        \E\left[2^{\sum_{i=1}^d \min\{J_{li},J'_{li}\} - l}\right].
    \end{equation}
    Similar to the binary case, the correlation function is strictly less than 1 unless $J_l$ and $J_l'$ share all splits (i.e., $\sum_{i=1}^d \min\{J_{li},J'_{li}\}= l$). Thus, the exogenous randomness helps increase the variability of $J_l$ and $J_l'$ and consequently reduces the forest estimator variance.
\end{remark}

\subsection{Additional insights via theory-guided simulations} \label{new.Sec.simu}


This section provides some additional insights into the success of RF over individual trees via simulation studies. Our simulation examples are designed and guided by the theoretical results established in the last subsection. 
The following three key insights about the success of RF are discussed, accompanied by simulations: 1) ensemble can reduce both squared bias and variance; 2) RF can outperform single trees even when $\gamma = 1$; and  3) exogenous randomness improves the performance of RF over tree both when  $\gamma = 1$ and $\gamma<1$, but the working mechanisms are different in these two cases.

We generate data from model (\ref{eq:binary}) and set dimensionality $d = 100$, $s = 5$, sample size $n = 1000$, the number of trees  $B = 100$, and noise variance $\sigma_0^2 = 1.69$. The true regression coefficient vector $\beta = (\beta_1, \beta_2, \ldots, \beta_5)$ has the following two different configurations: 

(I) \textit{Equal configuration:} $\beta_1 = \beta_2 = \beta_3 = \beta_4 = \beta_5 = 0.5$.

(II) \textit{Unequal configuration:} $\beta_1 = 2$, $\beta_2 = 1.8$, $\beta_3 = 1.6$, $\beta_4 = 1.4$, $\beta_5 = 1.2$.

\noindent We vary feature subsampling rate $\gamma$ from $0.1$ to $1$, and the tree depth from $l = 1$ to $9$. 

Motivated by our theoretical results, we record the values of the following performance measures across $1000$ Monte Carlo simulations for comparing RF to single trees:

   (a) \textit{Squared bias:} Squared bias terms in (\ref{eq:binaryRF}) for the binary case, and those in (\ref{eq:unifRF}) for the continuous case.
    
    (b) \textit{Number of unsplit signals in the binary case:}  $s - \sum_{j=1}^s I_{lj}$ for a single tree, and  $s -\sum_{j=1}^s \max\{I_{lj},I'_{lj}\}$ for a random forest, as discussed in Remark \ref{rem:1}.
    
    \textit{Squared diagonal signal length in the continuous case:}  $\sum_{i=1}^s 2^{-2J_{li}}$ for a single tree, and $\sum_{i=1}^s 2^{-2\max\{J_{li},J'_{li}\}}$ for a random forest, as discussed in Remark \ref{rem:4}.
    
    (c) \textit{Variance:} Single-tree variance and cross-tree covariance in (\ref{eq:binaryRF}) for the binary case, and those in (\ref{eq:unifRF}) for the continuous case.
    
    (d) \textit{Cross-tree correlation:} Expected cross-tree correlation (\ref{eq:crbinay}) for the binary case, and (\ref{eq:crunif}) for the continuous case. The cross-tree correlation for tree is defined as $1$. 
    
    (e) \textit{Number of shared splits:}  For a single tree,  $\sum_{j=1}^d I_{lj}$ for the binary case and $\sum_{i=1}^d J_{li}$ for the continuous case;  
    for a forest,  $\sum_{j=1}^d \min\{I_{lj},I'_{lj}\}$ for the binary case and $\sum_{i=1}^d \min\{J_{li},J'_{li}\}$ for the continuous case.

    (f) \textit{Global MSE bound:} The leading terms of the global MSE bounds in (\ref{eq:binaryRF}) and (\ref{eq:unifRF}), respectively, for the binary and continuous cases, with the remainder terms ignored.

    
    
    
    
    


\subsubsection{Random forests can reduce both squared bias and variance}\label{new.Sec.biasvarreduction}

\begin{figure}[htp]
    \centering
    \includegraphics[width = 0.75\textwidth]{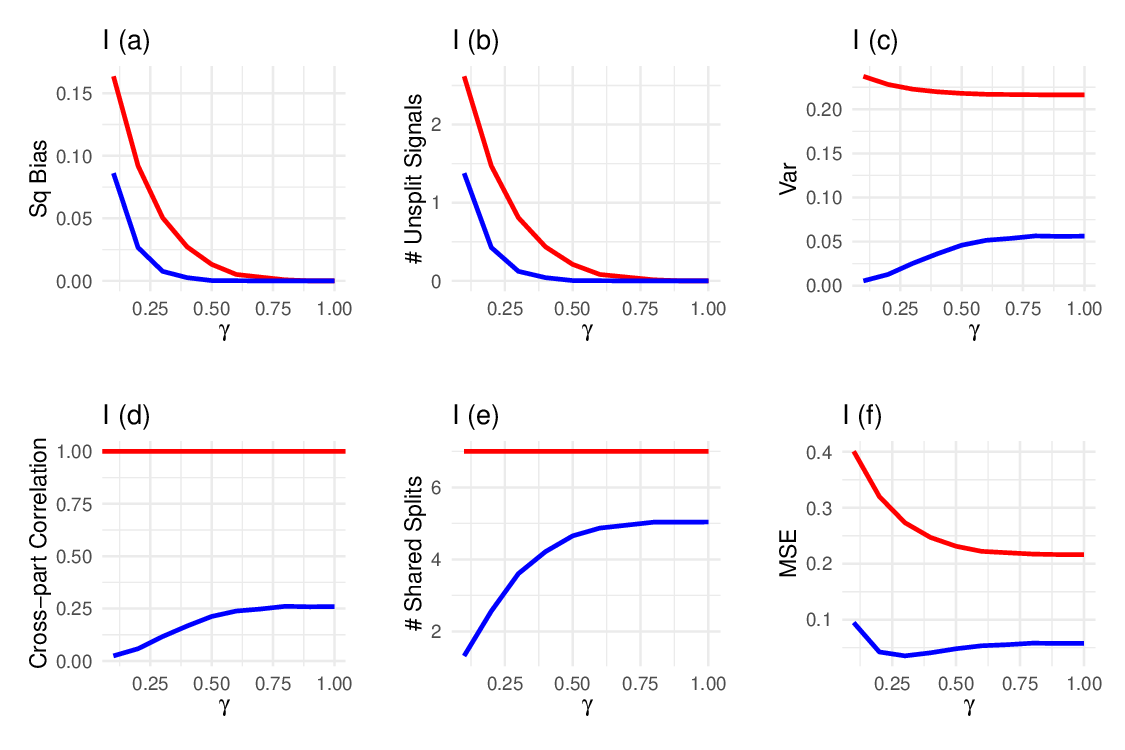}
    \caption{Performance measures (a)--(f) for  tree and  forests in the binary  case under configuration (I) as $\gamma$ varies. Tree depth is fixed at $l = 7$. Red: tree; blue: forest.}
    \label{fig:simu1}
\end{figure}

\begin{figure}[h]
    \centering
    \includegraphics[width = 0.75\textwidth]{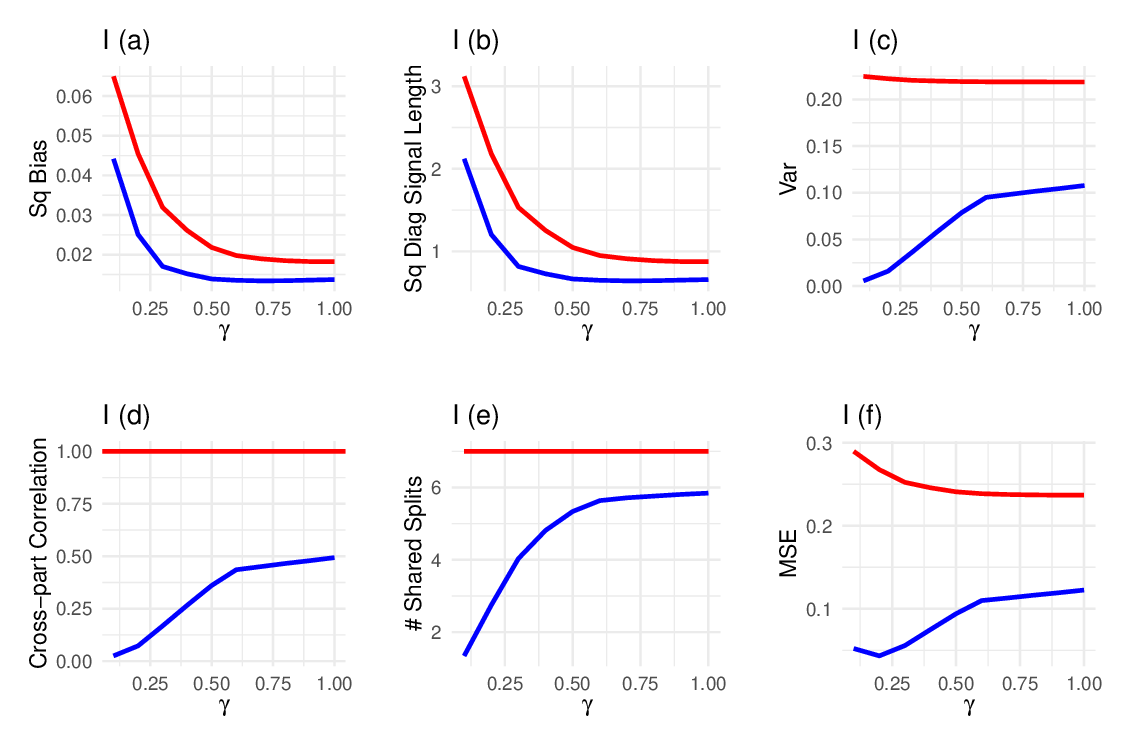}
    \caption{Performance measures (a)--(f) for tree and forest in the continuous case  under configuration (I) as $\gamma$ varies. Tree depth is fixed at $l = 7$. Red: tree; blue: forest.}
    \label{fig:simu4}
\end{figure}

This section aims at verifying Remarks \ref{rem:1} and \ref{rem:4} via simulation using Equal Coefficient Configuration (I).
We present the results for performance measures (a)--(f) in Figures \ref{fig:simu1} and \ref{fig:simu4} for the binary and the continuous case, respectively, with respect to varying $\gamma$.  

From Figures \ref{fig:simu1}(a) and (c), we observe that the squared bias and variance curves for RFs are consistently lower than those for tree,  leading to a substantial improvement in MSE as shown in Figure \ref{fig:simu1}(f). 
The reduction in squared bias is because the forest splits more signal features, as shown in Figure \ref{fig:simu1}(b). 
The reduction in variance is due to fewer shared splits between $I_l$ and $I_l'$, as illustrated in Figure \ref{fig:simu1}(e). These results are consistent with our theoretical analysis in Remark \ref{rem:1}. 
Additionally, the correlation between trees is low, as shown in Figure \ref{fig:simu1}(d), also explaining the reduction in variance.

For the continuous features, the trends observed in Figure \ref{fig:simu4} for performance measures (a)--(f) exhibit similar patterns to those in the binary case. These results also support our conclusion in Remark \ref{rem:4}, demonstrating that RF consistently achieves lower squared bias and variance compared to tree as $\gamma$ varies. They also confirm the improved effectiveness of RF across different types of feature distribution.


Additionally, as seen in Figures \ref{fig:simu1}(f) and \ref{fig:simu4}(f), the MSE curves for RFs exhibit $U$-shaped patterns, with the optimal MSE differing in these two settings. This observation indicates that the optimal $\gamma$ can be sensitive to the model structure, calling for further investigation.

\subsubsection{Random forests can outperform decision trees even with $\gamma=1$}
\label{new.Sec.gamma1}

\begin{figure}[h]
    \centering
    \includegraphics[width = 0.75\textwidth]{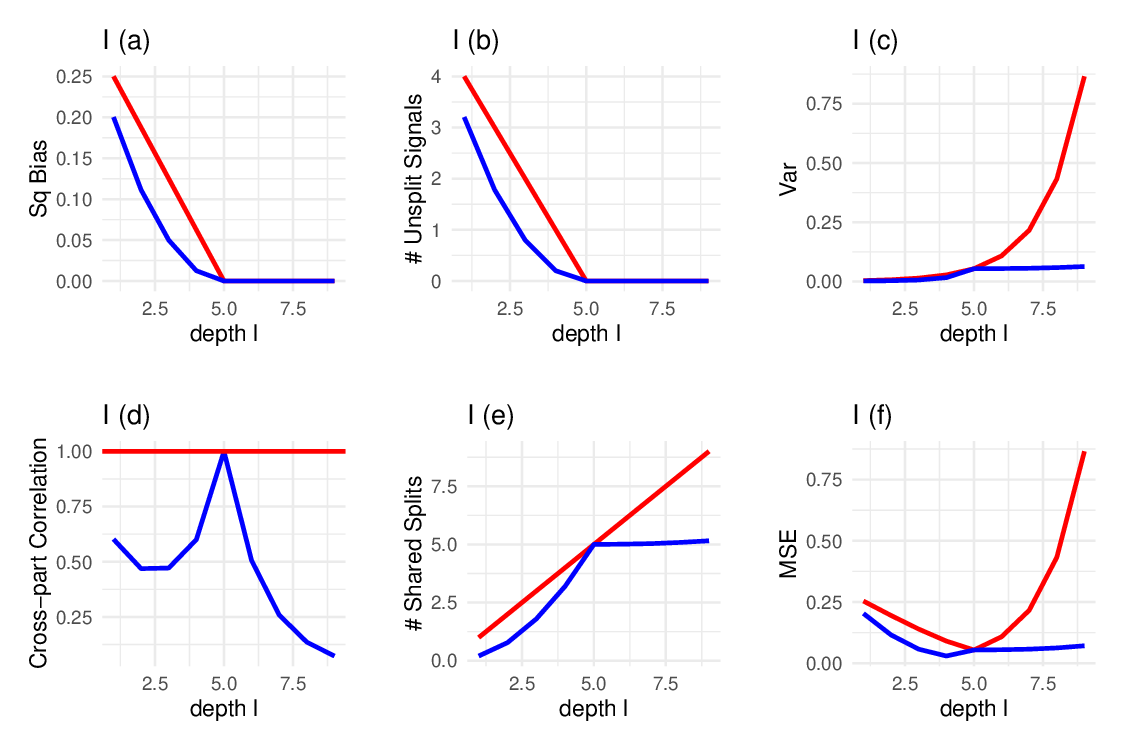}
    \caption{Performance measures (a)--(f) for tree and  forest in the binary case under configuration (I) as $l$ varies. We fix $\gamma = 1$.
    Red: tree; blue: forest.}
    \label{fig:simu2}
\end{figure}
\begin{figure}[h]
    \centering
    \includegraphics[width = 0.75\textwidth]{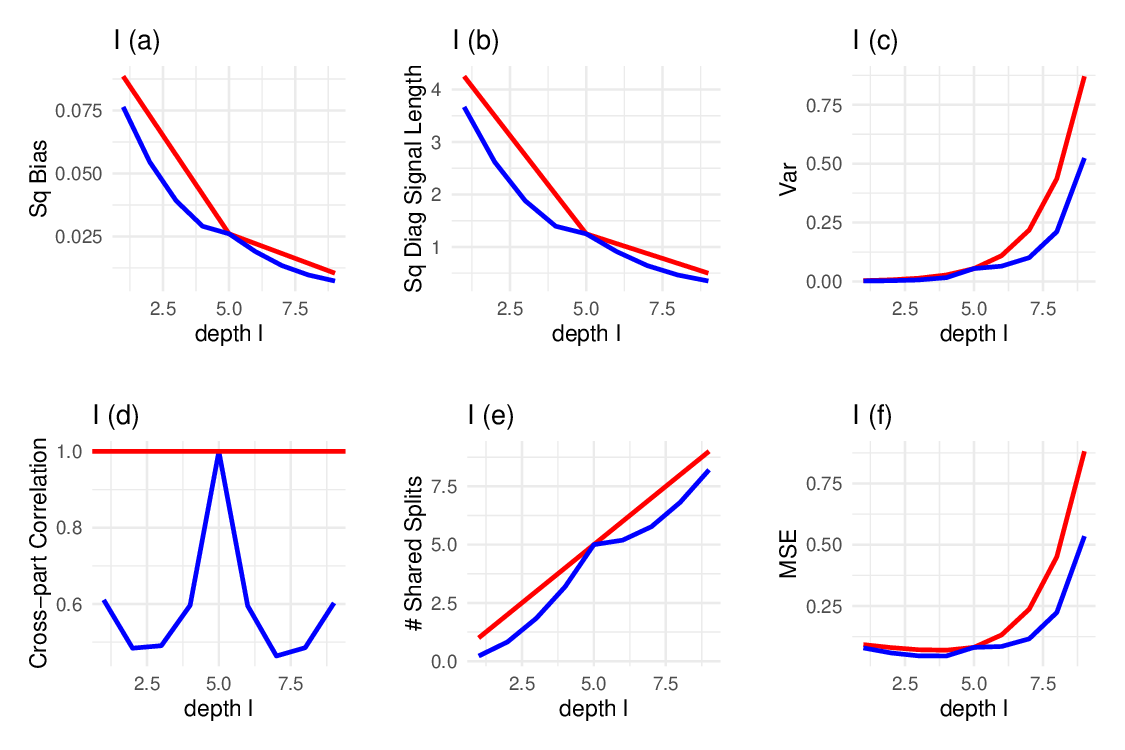}
    \caption{Performance measures (a)--(f) for  tree and forest in the continuous case under configuration (I) as $l$ varies. We fix $\gamma = 1$. Red: tree; blue: forest.}
    \label{fig:simu5}
\end{figure}

A surprising observation from Figures \ref{fig:simu1}(f) and \ref{fig:simu4}(f) is that, the MSE for forest is strictly less than that for tree even when $\gamma = 1$, contrary to the conventional wisdom that tree and forest are the same when $\gamma=1$. This improved performance is attributed to Type II exogenous randomness, which reduces correlations among trees. Note that there is no Type I exogenous randomness in this situation. To further illustrate these points, we compare the performance of forests and trees when
 $\gamma = 1$ for both binary and continuous feature distributions. How performance measures in (a)--(f) change as tree depth varies is displayed in Figures \ref{fig:simu2} and \ref{fig:simu5}, respectively, for the binary and continuous cases.

For the binary case,  Figure \ref{fig:simu2} reveals clearly that forests and tree differ across various tree depths. In particular, Figure \ref{fig:simu2}(d) shows an intriguing pattern in the cross-tree correlation -- the correlation reaches $1$ at 
$l = s= 5$, suggesting that only at this depth do the tree and forest behave identically. In contrast, for all other depths, the cross-tree correlation remains strictly below $1$, indicating a fundamental difference between the tree and forest. Furthermore, the correlation exhibits distinct trends on the two sides of 
$l = 5$, with a more rapid decrease on the RHS when $l > 5$. 

The cross-tree correlation pattern in Figure \ref{fig:simu2}(d)  can be explained theoretically using our findings in Section \ref{new.Sec.example1}.
Recall the CART partitioning rule and the fact that each binary feature is split at most once along each tree branch connecting a terminal cell and the root cell. When $\gamma = 1$, the binary CART process $I_l$ (Definition \ref{def:binary-cart}) has a specific property: it first splits the $s$ informative variables, selecting an optimizer randomly for each split in case of ties; from tree depth $s+1$ onward, it moves to split randomly-chosen non-informative features among available ones. This naturally explains different causes of Type II exogenous randomness before and after tree depth $s$.
In detail,

(i) When depth $l < 5$, 
        Type II exogenous randomness comes only from the similarity among informative features: since the $\beta_j$'s are identical, the random tie-breaking introduces randomness and reduces the correlation between trees. Such randomness allows the forest to split more signals than one single tree at any given depth $l<5$, thereby lowering the squared bias compared to a single tree (Figures \ref{fig:simu2}(a) and (b)). 

(ii) When $l= 5$, all the informative features have been split, and non-informative features have not been split yet. Thus, there is no randomness at this tree depth, leading to identical behavior between tree and forest in Figure \ref{fig:simu2}. 
        Since the squared bias is only determined by the number of unsplit signals, it reaches zero for both tree and forest (Figures \ref{fig:simu2}(a) and (b)).

(iii) When $l > 5$,  the non-informative features begin to contribute. These features have no influence on the bias reduction but can cause Type II exogenous randomness. When $d-s$ is large, such Type II exogenous randomness is large, which makes the number of shared splits $\sum_{j=1}^d \min\{I_{lj},I'_{lj}\}$ very close to $s=5$ (see Figure \ref{fig:simu2}(e)),  suggesting that it is generally hard for $I_l$ and $I_l'$ to share any split on non-informative variables. This leads to a sharp decline in cross-tree correlation $ 2 ^{\sum_{j=1}^d \min\{I_{lj},I'_{lj}\} - l} \approx 2^{- (l-5)}$ for $l > 5$ as shown in Figure \ref{fig:simu2}(d) and the significantly slow growth in variance and MSE in Figures \ref{fig:simu2}(c) and (f).

For the continuous case, we observe similar behavior in Figure \ref{fig:simu5}: RF consistently outperforms tree across various depths when $\gamma = 1$, although the specific patterns differ significantly from the binary case. Below we focus mainly on discussing the factors that contribute to the differences.
Recall that, unlike the binary case, informative features here can be split infinitely many times along a tree branch connecting a terminal cell with a root cell.
As a result, when $\gamma = 1$, the uniform CART process splits \textit{only} on the informative features during the tree-building process.
Thus, Type II exogenous randomness 
arises solely from the tie-breaking when searching for the informative feature with the largest impurity decrement. As shown in Figure \ref{fig:simu5}(d), i)  when $l<5$, the random tie-breaking among informative features ensures that the  cross-tree correlation is strictly less than 1;
ii) at $l =5$, each informative feature has been split exactly once and thus there is no Type II exogenous randomness, resulting in identical tree and forest performance; and 
iii) for $l>5$ but $l<10$, each of the $s=5$ informative features is split exactly one more time,  leading to a second round of reduction in correlation.

For the Equal Configuration (I), we observe the interesting phenomenon of quasi-periodic fluctuations in Figures \ref{fig:simu5}(d) and (e). Each time when trees grow from depth $ks + 1$ to depth $(k+1) s$ for $k=0, 1, ...$, each informative feature is split exactly once. As $k$ increases, the impurity decrement in each quasi-period gradually diminishes, resulting in slower and slower decreases each period in squared bias in Figure \ref{fig:simu5}(a). However, the impact of variance becomes increasingly significant as $l$ increases, because the leading term 
 $2^l /n$ in variance term keeps growing as $l$ increase. This explains the substantial rise in both variance and MSE when $l$ is large, as shown in Figures \ref{fig:simu5}(c) and (f).

We now move on to Unequal Configuration (II). 
In the binary setting, Type II exogenous randomness only happens when tree depth $l>s=5$ (i.e., non-informative features enter tree building), and in the continuous case, there is no Type II exogenous randomness. 
Thus, as shown in Figures \ref{fig:simu7} and \ref{fig:simu8} in Section \ref{new.SecC.2}, the tree and forest estimators have identical performance in the continuous case, and forest only outperforms tree when tree depth $l>s=5$ in the binary case. This once again confirms the contribution of Type II exogenous randomness to the success of random forests over trees.     


Our study also reveals an intriguing phenomenon that the existence of numerous non-informative features can be a \textit{blessing} to random forests' success, because of their contribution to reduced cross-tree correlations. We acknowledge that the blessing of high dimensionality is, to some extent, because of the population CART partitioning rule. For the sample CART partitioning rule, the effect of noise variables and high dimensionality can be more complicated because of the finite-sample estimation error involved. 

\subsubsection{Feature subsampling improves the performance of random forests}
\label{new.Sec.subsample}

\begin{figure}[h]
    \centering
    \includegraphics[width = 0.8\textwidth]{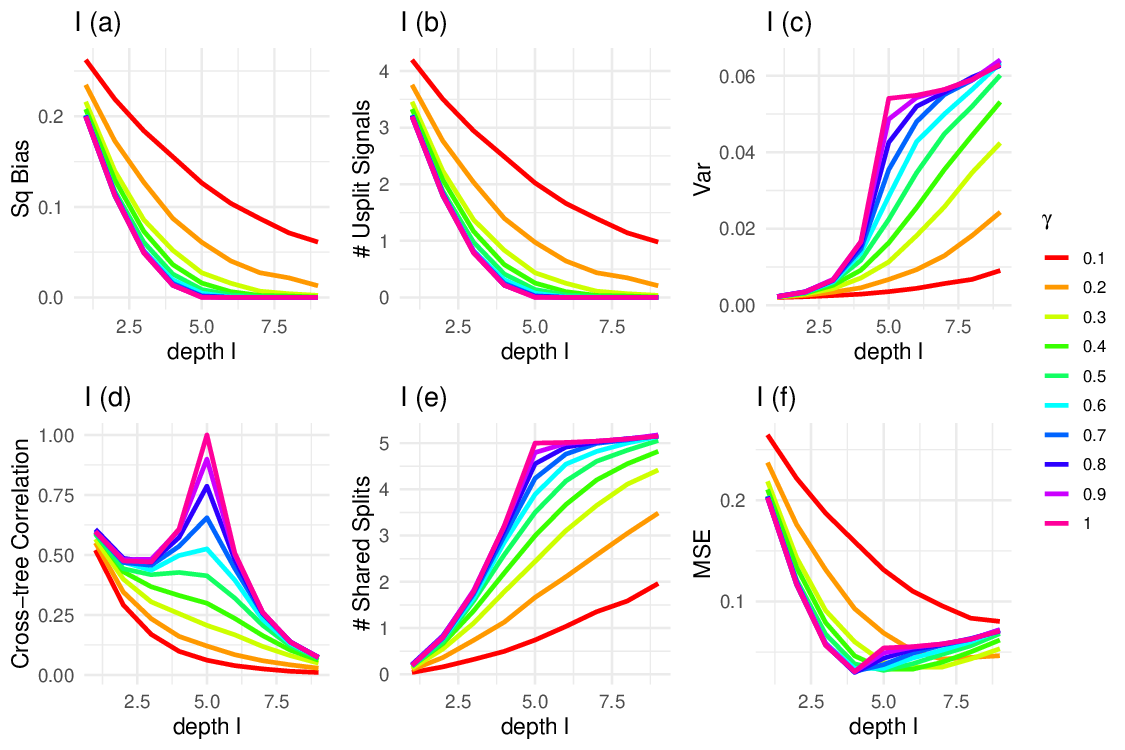}
    \caption{Performance measures (a)--(f) for forests in the binary case as $\gamma$ and $l$ vary under configuration (I).}
    \label{fig:simu9}
\end{figure}

\begin{figure}[h]
    \centering
    \includegraphics[width = 0.8\textwidth]{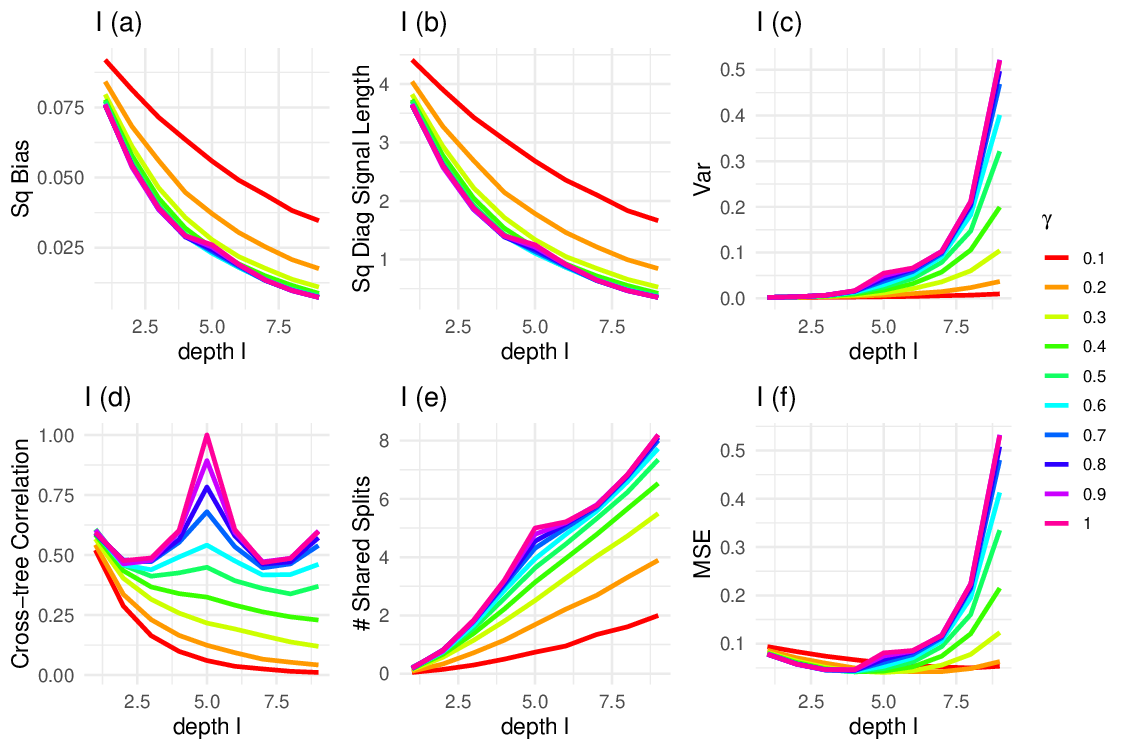}
    \caption{Performance measures (a)--(f) for forests in the continuous case when  $\gamma$ and $l$ vary under configuration (I).}
    \label{fig:simu10}
\end{figure}

This section provides further insights into the contribution of Type I exogenous randomness toward the success of RF. 
Our simulation results on the binary feature example in Section \ref{new.Sec.gamma1} suggest that when $\gamma=1$, non-informative features contribute to the tree-building only after depth $l>5$. Here, we demonstrate that feature subsampling allows non-informative features to contribute earlier than depth $l=5$, and thus further reduces cross-tree correlations across all tree depths (see Figure \ref{fig:simu9}(d)).  
The performance measures (a)--(f) are plotted in Figures \ref{fig:simu9} and \ref{fig:simu10} below for Equal Configuration (I), and in Figures \ref{fig:simu3} and \ref{fig:simu6} in Section \ref{new.SecC.3} for Unequal Configuration (II), respectively, as we vary $l$.

For configuration (I), recall that our simulation results in last subsection already demonstrate the effect of Type II exogenous randomness when $\gamma = 1$. The feature subsampling adds Type I exogenous randomness into the tree-building process, and thus, non-informative features now have a chance to be split earlier than informative ones at the first $s$ tree depths; see Figure \ref{fig:simu9}(b) for the increasing number of unsplit informative features at each fixed tree depth when $\gamma$ decreases.
This early involvement of non-informative features 
has a strong effect on reducing cross-tree correlation in high-dimensional settings, leading to a significant decrease in variance and correlation as $\gamma$ decreases at each fixed tree depth; see Figures \ref{fig:simu9}(c) and (d). Indeed,  Figure \ref{fig:simu9}(d) shows that the cross-tree correlation stays below 1 for all $\gamma<1$ even when tree depth $l=s=5$, a distinction from $\gamma=1$ scenario where $l=s=5$ has cross-tree correlation 1.  
The decreasing patterns of variance and cross-tree correlation are supported by the reduction in shared splits between $I_l$ and $I_l'$ as shown theoretically in (\ref{eq:binaryRF}) and empirically in Figure \ref{fig:simu9}(e). 
However, such variance reduction comes at the cost of increased squared bias (especially for small $l$), as shown in Figure \ref{fig:simu9}(a), because of the heavy involvement of non-informative features.  
This directly leads to a bias-variance tradeoff (Figure \ref{fig:simu9}(f)), and  $\gamma = 1$ is not optimal as $l$ increases. 

For the continuous case, recall the discussion in Section \ref{new.Sec.gamma1} that non-informative features never enter the tree-building process when $\gamma = 1$. Thanks to Type I exogenous randomness, for $\gamma < 1$, non-informative features have a chance to be split and thus reduce cross-tree correlation, leading to similar patterns in performance measures to those in the binary case; see Figure \ref{fig:simu10}. 

A surprising observation is that feature subsampling does not always increase the squared bias while reducing the variance at the early stage of tree-building. We present simulation evidence on this under Unequal Configuration (II) in Section \ref{new.SecC.3} and provide detailed discussions therein (see Figures \ref{fig:simu3} and \ref{fig:simu6}). Together with the preceding discussion, our finding clearly shows that feature subsampling can reduce \textit{both} bias and variance when $\gamma$ is not too small, where the bias reduction phenomenon may be model-specific and the variance reduction could be a robust phenomenon across models.


Although (I) and (II) represent two specific configurations, they share the same mechanisms where feature subsampling improves RF's performance by early involvement of non-informative variables. These results underscore the \textit{generality} of the feature subsampling effect in enhancing RF's performance across various coefficient configurations.

\section{General results on partitioning estimator ensemble}\label{new.Sec.modsetandpart}

The MSE expansion results in Section \ref{sec:CART} can hold more generally beyond the linear model and population CART partitioning rule. 
This section establishes these general results. 
We will work under the general setting presented in Section \ref{sec:partition-ensemble}. It is worth noting that the MSE results in Section \ref{sec:CART} are obtained by applying the general results here to the two specific examples therein.




\begin{assumption}[On $\vv{X}$]\label{assm:1}
   The random vector $\vv{X}$ can take either continuous or discrete values with bounded joint density function (or mass function) from the above and below.

\end{assumption}

\begin{assumption}[On $\mu$]\label{assm:2}
    The ground truth function $\mu(\cdot)$ belongs to the class $\mathcal{M}(\mathcal{X}^d,\mathcal{Y})$, the collection of measurable functions mapping from $\mathcal{X}^d$ to $\mathcal{Y}$. Continuity is not required.
\end{assumption}

\begin{assumption}[On $\sigma^2$]\label{assm:3}
     The function $\sigma^2(\cdot)$ is bounded from both above and below.
\end{assumption}

 Two sets $A$ and $B$ are \textit{indistinguishable} under probability $\P_{\vv{X}}$ if 
    $A\subset B$ and $\P(A) = \P(B)$.  We treat indistinguishable sets as equivalent (equal), which allows us to simplify the technical derivations.
In addition, two partitions $P$ and $P'$ are \textit{indistinguishable} if for any pair of $P_i \in P$ and $P'_j \in P'$ with $P_i\cap P'_j \neq \varnothing$, sets $P_i$ and $P'_j$ are indistinguishable.    

Recall the definition of partitioning rule $\vv{P}$, space $\mathcal{D}$ of decision factors, and exogenous random element $\Theta$ introduced right after Definition \ref{def:exogenous-randomness}. We add superscripts to $\vv{P}$ and $\mathcal{D}$ to emphasize their dependence on $n$.

\begin{assumption}[On $\vv{P}^n$]\label{assm:4}
(Regularity of the partitioning rule). 
Let $\mathcal{D}^n_{\Theta}\subset \mathcal{D}^n$ be the range of exogenous random element $\Theta$ and $\mathcal{P}^n_0 = \{\vv{P}^n(D):D \in \mathcal{D}^n_\Theta\}$ the collection of all potential outcomes of $\vv{P}^n(\Theta)$. The following regularity conditions hold:
    
    (1) Each cell in a partition $P \in \mathcal{P}_0$ has a positive probability under $\P_{\vv{X}}$.

    (2) For cells $P_1 \in P$ and $P_2 \in P$ with $P$ and $P'$ two distinct elements in $\mathcal{P}^n_0$, probability  $\P_{\vv{X}}(P_1 \cap P_2)>0$ if $P_1 \cap P_2 \neq \varnothing$. 

    (3) Any distinct partitions $P$ and $P'$ in $\mathcal{P}^n_0$  are distinguishable.
    
Moreover,  $\E_{\vv{X}}(\mu|\vv{P}^n_{\vv{x}}(\Theta))$, $\P_{\vv{X}}(\vv{P}^n_{\vv{x}}(\Theta))$, and $I_{\vv{P}^n_{\vv{x}}(\Theta)}$ are bivariate measurable functions on  $\mathcal{X}^d \times \Omega$, where $\vv{P}^n_{\vv{x}}(\Theta)$ is the cell in  $\vv{P}^n(\Theta)$ that contains $\vv{x}$.  
\end{assumption}




Same as in Section \ref{sec:partition-ensemble}, let $\hat\mu(\vv x; \vv Z, \vv{P}^n( \Theta))$ be either $\hat{\mu}_{\part}$ or $\hat{\mu}_{\ens}$ as in \eqref{defn:part0} and \eqref{defn:ens0}. 
In what follows, we analyze the GMSEs in (\ref{defn:MSE}) for both  $\hat{\mu}_{\part}$ and $\hat{\mu}_{\ens}$,  and reveal how exogenous randomness can help improve the performance of the ensemble estimator over the individual partitioning estimator. 
 For such purposes, we propose to consider 
 the following decomposition of GMSE
\begin{align}\label{G-MSE-decomp}
    \hbox{MSE}(\hat{\mu}) = \underbrace{\E_{\vv{X},\Theta}[\tilde{\mu}( \vv{X};\vv{P}^n(\Theta))- \mu(\vv{X})]^2}_{Bias} + \underbrace{\E_{\vv{X},\Theta}[{\rm Var}_{\vv Z}(\hat{\mu}(\vv{X};\vv{Z},\vv{P}^n(\Theta)))]}_{Var},
\end{align}
where 
$\tilde{\mu}(\vv x; \vv{P}^n( \Theta)) = \E_{\vv Z}[\hat{\mu}(\vv x;\vv{Z}, \vv{P}^n( \Theta))]$
with the expectation taken with respect to training sample $\vv Z$.
The major distinction of our MSE decomposition in (\ref{G-MSE-decomp}) above from the commonly used one in the literature is that we define the bias \textit{differently} as  $\tilde{\mu}(\vv x; \vv{P}^n(\Theta)) -\mu(\vv x)$ by first integrating out the training data $\vv Z$.

It is seen that $\tilde{\mu}(\vv x; \vv{P}^n( \Theta)) -\mu(\vv x)$ is precisely the bias of a single partitioning estimator conditional on   $\vv{P}^n(\Theta)$. Similarly, ${\rm Var}_{\vv Z}(\hat{\mu}(\vv x;\vv{Z},\vv{P}^n(\Theta)))$ is the variance of individual tree estimator conditional on  $\vv{P}^n(\Theta)$. In this sense, the decomposition in \eqref{G-MSE-decomp} averages over individual partitioning estimators' squared bias and variance. This innovation allows us to see the effect of ensemble more clearly and obtain the results in this paper, including the ones presented in the last section.

Next, we introduce some necessary notations to facilitate the presentation of our main results. In the same spirit as \eqref{eq:modindpcov} and \eqref{eq:modindpvar},
we define the global version of the signal-induced  and model-error-induced  cross-partition covariance functions, respectively, as 
    \begin{equation}\label{eq:globalCov}
    \begin{split}
        \text{Cov}_{\mu}(P,P') &= \sum_{\substack{P_i\in P,\\ P'_j \in P'}}
        \E_{\vv{X}}
        \left(
        (\mu - \E_{\vv{X}}(\mu|P_i))
        (\mu - \E_{\vv{X}}(\mu|P'_j)|
        P_i\cap P'_j
        \right)
        \frac{\P_{\vv{X}}(P_i\cap P'_j)^2}{ \P_{\vv{X}}(P_i) \P_{\vv{X}}(P'_j)} ,\\
        \text{Cov}_{\sigma^2}(P,P') &= \sum_{\substack{P_i\in P,\\ P'_j \in P'}} 
        \E_{\vv{X}}
        \left(\sigma^2|
        P_i\cap P'_j
        \right)
        \frac{\P_{\vv{X}}(P_i\cap P'_j)^2}{ \P_{\vv{X}}(P_i) \P_{\vv{X}}(P'_j)},
        \end{split}
    \end{equation}
    where $P = \{P_i\}$ and $P'=\{P'_j\}$ are elements in $\mathcal{P}_0^n$. In particular, when $P = P'$, we have the following signal-induced and model-error-induced variance functions 
    \begin{equation}\label{eq:globalVar}
    \begin{split}
    \text{Var}_{\mu}(P) &= \text{Cov}_{\mu}(P,P) = \sum_{\substack{P_i\in P}} \text{Var}_{\vv{X}}(\mu|P_i),\\
     \text{Var}_{\sigma^2}(P) &= \text{Cov}_{\sigma^2}(P,P) = \sum_{\substack{P_i\in P}} 
     \E_{\vv{X}}(\sigma^2| P_i).
     \end{split}
     \end{equation}
     The cross-partition covariance functions are bivariate functions on partitions and arise naturally in studying the covariance of two partitioning estimators. 
    Cross-partition covariance functions satisfy the Cauchy-Schwarz inequality, as formally summarized in the proposition below.     The proof is provided in Section \ref{new.SecA.prop2}.  
    \begin{proposition}
    \label{prop:globaltrieq}
    Let $P$ and $P'$ be two components in $\mathcal{P}^n_0$. Then it holds that 
        \begin{equation}\label{eq:globaltri}
     \begin{split}
     \ \text{Cov}_{\mu}(P,P') \leq \sqrt{\text{Var}_{\mu}(P) \text{Var}_{\mu}(P')}, \ \
           \text{Cov}_{\sigma^2}(P, P) \leq \sqrt{\text{Var}_{\sigma^2}(P) \text{Var}_{\sigma^2}(P')}.
     \end{split}
     \end{equation}
For the first result, the equality holds if and only if for any $P_i\in P$, $P'_j\in P'$ with $P_i\cap P'_j\neq\varnothing$, either $P_i$ and $P'_j$ are indistinguishable or
    $\mu$ is constant on $P_i\cup P'_j$. For the second result,
    the equality holds if and only if $P$ and $P'$ are indistinguishable. Consequently, Assumption \ref{assm:4}(3) ensures that the second result in \eqref{eq:globaltri} is a strict inequality, and that when $\mu$ is a non-constant function, the first result in \eqref{eq:globaltri} is a strict inequality. 
    \end{proposition}

\begin{theorem} 
\label{new.thm2} 
Assume that the following 
condition holds 
\begin{equation}\label{cond:globaluniform}
       \sup_{n} \E_{\Theta}\left[n^{-1}|\vv{P}^n(\Theta)|\right] < \infty, 
    \end{equation}
    where $|\vv{P}^n(\Theta)|$ is the total number of cells in $\vv{P}^n(\Theta)$.
    Then the MSE of $\hat{\mu}_{ens}$ satisfies
    \begin{equation}\label{eq:MSEens}
        \begin{split}
            \hbox{MSE}(\hat{\mu}_{ens}) = & \underbrace{\frac{B-1}{B}\E_{\vv{X}}\left[(
            \mu - \E_{\Theta}(\mu_{\vv{P}^n(\Theta)}))^2\right]}_{\text{squared bias of ensemble}} + \frac{1}{B}\underbrace{ \E_{\vv{X},\Theta}\left[(\mu - \mu_{\vv{P}^n(\Theta)})^2\right]}_{\text{squared bias of a single partition}} \\
            + & \underbrace{\frac{B-1}{B n} \E_{\Theta,\Theta'}\left\{\text{Cov}_{\mu}(\vv{P}^n(\Theta'),\vv{P}^n(\Theta')) + \text{Cov}_{\sigma^2}(\vv{P}^n(\Theta'),\vv{P}^n(\Theta'))\right\}}_{\text{cross-partition covariance}}\\
             +&\frac{1}{B}\underbrace{\E_\Theta\left\{\frac{1}{ n}\text{Var}_{\mu}(\vv{P}^n(\Theta)) + \frac{1}{ n}\text{Var}_{\sigma^2}(\vv{P}^n(\Theta))\right\}}_{\text{single-partition variance}}
             + \mathcal{R}_{\ens}(\mu)
             ,
        \end{split}
    \end{equation}
    where $\Theta$ and $\Theta'$ are independent copies of the exogenous random effects, and 
    \begin{align*}
        &\mathcal{R}_{\ens}(\mu) \lesssim (\|\mu\|_\infty + \|\sigma^2\|_\infty)^2 \times \Bigg\{\E_{\Theta}\left(\sum_{P_i\in \vv{P}(\Theta)} (1 - \P_{\vv{X}}(P_i))^n \vv{P}_{\vv{X}}(P_i)\right) \\
        & + \frac{B-1}{B}\E_{\Theta}\left(\frac{1}{n}\sum_{P_i \in \vv{P}(\Theta)} \frac{1}{(1 + (n-1)\P_{\vv{X}}(P_i))^{1/2}}\right) + \E_{\Theta}\left(\frac{1}{n}\sum_{P_i \in \vv{P}(\Theta)} \frac{1}{1 + (n-1)\P_{\vv{X}}(P_i)}\right)\Bigg\}
    \end{align*}
    with the constant in $\lesssim$ independent of $\vv{x}_0$, $\mu$, $\sigma^2$, $\vv{P}^n(\Theta)$, and $n$. 
    The MSE of a single partitioning estimator $\hat{\mu}_{\part}$ corresponds to the special case of $B=1$ in (\ref{eq:MSEens}).
  
\end{theorem}
   
Theorem \ref{new.thm2} above is built on a local MSE bound for ensemble estimators detailed in Section \ref{new.Sec.local}, and its proof is provided in Section \ref{new.SecA.3}. 
    Condition \eqref{cond:globaluniform} indicates that the partitioning rule is allowed to vary as sample size $n$ increases. Under Assumption \ref{assm:4}(3) and when $\mu$ is non-constant, the results in Proposition \ref{prop:globaltrieq} become strict inequalities. Thus, the leading terms of both squared bias and variance for the ensemble estimator are strictly less than those of a single partitioning estimator, showing the advantage of the ensemble estimator. Same as in the two examples presented in Section \ref{sec:CART}, as $B\rightarrow\infty$, the leading terms in MSE decomposition for ensemble estimator are driven by the pairwise interactions among partitions resulted from exogenous randomness, and the effect from single partitioning estimators vanishes completely. These theoretical results support the new insights discussed in the Introduction.

We next present the consistency results below, where we will make use of the model-independent cross-partition covariance and variance functions defined in Section \ref{sec:partition-ensemble}.

\begin{theorem}
\label{new.thm3}
Suppose Assumptions \ref{assm:1}--\ref{assm:4} hold. Assume that condition (\ref{cond:globaluniform}) holds for each $n\geq 1$, and with probability one, each cell $P^n_i$ in partition $\vv{P}^n(\Theta)$ satisfies
    \begin{equation}\label{cond:cellprob}
        n \P_{\vv{X}}(P^n_i) \to +\infty.
    \end{equation}
    Then
    for the partitioning estimator $\hat{\mu}_{\part,n}$, the sufficient and necessary condition for the weak consistency of $\hat{\mu}_{\part,n}$ is
\begin{align}
    \E_{\vv{X},\Theta}\left[(\mu - \mu_{\vv{P}^n(\Theta)})^2\right] \to 0~~~\text{and}~~~\frac{\E_{\Theta}[|\vv{P}^n(\Theta)|]}{n}\to 0.\label{eq:partbiasconv}
\end{align}    
    Meanwhile, as $B \to \infty$,
    the ensemble estimator $\hat{\mu}_{\ens,n}$ is weakly consistent if and only if
\begin{align}
    \E_{\vv{X}}\left[(\mu - \E_{\Theta}(\mu_{\vv{P}^n(\Theta)}))^2\right] \to 0~~~\text{and}~~~\frac{1}{n}\E_{\Theta,\Theta'}\left[\text{Cov}(\vv{P}^n(\Theta),\vv{P}^n(\Theta'))\right] \to 0.\label{eq:ensbiasconv}
\end{align}
In particular, condition (\ref{eq:partbiasconv}) implies condition (\ref{eq:ensbiasconv}).
\end{theorem}

The proof of Theorem \ref{new.thm3} above is provided in Section \ref{new.SecA.4}. Since condition \eqref{eq:partbiasconv} for individual partitioning estimator consistency is stronger than condition \eqref{eq:ensbiasconv} for ensemble estimator consistency, we obtain immediately that the ensemble estimator may be consistent under weaker conditions than its individual partitioning estimator. Indeed, the consistency rates in Theorems \ref{new.thm.binary} and \ref{new.thm.unif} are obtained by bounding the individual tree estimator, which automatically provides an upper bound for the random forests estimator.

\bibliographystyle{chicago}
\bibliography{RF}

	
\newpage
\appendix
\setcounter{page}{1}
\setcounter{section}{0}
\renewcommand{\theequation}{A.\arabic{equation}}
\setcounter{equation}{0}

\renewcommand{\thetheorem}{A.\arabic{theorem}}
\setcounter{theorem}{0}

\renewcommand{\theproposition}{A.\arabic{proposition}}
\setcounter{proposition}{0}

\begin{center}{\bf \Large Supplementary Material to ``Exogenous Randomness Empowering Random Forests''}
		
\bigskip
		
Tianxing Mei, Yingying Fan and Jinchi Lv 
\end{center}
	
\noindent This Supplementary Material contains the proofs of Theorems \ref{new.thm.binary}--\ref{new.thm3} and Proposition \ref{prop:globaltrieq}, some key lemmas and their proofs, and additional materials. All the notation is the same as defined in the main text.

\section{Local MSE bounds for ensemble estimators}
\label{new.Sec.local}

\subsection{Main results}
Let $\hat\mu(\vv x; \vv Z, \vv{P}^n( \Theta))$ be either $\hat{\mu}_{\part}$ or $\hat{\mu}_{\ens}$ as in \eqref{defn:part0} and \eqref{defn:ens0} in Section \ref{sec:partition-ensemble}. 
For each given $\vv{x}_0\in\mathcal{X}^d$, the \textit{local mean squared error (LMSE)} 
is defined as
\begin{equation}\label{defn:LMSE}
    \hbox{MSE}(\hat{\mu} ; \vv{x}_0) = \E_{\vv{Z},{\Theta}}\left[(\mu(\vv{x}_0) - \hat{\mu}(\vv{x}_0; \vv Z, \vv{P}^n( \Theta)))^2\right].
\end{equation}
In addition, we define the signal-induced  and model-error-induced local cross-partition covariance functions, respectively, as
    \begin{equation}\label{eq:localCov}
    \begin{split}
        \text{Cov}_{\mu}(P,P';\vv{x}_0) &= \E_{\vv{X}}
        \left(
        (\mu - \E_{\vv{X}}(\mu|P_{\vv{x}_0}))
        (\mu - \E_{\vv{X}}(\mu|P'_{\vv{x}_0}))|
        P_{\vv{x}_0} \cap P'_{\vv{x}_0}
        \right)\\
        & \times
        \frac{\P_{\vv{X}}(P_{\vv{x}_0} \cap P'_{\vv{x}_0})}{ \P_{\vv{X}}(P_{\vv{x}_0}) \P_{\vv{X}}(P'_{\vv{x}_0})} ,\\
        \text{Cov}_{\sigma^2}(P,P';\vv{x}_0) &= \E_{\vv{X}}
        \left(\sigma^2|
        P_{\vv{x}_0} \cap P'_{\vv{x}_0}
        \right)
        \frac{\P_{\vv{X}}(P_{\vv{x}_0} \cap P'_{\vv{x}_0})}{ \P_{\vv{X}}(P_{\vv{x}_0}) \P_{\vv{X}}(P'_{\vv{x}_0})},
    \end{split}
    \end{equation}
    where $P$ and $P'$ are elements in $\mathcal{P}_0$. In particular, when $P = P'$, denote by 
    \begin{equation}\label{eq:localVar}
        \begin{split}
            \text{Var}_{\mu}(P;\vv{x}_0) & = \text{Cov}_{\mu}(P,P;\vv{x}_0) = \frac{\text{Var}_{\vv{X}}(\mu | P_{\vv{x}_0})}{\P_{\vv{X}}(P_{\vv{x}_0})},\\
            \text{Var}_{\sigma^2}(P;\vv{x}_0) & = \text{Cov}_{\sigma^2}(P,P;\vv{x}_0) = \frac{\E_{\vv{X}}(\sigma^2 | P_{\vv{x}_0})}{\P_{\vv{X}}(P_{\vv{x}_0})}.
        \end{split}
    \end{equation}
      The cross-partition covariance functions capture the covariance of partitioning estimators generated by $P$ and $P'$ locally around the target point $\vv{x}_0$, and they also exhibit a similar Cauchy--Schwarz property as the global cross-partition covariance functions. 
    \begin{proposition}[Cauchy--Schwarz inequality for local cross-partition covariance functions]\label{prop:localtrieq}
    Let $P$ and $P'$ be two components in $\mathcal{P}^n_0$ (defined in Assumption \ref{assm:4}). Then the following inequality holds
        \begin{equation}\label{eq:localtri}
     \begin{split}
     \text{Cov}_{\mu}(P,P';\vv{x}_0) &\leq \sqrt{\text{Var}_{\mu}(P;\vv{x}_0) \text{Var}_{\mu}(P';\vv{x}_0)},\\
           \text{Cov}_{\sigma^2}(P,P';\vv{x}_0) &\leq \sqrt{\text{Var}_{\sigma^2}(P;\vv{x}_0) \text{Var}_{\sigma^2}(P';\vv{x}_0)}.
     \end{split}
     \end{equation}
For the first result, the equality holds if and only if $P_{\vv{x}_0}$ are $P'_{\vv{x}_0}$ are indistinguishable, or
    $\mu$ is constant on $P_{\vv{x}_0}\cup P'_{\vv{x}_0}$. For the second result, the equality holds if and only if $P_{\vv{x}_0}$ are $P'_{\vv{x}_0}$ are indistinguishable. Consequently, Assumption \ref{assm:4}(3) ensures that the second result in \eqref{eq:localtri} is a strict inequality, and that when $\mu$ is a non-constant function, the first result in \eqref{eq:localtri} is a strict inequality. 
    \end{proposition}
     The proof of Proposition \ref{prop:localtrieq} above is provided later in Section \ref{new.SecA.prop1}.

\begin{theorem}[\color{black} Local MSE bounds] \label{new.thm1}
Assume that for any fixed target point $\vv{x}_0\in\mathcal{X}^d$, the random cell $\vv{P}^n_{\vv{x}_0}(\Theta)$ containing the target point satisfies 
    \begin{equation}\label{localuniform}
       \sup_{n} \E_{\Theta}\left[\frac{1}{n \P_{\vv{X}}(\vv{P}^n_{\vv{x}_0}(\Theta))}\right] < \infty. 
    \end{equation}
    Then under Assumptions \ref{assm:1}--\ref{assm:4}, the local MSE of the ensemble estimator satisfies 
    \begin{equation}\label{eq:localMSEens}
        \begin{split}
            \hbox{MSE}(\hat{\mu}_{\ens};\vv{x}_0) 
             =& \underbrace{\frac{B-1}{B} \left(\mu(\vv{x}_0) - \E_{\Theta}\left(\E_{\vv{X}}(\mu|\vv{P}^n_{\vv{x}_0}(\Theta))\right)\right)^2}_{\text{squared bias of ensemble}}
             \\
             + &  \underbrace{\frac{1}{B}\E_{\Theta}\left[(\mu(\vv{x}_0) - \E_{\vv{X}}(\mu|\vv{P}^n_{\vv{x}_0}(\Theta)))^2\right]}_{\text{squared bias of a single partition}}\\
            +&  \underbrace{\frac{B - 1}{B n} \mathbb{E}_{\Theta,\Theta'}\left[\text{Cov}_{\mu}(\vv{P}^n(\Theta), \vv{P}^n(\Theta');\vv{x}_0) + \text{Cov}_{\sigma^2}(\vv{P}^n(\Theta), \vv{P}^n(\Theta');\vv{x}_0)\right]}_{\text{cross-partition covariance}}\\
            +& \underbrace{\frac{1}{B n} \E_{\Theta}\left[\text{Var}_{\mu}(\vv{P}^n(\Theta);\vv{x}_0) + \text{Var}_{\sigma^2}(\vv{P}^n(\Theta);\vv{x}_0)\right]}_{\text{single-partition variance}}\\
            +&\mathcal{R}_{\ens}(\mu;\vv{x}_0)
             ,
        \end{split}
    \end{equation}
    in which $\Theta$ and $\Theta'$ are mutually independent and the remainder $\mathcal{R}_{\ens}(\mu; \vv{x}_0)$ satisfies
    \begin{align*}
        \mathcal{R}_{\ens}(\mu;\vv{x}_0) & 
        \lesssim (\|\mu\|_{\infty} + \|\sigma^2\|_{\infty})^2\\
        & \times \Bigg\{\E_{\Theta}((1-\P_{\vv{X}}(\vv{P}^n_{\vv{x}_0}(\Theta)))^n) \\
        & + \frac{B-1}{B} \E_{\Theta,\Theta'}\left(\frac{\P_{\vv{X}}(\vv{P}^n_{\vv{x}_0}(\Theta)\cap \vv{P}^n_{\vv{x}_0}(\Theta'))}{n \P_{\vv{X}}(\vv{P}^n_{\vv{x}_0}(\Theta)) \P_{\vv{X}}(\vv{P}^n_{\vv{x}_0}(\Theta'))(1 + (n-1)\P_{\vv{X}}(\vv{P}^n_{\vv{x}_0}(\Theta)))}\right)\\
        & + \frac{B-1}{B}\E_{\Theta}\left(\frac{1}{(1+(n-1)\P_{\vv{X}}(\vv{P}^n_{\vv{x}_0}(\Theta)))^{3/2}}\right)\\
        &+\frac{1}{B}\E_{\Theta}\left( \frac{1}{n \P_{\vv{X}}(\vv{P}^n_{\vv{x}_0}(\Theta))(1 + (n -1) \P_{\vv{X}}(\vv{P}^n_{\vv{x}_0}(\Theta))}\right)\Bigg\}
    \end{align*}
   with the constant in $\lesssim$ independent of $\vv{x}_0$, $\mu$, $\sigma^2$, $\vv{P}^n(\Theta)$, and $n$. 
   In particular, the local MSE of a single partitioning estimator $\mu_{\part}$ is the special case of $B=1$ in (\ref{eq:localMSEens}).
   


\end{theorem}
The proof of Theorem \ref{new.thm1} above is provided in Section \ref{new.SecA.2} later.

\subsection{Technical preparation} \label{new.SecA.1}

This section is devoted to providing an expansion of the local MSE, based on which our main results are built. 
In what follows, denote by $\E_{\vv{X}|\Theta,\Theta'}(\mu) = \E(\mu(\vv{X};\Theta,\Theta')|\Theta,\Theta')$ the conditional expectation of function $\mu(\vv{X};\Theta,\Theta')$ given the exogenous factors $(\Theta,\Theta')$. In other words, given the remaining two $\Theta$'s, we integrate with respect to $\vv{X}$. Similarly, $\E_{\epsilon|\vv{X},\Theta,\Theta'}(\cdot)$ is the conditional expectation with respect to $\epsilon$ given $(\vv{X},\Theta,\Theta')$. To simplify the notation, we omit the superscript $n$ in $\vv{P}^n$ throughout the proof whenever there is no confusion.

We consider the local-version decomposition of MSE as in (\ref{G-MSE-decomp}) 
\begin{equation}\label{eq:decomposition}
 \begin{split}
      \hbox{MSE}(\hat{\mu}_{ens}; \vv{x}_0) & = \E_{\vv{Z},\Theta_{1:B}}[(\mu(\vv{x}_0) - \hat{\mu}_{ens}(\vv{x}_0;\vv{Z},\vv{P}(\Theta_{1:B}))^2] \\
     &= \underbrace{\E_{\Theta_{1:B}}\left[(\mu(\vv{x}_0)) - \E_{\vv{Z}|\Theta_{1:B}}[\hat{\mu}_{ens}(\vv{x}_0;\vv{Z},\vv{P}(\Theta_{1:B}))])^2\right]}_{\text{Bias}^2(\vv{x}_0)} \\
     & + \underbrace{\E_{\Theta_{1:B}}\left[\hbox{Var}_{\vv{Z}|\Theta_{1:B}}[\hat{\mu}_{ens}(\vv{x}_0;\vv{Z},\vv{P}(\Theta_{1:B}))]\right]}_{\text{Var}(\vv{x}_0)},
\end{split}
\end{equation}
where 
$\tilde{\mu}_{\ens}(\vv x_0; \vv{P}^n( \Theta)) = \E_{\vv Z}[\hat{\mu}_{\ens}(\vv x_0;\vv{Z}, \vv{P}^n( \Theta))]$
with the expectation taken with respect to the training sample $\vv Z$.

For further investigation of the structure of the local MSE in (\ref{eq:decomposition}), let us introduce the following notation
\begin{align}
    \tilde{\mu}(\vv{x}_0;\Theta) & :=  \E_{\vv{Z}|\Theta}(\hat{\mu}_{\part}(\vv{x}_0;\vv{Z},\vv{P}(\Theta))),
    \label{eq:defntildemu}\\
    V_1(\vv{x}_0;\Theta) &:= \text{Var}_{\vv{Z}|\Theta}[\hat{\mu}_{\part}(\vv{x}_0;\vv{Z},\vv{P}(\Theta))],
    \label{eq:defnV1}\\
    V_2(\vv{x}_0;\Theta,\Theta') &: = \text{Cov}_{\vv{Z}|\Theta,\Theta'}[\hat{\mu}_{\part}(\vv{x}_0;\vv{Z},\vv{P}(\Theta)), \hat{\mu}_{\part}(\vv{x}_0;\vv{Z},\vv{P}(\Theta'))].
    \label{eq:defnV2}
\end{align}
The meanings of these quantities are given below. 
The notation $\tilde{\mu}(\vv{x}_0;\Theta)$ stands for a function of the exogenous factor $\Theta$ by integrating over the training data of a single partitioning estimator. 
The function $V_1(\vv{x}_0;\Theta)$ refers to the conditional variance of the partitioning estimator given the partition $\vv{P}(\Theta)$ and thus is also a function of $\Theta$.
The bivariate function $V_2(\vv{x}_0;\Theta,\Theta')$ represents the conditional covariance of two partitioning estimators when the corresponding partitions $\vv{P}(\Theta)$ and $\vv{P}(\Theta')$ are fixed.

On the one hand, for the conditional expectation part, it holds that 
\begin{align*}
    \E_{\vv{Z}|\Theta_{1:B}}(\hat{\mu}_{\ens}(\vv{x}_0;\vv{Z},\vv{P}(\Theta_{1:B}))) &= \frac{1}{B}\sum_{b=1}^B \E_{\vv{Z}|\Theta_{b}}(\hat{\mu}_{\part}(\vv{x}_0;\vv{Z},\vv{P}(\Theta_{b})))\\
    & = \frac{1}{B}\sum_{b=1}^B \tilde{\mu}(\vv{x}_0;\Theta_b),
\end{align*}
in which the right-hand side (RHS) is a sum of independent and identically distributed (i.i.d.) random variables. Hence, the conditional expectation part in (\ref{eq:decomposition}) reduces to
\begin{align*}
    &\E_{\Theta_{1:B}}\left[(\mu(\vv{x}_0) - \E_{\vv{Z}|\Theta_{1:B}}(\hat{\mu}_{\ens}(\vv{x}_0;\vv{Z},\vv{P}(\Theta_{1:B}))))^2\right]\\
    = & \E_{\Theta_{1:B}}\left\{\left( 
    \mu(\vv{x}_0) - 
    \frac{1}{B} \sum_{b=1}^B \tilde{\mu}(\vv{x}_0; \Theta_b) \right)^2\right\}\\
    = & \frac{1}{B} \E_{\Theta}\left\{(\mu(\vv{x}_0) - \tilde{\mu}(\vv{x}_0;\Theta))^2\right\} \\
    + & \frac{B-1}{B} \E_{\Theta,\Theta'}\left\{(\mu(\vv{x}_0) - \tilde{\mu}(\vv{x}_0;\Theta))(\mu(\vv{x}_0) - \tilde{\mu}(\vv{x}_0;\Theta'))\right\}\\
    = & \frac{1}{B} \E_{\Theta}\left\{(\mu(\vv{x}_0) - \tilde{\mu}(\vv{x}_0;\Theta))^2\right\}
    + \frac{B -1}{B} (\mu(\vv{x}_0) - \E_{\Theta}[\tilde{\mu}(\vv{x}_0;\Theta)])^2.
\end{align*}
Observe that $\tilde{\mu}(\vv{x}_0;\Theta)$ and $\tilde{\mu}(\vv{x}_0;\Theta')$ are independent and identically distributed. We immediately see that 
\begin{equation} \label{eq:BiasDecomposition}
 \begin{split}   \text{Bias}^2(\vv{x}_0) = &\E_{\Theta_{1:B}}\left[(\mu(\vv{x}_0) - \E_{\vv{Z}|\Theta_{1:B}}(\hat{\mu}_{\ens}(\vv{x}_0;\vv{Z},\vv{P}(\Theta_{1:B}))))^2\right]\\
    = & \frac{1}{B} \E_{\Theta}\left\{(\mu(\vv{x}_0) - \tilde{\mu}(\vv{x}_0;\Theta))^2\right\}
    + \frac{B -1}{B} (\mu(\vv{x}_0) - \E_{\Theta}[\tilde{\mu}(\vv{x}_0;\Theta)])^2\\
    = & (\mu(\vv{x}_0) - \E_{\Theta}[\tilde{\mu}(\vv{x}_0;\Theta)])^2 + \frac{1}{B} \text{Var}_{\Theta}(\tilde{\mu}(\vv{x}_0;\Theta)).
    \end{split}
\end{equation}

On the other hand, for the conditional variance part, we have
\begin{align*}
    & \text{Var}_{\vv{Z}|\Theta_{1:B}}(\hat{\mu}_{\ens}(\vv{x}_0);\vv{Z},\vv{P}(\Theta_{1:B})))\\
   = & \frac{1}{B^2} \sum_{b=1}^B \text{Var}_{\vv{Z}|\Theta_{b}}\left[\hat{\mu}_{\part}(\vv{x}_0;\vv{Z},\vv{P}(\Theta_{b}))\right]\\
   + & \frac{1}{B^2} \sum_{b\neq b'} \text{Cov}_{\vv{Z}|\Theta_{b},\Theta_{b'}}\left[\hat{\mu}_{\part}(\vv{x}_0;\vv{Z},\vv{P}(\Theta_{b})),\hat{\mu}_{\part}(\vv{x}_0;\vv{Z},\vv{P}(\Theta_{b'}))\right]\\
   =& \frac{1}{B^2} \sum_{b=1}^B V_1(\vv{x}_0;\Theta_b) + \frac{1}{B^2} \sum_{b\neq b'} V_2(\vv{x}_0;\Theta_b,\Theta_{b'}).
\end{align*}
Since $\Theta_1,\ldots,\Theta_B$ are i.i.d. copies of the exogenous decision factors, both $\{V_1(\vv{x}_0;\Theta_b)\}$ and $\{V_2(\vv{x}_0;\Theta_b,\Theta_{b'})\}$ are collections of identically distributed variables, and as such, the conditional variance term in (\ref{eq:decomposition}) can be simplified as 
\begin{equation} \label{eq:VarDecomposition}
\begin{split}
   \text{Var}(\vv{x}_0) =  &\E_{\Theta_{1:B}}[\text{Var}_{\vv{Z}|\Theta_{1:B}}(\hat{\mu}_{\ens}(\vv{x}_0;\vv{Z},\vv{P}(\Theta_{1:B})))]\\
    =& \frac{1}{B}\E_{\Theta}[V_1(\vv{x}_0;\Theta)] + 
    \frac{B-1}{B}\E_{\Theta,\Theta'}[V_2(\vv{x}_0;\Theta,\Theta')].
\end{split}    
\end{equation}

\subsection{Proof of Theorem \ref{new.thm1}} \label{new.SecA.2}

According to (\ref{eq:decomposition}), (\ref{eq:BiasDecomposition}), and (\ref{eq:VarDecomposition}), it holds that 
\begin{equation}\label{eq:thm1.0}
\begin{split}
    \text{MSE}(\hat{\mu}_{\ens};\vv{x}_0) &= \underbrace{(\mu(\vv{x}_0) - \E_\Theta(\tilde{\mu}(\vv{x}_0;\Theta))^2 + \frac{1}{B}\text{Var}_{\Theta} (\tilde{\mu}(\vv{x}_0;\Theta))}_{\text{Bias}^2(\vv{x}_0)}\\
    & + \underbrace{\frac{B-1}{B} \E_{\Theta,\Theta'}(V_2(\vv{x}_0;\Theta,\Theta')) + \frac{1}{B} \E_{\Theta}(V_1(\vv{x}_0;\Theta))}_{\text{Var}(\vv{x}_0)},
    \end{split}
\end{equation}
in which $\tilde{\mu}(\vv{x}_0;\Theta)$, $V_1(\vv{x}_0;\Theta)$, and $V_2(\vv{x}_0;\Theta,\Theta')$ are defined as in (\ref{eq:defntildemu}), (\ref{eq:defnV1}), and (\ref{eq:defnV2}), respectively. 
The proof of Theorem \ref{new.thm1} primarily involves deriving the asymptotic expansions for these three quantities. We omit the superscript $n$ from $\vv{P}^n$ whenever there is no ambiguity.

\subsubsection*{Asymptotic expansion of $\tilde{\mu}(\vv{x}_0;\Theta)$ in (\ref{eq:defntildemu})}
This section is devoted to showing the connection in (\ref{eq:1storderasymp}) of $\tilde{\mu}(\vv{x}_0;\Theta)$ and $\E_{\vv{X}}[\mu|\vv{P}_{\vv{x}_0}(\Theta)]$, the conditional expectation of the ground truth $\mu$ over the target cell $\vv{P}_{\vv{x}_0}(\Theta)$.

In view of the definitions in (\ref{defn:part0}) and (\ref{eq:defntildemu}), we have 
\begin{align*}
    \tilde{\mu}(\vv{x}_0;\Theta) &=\E_{\vv{Z}|\Theta}[\hat{\mu}_{\part}(\vv{x}_0;\vv{Z},\vv{P}(\Theta))]\\
    & = \E_{\vv{Z}|\Theta}\left[\frac{1}{N_{\vv{x}_0}(\Theta)}\sum_{i=1}^n Y_i I\{\vv{X}_i\in \vv{P}_{\vv{x}_0}(\Theta)\}\right]\\
    & = \E_{\vv{Z}|\Theta}\left[\frac{1}{N_{\vv{x}_0}(\Theta)}\sum_{i=1}^n (\mu(\vv{X}_i) + \epsilon_i) I\{\vv{X}_i\in \vv{P}_{\vv{x}_0}(\Theta)\}\right]\\
    &= \E_{\vv{Z}|\Theta}\left[\frac{1}{N_{\vv{x}_0}(\Theta)}\sum_{i=1}^n \mu(\vv{X}_i) I\{\vv{X}_i\in \vv{P}_{\vv{x}_0}(\Theta)\}\right],
\end{align*}
since $\E[\vv{\epsilon}|\vv{X}]=0$. 
Recall that the cell $\vv{P}_{\vv{x}_0}(\Theta)$ is independent of observations $\{\vv{X}_j\}$, and hence, $N_{\vv{x}_0}(\Theta) = \sum_{i=1}^n I\{\vv{X}_i \in \vv{P}_{\vv{x}_0}(\Theta)\}$ follows a binomial distribution $\mathcal{B}(n,p(\vv{x}_0;\Theta))$ with  $p(\vv{x}_0;\Theta)=\P_{\vv{X}|\Theta}(\vv{X}\in \vv{P}_{\vv{x}_0}(\Theta))$ when the partition is given. Thus, it holds that
\begin{align*}
    &\E_{\vv{Z}|\Theta}\left[\frac{1}{N_{\vv{x}_0}(\Theta)}\sum_{i=1}^n \mu(\vv{X}_i) I\{\vv{X}_i\in \vv{P}_{\vv{x}_0}(\Theta)\}\right] \\
    =&\E_{\vv{Z}|\Theta}\left[\sum_{k=1}^n \left(\frac{1}{k}\sum_{i=1}^n \mu(\vv{X}_i) I\{\vv{X}_i\in \vv{P}_{\vv{x}_0}(\Theta)\}\right) I\{N_{\vv{x}_0}(\Theta) = k\}\right]\\
    =&\sum_{k=1}^n \binom{n}{k}  \E_{\vv{Z}|\Theta}\left[\left(\frac{1}{k}\sum_{i=1}^k \mu(\vv{X}_i)\right)I\{\vv{X}_{1:k}\in \vv{P}_{\vv{x}_0}(\Theta);\vv{X}_{(k+1):n}\notin \vv{P}_{\vv{x}_0}(\Theta)\}\right]\\
    =& \E_{\vv{X}}[\mu|\vv{P}_{\vv{x}_0}(\Theta)] \sum_{k=1}^n \binom{n}{k} p(\vv{x}_0;\Theta)^k (1-p(\vv{x}_0;\Theta))^{n-k}\\
    =& \E_{\vv{X}}[\mu|\vv{P}_{\vv{x}_0}(\Theta)]\left( 1- (1-p(\vv{x}_0;\Theta))^n\right).
\end{align*}
In conclusion, we now find that 
\begin{equation}\label{eq:1storderasymp}
\begin{split}
    \tilde{\mu}(\vv{x}_0;\Theta) = \E_{\vv{X}}(\mu|\vv{P}_{\vv{x}_0}(\Theta))\left( 1- (1-p(\vv{x}_0;\Theta))^n\right)
    \end{split}
\end{equation}
with $p(\vv{x}_0;\Theta) = \P_{\vv{X}|\Theta}(\vv{X} \in \vv{P}_{\vv{x}_0}(\Theta))$.

Back to (\ref{eq:thm1.0}), it follows immediately from (\ref{eq:1storderasymp}) that 
\begin{equation}\label{eq:thm1.1}
    \begin{split}
        (\mu(\vv{x}_0) - \E_\Theta(\tilde{\mu}(\vv{x}_0;\Theta)))^2 & = (\mu(\vv{x}_0) - \E_\Theta(\E_{\vv{X}}(\mu|\vv{P}_{\vv{x}_0}(\Theta))))^2 \\
        & + \|\mu\|_\infty^2 O(\E_{\Theta}((1- p(\vv{x}_0;\Theta))^n)),\\
        \text{Var}_\Theta(\tilde{\mu}(\vv{x}_0;\Theta)) & = \text{Var}_\Theta(\E_{\vv{X}}(\mu|\vv{P}_{\vv{x}_0}(\Theta)))) \\
        & +  \|\mu\|_\infty^2 O(\E_{\Theta}((1- p(\vv{x}_0;\Theta))^n)),
    \end{split}
\end{equation}
which gives the asymptotic expansion for terms in the squared bias part.

\subsubsection*{Asymptotic expansion of variance $V_1(\vv{x}_0;\Theta)$ in (\ref{eq:defnV1})}

In this section, we focus on the nonasymptotic expansion of $V_1(\vv{x}_0;\Theta)$ by deriving its leading term and the order of remainders in (\ref{eq:V1asymp}) and (\ref{eq:RV1}).

Recall that
\begin{equation}\label{eq:varT}
\begin{split}
    V_1(\vv{x}_0;\Theta) & = \text{Var}_{\vv{Z}|\Theta}(\hat{\mu}_{\part}(\vv{x}_0;\vv{Z},\vv{P}(\Theta)))\\
     & = \E_{\vv{Z}|\Theta}[(\hat{\mu}_{\part}(\vv{x}_0;\vv{Z},\vv{P}(\Theta))^2] 
    - (\tilde{\mu}(\vv{x}_0;\Theta))^2,
\end{split}    
\end{equation}
in which the expansion of $\tilde{\mu}(\vv{x}_0;\Theta)$ has been derived in the last section. Hence, it suffices to focus on the first squared term.

Since $\E[\epsilon|\vv{X}] =0$ and noises $\epsilon_1,\ldots,\epsilon_n$ are mutually independent, we can deduce that 
\begin{align*}
    & \E_{\vv{Z}|\Theta}[(\hat{\mu}_{\part}(\vv{x}_0;\vv{Z},\vv{P}(\Theta))^2]\\
    = & \E_{\vv{Z}|\Theta}\left[\left(\frac{1}{N_{\vv{x}_0}(\Theta)}\sum_{i=1}^n (\mu(\vv{X}_i)+\epsilon_i) I\{\vv{X}_i\in \vv{P}_{\vv{x}_0}(\Theta)\}\right)^2\right]\\
    = & \E_{\vv{Z}|\Theta}\left[\frac{1}{N_{\vv{x}_0}(\Theta)^2}\sum_{i,j=1}^n (\mu(\vv{X}_i)+\epsilon_i)(\mu(\vv{X}_j)+\epsilon_j)I\{\vv{X}_i,\vv{X}_j\in \vv{P}_{\vv{x}_0}(\Theta)\}\right]\\
    = & \E_{\vv{Z}|\Theta}\left[\frac{1}{N_{\vv{x}_0}(\Theta)^2}\sum_{i=1}^n (\mu(\vv{X}_i)^2+\epsilon_i^2)I\{\vv{X}_i\in \vv{P}_{\vv{x}_0}(\Theta)\}\right]\\
    + & \E_{\vv{Z}|\Theta}\left[\frac{1}{N_{\vv{x}_0}(\Theta)^2}\sum_{1\leq i\neq j\leq n} \mu(\vv{X}_i)\mu(\vv{X}_j)I\{\vv{X}_i,\vv{X}_j\in \vv{P}_{\vv{x}_0}(\Theta)\}\right]\\
    = & \E_{\vv{Z}|\Theta}\left[\frac{1}{N_{\vv{x}_0}(\Theta)^2}\sum_{i=1}^n (\mu(\vv{X}_i)^2+\sigma^2(\vv{X}_i))I\{\vv{X}_i\in \vv{P}_{\vv{x}_0}(\Theta)\}\right]\\
    + & \E_{\vv{Z}|\Theta}\left[\frac{1}{N_{\vv{x}_0}(\Theta)^2}\sum_{1\leq i\neq j\leq n} \mu(\vv{X}_i)\mu(\vv{X}_j)I\{\vv{X}_i,\vv{X}_j\in \vv{P}_{\vv{x}_0}(\Theta)\}\right]\\
     =:& M_1 + M_2.
\end{align*}

Applying a similar technique as in the proof of (\ref{eq:1storderasymp}), 
we immediately find that the first term satisfies 
\begin{align*}
    M_1& = \E_{\vv{Z}|\Theta}\left[\frac{1}{N_{\vv{x}_0}(\Theta)^2}\sum_{i=1}^n (\mu(\vv{X}_i)^2+\sigma^2(\vv{X}_i))I\{\vv{X}_i\in \vv{P}_{\vv{x}_0}(\Theta)\}\right]\\
    =&\sum_{k=1}^n \binom{n}{k}
    \E_{\vv{Z}|\Theta}\left[\left(\frac{1}{k^2}\sum_{i=1}^k (\mu(\vv{X}_i)^2+\sigma^2(\vv{X}_i)\right)I\{\vv{X}_{1:k}\in \vv{P}_{\vv{x}_0}(\Theta);\vv{X}_{(k+1):n}\notin \vv{P}_{\vv{x}_0}(\Theta)\}\right]\\
    =&\E_{\vv{X}}[\mu^2 + \sigma^2| \vv{P}_{\vv{x}_0}(\Theta)] \sum_{k=1}^n \frac{1}{k}\binom{n}{k} p(\vv{x}_0;\Theta)^k(1-p(\vv{x}_0;\Theta))^{n-k}\\
    =&\E_{\vv{X}}[\mu^2 + \sigma^2| \vv{P}_{\vv{x}_0}(\Theta)]\E_{\vv{Z}|\Theta}\left[\frac{I\{N_{\vv{x}_0}(\Theta)\geq 1\}}{N_{\vv{x}_0}(\Theta)}\right],
\end{align*}
where $p(\vv{x}_0;\Theta) = \P_{\vv{X}|\Theta}(\vv{X} \in \vv{P}_{\vv{x}_0}(\Theta))$ and $N_{\vv{x}_0}(\Theta) \sim \mathcal{B}(n,p(\vv{x}_0;\Theta))$ given $\Theta$. 

Meanwhile, for the second term, we can also show that 
\begin{align*}
    M_2 &=\E_{\vv{Z}|\Theta}\left[\frac{1}{N_{\vv{x}_0}(\Theta)^2}\sum_{1\leq i\neq j\leq n} \mu(\vv{X}_i)\mu(\vv{X}_j)I\{\vv{X}_i,\vv{X}_j\in \vv{P}_{\vv{x}_0}(\Theta)\}\right]\\
    =&\sum_{k=1}^n \binom{n}{k} \frac{k(k-1)}{k^2} \E_{\vv{Z}|\Theta}\left[\mu(\vv{X}_1)\mu(\vv{X}_2)I\{\vv{X}_{1:k}\in \vv{P}_{\vv{x}_0}(\Theta);\vv{X}_{(k+1):n}\notin \vv{P}_{\vv{x}_0}(\Theta)\}\right]\\
    =&(\E_{\vv{X}}[\mu|\vv{P}_{\vv{x}_0}(\Theta)])^2\sum_{k=1}^n \frac{k-1}{k} \binom{n}{k}p(\vv{x}_0;\Theta)^k(1-p(\vv{x}_0;\Theta))^{n-k}\\
    =& (\E_{\vv{X}}[\mu|\vv{P}_{\vv{x}_0}(\Theta)])^2 \E_{\vv{Z}|\Theta}\left[\frac{(N_{\vv{x}_0}(\Theta)-1)I\{N_{\vv{x}_0}(\Theta)\geq 1\}}{N_{\vv{x}_0}(\Theta)}\right]\\
    =& (\E_{\vv{X}}[\mu|\vv{P}_{\vv{x}_0}(\Theta)])^2(1-(1-p(\vv{x}_0;\Theta))^n)\\
    - & (\E_{\vv{X}}[\mu|\vv{P}_{\vv{x}_0}(\Theta)])^2\E_{\vv{Z}|\Theta}\left[\frac{I\{N_{\vv{x}_0}(\Theta)\geq 1\}}{N_{\vv{x}_0}(\Theta)}\right].
\end{align*}
Substituting the results above into (\ref{eq:varT}), we can obtain that
\begin{equation}\label{eq:varV1}
    \begin{split}
    V_1(\vv{x}_0;\Theta) &=\left\{\text{Var}_{\vv{X}}[\mu|\vv{P}_{\vv{x}_0}(\Theta)] + \E_{\vv{X}}[\sigma^2|\vv{P}_{\vv{x}_0}(\Theta)]\right\} \E_{\vv{Z}|\Theta}\left[\frac{I\{N_{\vv{x}_0}(\Theta)\geq 1\}}{N_{\vv{x}_0}(\Theta)}\right]\\
    &+(\E_{\vv{X}}[\mu|\vv{P}_{\vv{x}_0}(\Theta)])^2
    (1-(1-p(\vv{x}_0;\Theta))^n) (1 - p(\vv{x}_0;\Theta))^n.
    \end{split}
\end{equation}

To further expand $V_1(\vv{x}_0;\Theta)$, we need to deal with term 
\[
\E_{\vv{Z}|\Theta}\left[\frac{I\{N_{\vv{x}_0}(\Theta)\geq 1\}}{N_{\vv{x}_0}(\Theta)}\right].
\]
The lemma below provides us with a \textit{sharp} approximation of this inverse moment of a binomial variable, and its proof is contained in Section \ref{new.SecB.1}. 

\begin{lemma}\label{keylem1}
Assume that $N \sim \mathcal{B}(n,p)$ with $p>0$. Then we have
\begin{equation*}
    \begin{split}
        \left|\E\left\{\frac{I\{N\geq 1\}}{N}\right\} - \frac{1}{np}\right| \leq C\left(1 + \frac{1}{np}\right) \frac{1}{(1 + (n-1)p)^2},
    \end{split}
\end{equation*}
in which $C$ is a positive constant independent of $n$ and $p$.
\end{lemma}

Combining the discussion above and the definition in (\ref{eq:localVar}), we conclude that
\begin{equation}\label{eq:V1asymp}
    \begin{split}
        V_1(\vv{x}_0;\Theta) &=\frac{\text{Var}_{\vv{X}}[\mu|\vv{P}_{\vv{x}_0}(\Theta)] + \E_{\vv{X}}[\sigma^2|\vv{P}_{\vv{x}_0}(\Theta)]}{n p(\vv{x}_0;\Theta)} + \mathcal{R}_{V1}(\vv{x}_0;\Theta) \\
        & = \frac{1}{n}\text{Var}_{\mu}(\vv{P}(\Theta);\vv{x}_0) + \frac{1}{n}\text{Var}_{\sigma^2}(\vv{P}(\Theta);\vv{x}_0) + \mathcal{R}_{V_1}(\vv{x}_0;\Theta),
    \end{split}
\end{equation}
where the remainder satisfies
\begin{equation}\label{eq:RV1}
\begin{split}
     \mathcal{R}_{V1}(\vv{x}_0;\Theta)& \lesssim (\|\mu\|_{\infty} + \|\sigma^2\|_{\infty})\\
     & \times \left(\frac{1}{(1 +(n-1)p(\vv{x}_0;\Theta))^2} + (1 - p(\vv{x}_0;\Theta))^n\right)\left(1 + \frac{1}{np(\vv{x}_0;\Theta)}\right)
    \end{split}
\end{equation}
with the positive constant in $\lesssim$ independent of $\mu$, $\sigma^2$, $\vv{P}$, $\Theta$, and $n$.

Back to (\ref{eq:thm1.0}), we can immediately deduce that 
\begin{equation}\label{eq:thm1.2}
    \begin{split}
        \E_\Theta(V_1(\vv{x}_0;\Theta)) & =
        \frac{1}{n}\E_{\Theta}\left\{\text{Var}_{\mu}(\vv{P}(\Theta);\vv{x}_0) + \text{Var}_{\sigma^2}(\vv{P}(\Theta);\vv{x}_0)\right\}\\
        & + (\|\mu\|_\infty + \|\sigma^2\|_\infty)^2 \\
        & \times O\Bigg(\E_\Theta\left(\frac{1}{np(\vv{x}_0;\Theta)(1 + (n-1)p(\vv{x}_0;\Theta))}\right)\\
        & ~~~~~~+ \E_\Theta((1 - p(\vv{x}_0;\Theta))^n)\Bigg).
    \end{split}
\end{equation}

\subsubsection*{Asymptotic expansion of covariance $V_2(\vv{x}_0;\Theta,\Theta')$ in (\ref{eq:defnV2})}

Recall that $V_2(\vv{x}_;\Theta,\Theta')$ is a covariance function and thus possesses the decomposition 
\begin{equation}\label{eq:covdecomp}
    \begin{split}
   V_2(\vv{x}_0;\Theta,\Theta') = & \text{Cov}_{\vv{Z}|\Theta,\Theta'}[\hat{\mu}_{\part}(\vv{x}_0;\vv{Z},\vv{P}(\Theta)), \hat{\mu}_{\part}(\vv{x}_0;\vv{Z},\vv{P}(\Theta'))]\\
    = & \E_{\vv{Z}|\Theta,\Theta'}[\hat{\mu}_{\part}(\vv{x}_0;\vv{Z},\vv{P}(\Theta))\hat{\mu}_{\part}(\vv{x}_0;\vv{Z},\vv{P}(\Theta'))]\\ 
    - & \tilde{\mu}(\vv{x}_0;\Theta) \tilde{\mu}(\vv{x}_0;\Theta').
    \end{split}
\end{equation}
Since we have already derived in (\ref{eq:1storderasymp}) the nonasymptotic expansion of $\tilde{\mu}(\vv{x}_0;\Theta)$ in the second term above for the product of expectations, it suffices to focus on the first term for the expectation of products. To this end, observe that 
\begin{align*}
    & \E_{\vv{Z}|\Theta,\Theta'}[\hat{\mu}_{\part}(\vv{x}_0;\vv{Z},\vv{P}(\Theta))\hat{\mu}_{\part}(\vv{x}_0;\vv{Z},\vv{P}(\Theta'))]\\
    = & \E_{\vv{Z}|\Theta,\Theta'}
    \left[\frac{1}{N_{\vv{x}_0}(\Theta)N_{\vv{x}_0}(\Theta')}\sum_{i,j=1}^n Y_iY_j I\{\vv{X}_i\in \vv{P}_{\vv{x}_0}(\Theta),\vv{X}_j\in \vv{P}_{\vv{x}_0}(\Theta')\}\right]\\
    = & \E_{\vv{Z}|\Theta,\Theta'}\left[
    \frac{1}{N_{\vv{x}_0}(\Theta)N_{\vv{x}_0}(\Theta')}
    \sum_{i=1}^n Y_i^2 
    I\{\vv{X}_i\in \vv{P}_{\vv{x}_0}(\Theta)\cap \vv{P}_{\vv{x}_0}(\Theta')\}
    \right]\\
    + & \E_{\vv{Z}|\Theta,\Theta'}\left[
    \frac{1}{N_{\vv{x}_0}(\Theta)N_{\vv{x}_0}(\Theta')}
    \sum_{1\leq i\neq j \leq n} Y_iY_j 
    I\{\vv{X}_i\in \vv{P}_{\vv{x}_0}(\Theta),\vv{X}_j\in \vv{P}_{\vv{x}_0}(\Theta')\}
    \right].
\end{align*}

To further elucidate the structure of the equations, we introduce some additional notation for simplicity. 
Without ambiguity, 
let us denote $\vv{P}_{\vv{x}_0}(\Theta)$ and $N_{\vv{x}_0}(\Theta)$ by $P$ and $N$, respectively, 
and likewise, $\vv{P}_{\vv{x}_0}(\Theta')$ and $N_{\vv{x}_0}(\Theta')$ by $P'$ and $N'$, respectively. Next, define $P_0 = P \cap P'$, $P_1 = P \setminus P'$, and $P_2 = P' \setminus P$. For $j = 0, 1, 2$, let $N_j = \sum_{i=1}^n I\{\vv{X}_i \in P_j\}$ be the number of observations in cell $P_j$. 
Clearly, given $\Theta$ and $\Theta'$, it holds that $N_j \sim \mathcal{B}(n, p_j)$, where $p_j = \P_{\vv{X}|\Theta,\Theta'}(\vv{X} \in P_j)$ for $j = 0, 1, 2$. Moreover, we notice that $N = N_0 + N_1$ and $N' = N_0 + N_2$.

The first term in $Y_i^2$'s can be further simplified as
\begin{equation}\label{eq:covterm1}
    \begin{split}
    &\E_{\vv{Z}|\Theta,\Theta'}\left[\frac{1}{N N'}\sum_{i=1}^n Y_i^2 I\{\vv{X}_i\in P_0 \}\right]\\
    =&\E_{\vv{Z}|\Theta,\Theta'}\left[\frac{1}{N N'}\sum_{i=1}^n (\mu(\vv{X}_i)^2+\epsilon_i^2) I\{\vv{X}_i\in P_0\}\right]\\
    =&\E_{\vv{Z}|\Theta,\Theta'}\left[\frac{1}{N N'}\sum_{i=1}^n (\mu(\vv{X}_i)^2+\sigma^2(\vv{X}_i)) I\{\vv{X}_i\in P_0\}\right]\\
    =&\E_{\vv{Z}|\Theta,\Theta'}\left[\frac{1}{(N_0+N_1)(N_0+N_2)}\sum_{i=1}^n (\mu(\vv{X}_i)^2+\sigma^2(\vv{X}_i)) I\{\vv{X}_i\in P_0\}\right]\\
    =&\E_{\vv{X}}[\mu^2+\sigma^2|P_0] \E_{\vv{Z}|\Theta,\Theta'}\left[\frac{N_0}{(N_0+N_1)(N_0+N_2)}\right].
    \end{split}
\end{equation}

The second term in the cross-product $Y_iY_j$'s can be further decomposed according to the different configurations of the two distinct observations $\vv{X}_i$ and $\vv{X}_j$ across the disjoint parts $P_0$, $P_1$, and $P_2$ 
\begin{align*}
    &\E_{\vv{Z}|\Theta,\Theta'}\left[\frac{1}{NN'}\sum_{i\neq j} Y_iY_j I\{\vv{X}_i\in P,\vv{X}_j\in P'\}\right]\\
    = & \E_{\vv{Z}|\Theta,\Theta'}\left[\frac{1}{N N'}\sum_{i\neq j} \mu(\vv{X}_i)\mu(\vv{X}_j) I\{\vv{X}_i\in P,\vv{X}_j\in P'\}\right]\\
    =&\E_{\vv{Z}|\Theta,\Theta'}\left[\frac{1}{N N'}\sum_{i\neq j} \mu(\vv{X}_i)\mu(\vv{X}_j) I\{\vv{X}_i, \vv{X}_j\in P_0\}\right]\\
    +&\E_{\vv{Z}|\Theta,\Theta'}\left[\frac{1}{N N'}\sum_{i\neq j} \mu(\vv{X}_i)\mu(\vv{X}_j) I\{\vv{X}_i\in P_0,\vv{X}_j\in P_2\}\right]\\
    +&\E_{\vv{Z}|\Theta,\Theta'}\left[\frac{1}{N N'}\sum_{i\neq j} \mu(\vv{X}_i)\mu(\vv{X}_j) I\{\vv{X}_i\in P_1,\vv{X}_j\in P_0\}\right]\\
    +&\E_{\vv{Z}|\Theta,\Theta'}\left[\frac{1}{N N'}\sum_{i\neq j} \mu(\vv{X}_i)\mu(\vv{X}_j) I\{\vv{X}_i\in P_1,\vv{X}_j\in P_2\}\right]\\
    =: & I_1 + I_2 + I_3 + I_4,
\end{align*}
in which
\begin{equation*}
    \begin{split}
    I_1 & = \E_{\vv{Z}|\Theta,\Theta'}\left[\frac{1}{N N'}\sum_{i\neq j} \mu(\vv{X}_i)\mu(\vv{X}_j) I\{\vv{X}_i, \vv{X}_j\in P_0\}\right]\\
    &=(\E_{\vv{X}}[\mu|P_0])^2\E_{\vv{Z}|\Theta,\Theta'}\left[\frac{N_0(N_0-1)}{(N_0+N_1)(N_0+N_2)}\right] ,\\
    I_2 & = \E_{\vv{Z}|\Theta,\Theta'}\left[\frac{1}{N N'}\sum_{i\neq j} \mu(\vv{X}_i)\mu(\vv{X}_j) I\{\vv{X}_i \in P_0, \vv{X}_j\in P_2\}\right]\\
    &=(\E_{\vv{X}}[\mu|P_0])(\E_{\vv{X}}[\mu|P_2])\E_{\vv{Z}|\Theta,\Theta'}\left[\frac{N_0N_2}{(N_0+N_1)(N_0+N_2)}\right],\\
    I_3 & = \E_{\vv{Z}|\Theta,\Theta'}\left[\frac{1}{N N'}\sum_{i\neq j} \mu(\vv{X}_i)\mu(\vv{X}_j) I\{\vv{X}_i \in P_1, \vv{X}_j\in P_0\}\right]\\
    &=(\E_{\vv{X}}[\mu|P_1])(\E_{\vv{X}}[\mu|P_0])\E_{\vv{Z}|\Theta,\Theta'}\left[\frac{N_1N_0}{(N_0+N_1)(N_0+N_2)}\right],\\
    I_4 & = \E_{\vv{Z}|\Theta,\Theta'}\left[\frac{1}{N N'}\sum_{i\neq j} \mu(\vv{X}_i)\mu(\vv{X}_j) I\{\vv{X}_i \in P_1, \vv{X}_j\in P_2\}\right]\\
    &=(\E_{\vv{X}}[\mu|P_1])(\E_{\vv{X}}[\mu|P_2])\E_{\vv{Z}|\Theta,\Theta'}\left[\frac{N_1N_2}{(N_0+N_1)(N_0+N_2)}\right].
    \end{split}
\end{equation*}

Combining the decompositions above, we can deduce that  
\begin{equation}\label{eq:cov0}
    \begin{split}
        &\E_{\vv{Z}|\Theta,\Theta'}[\hat{\mu}_{\part}(\vv{x}_0;\vv{Z},\vv{P}(\Theta))\hat{\mu}_{\part}(\vv{x}_0;\vv{Z},\vv{P}(\Theta'))]\\
        =&(\E_{\vv{X}}(\sigma^2|P_0) + \text{Var}_{\vv{X}}(\mu|P_0)) \E_{\vv{Z}|\Theta,\Theta'}\left[\frac{N_0}{(N_0+N_1)(N_0+N_2)}\right]\\
        +&\E_{\vv{Z}|\Theta,\Theta'}\Bigg[ \left(
        \E_{\vv{X}}(\mu|P_0) \frac{N_0}{N_0 + N_1} + \E_{\vv{X}}(\mu|P_1) \frac{N_1}{N_0 + N_1}\right) \\
        & ~~~~~~~~~\times \left(\E_{\vv{X}}(\mu|P_0) \frac{N_0}{N_0 + N_2} + \E_{\vv{X}}(\mu|P_2) \frac{N_2}{N_0 + N_2}\right)\Bigg].
    \end{split}
\end{equation}
Consequently, an application of (\ref{eq:1storderasymp}), (\ref{eq:covdecomp}), and (\ref{eq:cov0}) leads to the covariance formula 
\begin{equation}\label{eq:cov}
    \begin{split}
        V_2(\vv{x}_0;\Theta,\Theta') = & \text{Cov}_{\vv{Z}|\Theta,\Theta'}[\hat{\mu}_{\part}(\vv{x}_0;\vv{Z},\vv{P}(\Theta)), \hat{\mu}_{\part}(\vv{x}_0;\vv{Z},\vv{P}(\Theta'))]\\
        = & (\E_{\vv{X}}(\sigma^2|P_0) + \text{Var}_{\vv{X}}(\mu|P_0)) \E_{\vv{Z}|\Theta,\Theta'}\left[\frac{N_0}{(N_0+N_1)(N_0+N_2)}\right]\\
        +&\E_{\vv{Z}|\Theta,\Theta'}\Bigg[ \left(
        \E_{\vv{X}}(\mu|P_0) \frac{N_0}{N_0 + N_1} + \E_{\vv{X}}(\mu|P_1) \frac{N_1}{N_0 + N_1}\right) \\
        & ~~~~~~~~~\times \left(\E_{\vv{X}}(\mu|P_0) \frac{N_0}{N_0 + N_2} + \E_{\vv{X}}(\mu|P_2) \frac{N_2}{N_0 + N_2}\right)\Bigg]\\
        - & (\E_{\vv{X}}(\mu|P))(\E_{\vv{X}}(\mu|P'))\left( 1- (1-p)^n\right)\left( 1- (1-p')^n\right),
    \end{split}
\end{equation}
where $p=\P_{\vv{X}|\Theta}(\vv{X}\in P)$ and $p'=\P_{\vv{X}|\Theta'}(\vv{X}\in P')$.  

\begin{lemma}\label{keylem2}
    Assume that we have an i.i.d. sample $\vv{Z} = \{\vv{X}_i\}_{1\leq i\leq n}$ from a population $\vv{X}$ in $\mathcal{X}^d$.
    Let $P$ and $P'$ be two non-empty cells in $\mathcal{X}^d$ with non-empty intersection $P_0 = P\cap P' \neq \varnothing$.
    Denote by 
    $N = \sum_{i=1}^n I\{\vv{X}_i \in P\}$ with $p = \P(\vv{X} \in P)$, and similarly, $N'$ and $p'$ for $P'$.
    In addition, define $P_1 = P \setminus P'$ and $P_2 = P' \setminus P$. Let $N_j = \sum_{i=1}^n I\{\vv{X}_i \in P_j\}$ and $p_j = \P(\vv{X} \in P_j)$. Then when $p_0 >0$, it holds that 
    \begin{itemize}
        \item[(1)] 
    
    \begin{equation}\label{lemma2.1}
            \begin{split}
            \E\left[\frac{N_0}{N N'}\right] =  \frac{p_0}{n p p'} + R_{21},
            \end{split}
        \end{equation}
        where the remainder satisfies
        \[
        R_{21} \lesssim \frac{p_0}{npp'}\left(\frac{1}{1 + (n-1)p} + \frac{1}{1 + (n-1)p'}\right)
        \]
        with the constant $C$ in $\lesssim$ independent of cells $P$, $P'$, and the population distribution of $\vv{X}$.
        
        \item[(2)] For any real numbers $\alpha$, $\beta$, and $\gamma$, we have 
        \begin{equation}\label{lemma2.2}
            \begin{split}
             &\E\left[\left(\alpha \frac{N_0}{N} + \beta \frac{N_1}{N}\right) \left(\alpha \frac{N_0}{N'} + \gamma \frac{N_1}{N'}\right) \right] \\
             =
             & \left(\alpha \frac{p_0}{p} + \beta \frac{p_1}{p}\right)
             \left(\alpha \frac{p_0}{p'} + \gamma \frac{p_2}{p'}\right)\\
             +& (\alpha - \beta)(\alpha - \gamma) \left(\frac{p_1p_2}{pp'}\right)\left(\frac{p_0}{n p p'}\right) + R_{22},
            \end{split}
        \end{equation}
        where the remainder $R_{22}$ satisfies
        \begin{align*}
            R_{22} \lesssim  \max\{|\alpha|,|\beta|,|\gamma|\}^2 &\Bigg\{\frac{p_0}{npp'}\left\{\frac{1}{1 + (n-1)p} + \frac{1}{1+ (n-1)p'}\right\}\\
            &+ \frac{1}{(1 + (n-1)p)^{3/2}} + \frac{1}{(1 + (n-1)p')^{3/2}}\\
    &+(1 - p)^n + (1 - p')^n\Bigg\}.
        \end{align*}
    \end{itemize} 
\end{lemma}

The proof of Lemma \ref{keylem2} above is provided in Section \ref{new.SecB.2}. 
Substituting (\ref{lemma2.1}) and (\ref{lemma2.2}) into (\ref{eq:cov}), by setting 
\[
\alpha = \E_{\vv{X}}(\mu|P_0),~~~~\beta = \E_{\vv{X}}(\mu|P_1),~~~~\gamma = \E_{\vv{X}}(\mu|P_3),
\]
we can deduce that 
\begin{align*}
    V_2(\vv{x}_0;\Theta,\Theta') & =  (\E_{\vv{X}}(\sigma^2|P_0) + \text{Var}_{\vv{X}}(\mu|P_0))  \frac{p_0}{n p p'}\\
        & + (\E_{\vv{X}}(\mu|P_0) - \E_{\vv{X}}(\mu|P_1))(\E_{\vv{X}}(\mu|P_0) - \E_{\vv{X}}(\mu|P_2)) \frac{p_1p_2}{pp'} \frac{p_0}{npp'}\\
        & + (\|\mu\|_\infty + \|\sigma^2\|_\infty)^2\\ & \times O\Bigg(\frac{p_0}{npp'}\left\{\frac{1}{1 + (n-1)p} + \frac{1}{1+ (n-1)p'}\right\}\\
        &~~~~~+ \frac{1}{(1 + (n-1)p)^{3/2}} + \frac{1}{(1 + (n-1)p')^{3/2}}\\
    &~~~~~+(1 - p)^n - (1 - p')^n\Bigg).
\end{align*}
Note that 
\begin{equation}\label{eq:keydecompose}
\begin{split}
    &\E_{\vv{X}}((\mu - \E_{\vv{X}}(\mu|P))(\mu - \E_{\vv{X}}(\mu|P'))| P_0) \\
    = &\text{Var}_{\vv{X}}(\mu|P_0) 
    + (\E_{\vv{X}}(\mu|P_0) - \E_{\vv{X}}(\mu|P))(\E_{\vv{X}}(\mu|P_0) - \E_{\vv{X}}(\mu|P'))
\end{split}
\end{equation}
and
\begin{align*}
    \E_{\vv{X}}(\mu|P) = \E_{\vv{X}}(\mu|P_0) \frac{p_0}{p} + \E_{\vv{X}}(\mu|P_1) \frac{p_1}{p},\\
    \E_{\vv{X}}(\mu|P') = \E_{\vv{X}}(\mu|P_0) \frac{p_0}{p'} + \E_{\vv{X}}(\mu|P_2) \frac{p_2}{p'}.
\end{align*}

We can then obtain the decomposition formula 
\begin{align*}
&\E_{\vv{X}}((\mu - \E_{\vv{X}}(\mu|P))(\mu - \E_{\vv{X}}(\mu|P'))| P_0) \\
&= {\rm Var}_{\vv{X}}(\mu|P_0)\\
& + (\E_{\vv{X}}(\mu|P_0) - \E_{\vv{X}}(\mu|P_1))(\E_{\vv{X}}(\mu|P_0) - \E_{\vv{X}}(\mu|P_2)) \frac{p_1p_2}{pp'}, 
\end{align*}
which along with (\ref{eq:localCov}) entails that
\begin{align*}
    V_2(\vv{x}_0;\Theta,\Theta') & =  \E_{\vv{X}}(\sigma^2|P_0)\frac{p_0}{n p p'}  +  \E_{\vv{X}}((\mu - \E_{\vv{X}}(\mu|P))(\mu - \E_{\vv{X}}(\mu|P'))| P_0) \frac{p_0}{npp'}\\
        &  + (\|\mu\|_\infty + \|\sigma^2\|_\infty)^2\\ & \times O\Bigg(\frac{p_0}{npp'}\left\{\frac{1}{1 + (n-1)p} + \frac{1}{1+ (n-1)p'}\right\}\\
        &~~~~~+ \frac{1}{(1 + (n-1)p)^{3/2}} + \frac{1}{(1 + (n-1)p')^{3/2}}\\
    &~~~~~+(1 - p)^n + (1 - p')^n\Bigg)\\
        & = \frac{1}{n} \left\{\text{Cov}_{\mu}(\vv{P}(\Theta),\vv{P}(\Theta');\vv{x}_0)
        + \text{Cov}_{\sigma^2}(\vv{P}(\Theta),\vv{P}(\Theta');\vv{x}_0)\right\} \\
        & + (\|\mu\|_\infty + \|\sigma^2\|_\infty)^2\\ & \times O\Bigg(\frac{p_0}{npp'}\left\{\frac{1}{1 + (n-1)p} + \frac{1}{1+ (n-1)p'}\right\}\\
        &~~~~~+ \frac{1}{(1 + (n-1)p)^{3/2}} + \frac{1}{(1 + (n-1)p')^{3/2}}\\
    &~~~~~+(1 - p)^n + (1 - p')^n\Bigg).
\end{align*}

Finally, by integrating with respect to the exogenous factors $\Theta$ and $\Theta'$, we can conclude that
\begin{equation}\label{eq:thm1.3}
    \begin{split}
        \E_{\Theta,\Theta'}(V_2(\vv{x}_0;\Theta,\Theta')) & = \frac{1}{n} \E_{\Theta,\Theta'}\Big\{\text{Cov}_{\mu}(\vv{P}(\Theta),\vv{P}(\Theta');\vv{x}_0)\\
        &~~~~~~~~~~~~~
        + \text{Cov}_{\sigma^2}(\vv{P}(\Theta),\vv{P}(\Theta');\vv{x}_0)\Big\} \\
        & + (\|\mu\|_\infty + \|\sigma^2\|_\infty)^2\\ & \times O\Bigg(\E_{\Theta,\Theta'}\left(\frac{p(\vv{x}_0;\Theta,\Theta')}{n p(\vv{x}_0;\Theta) p(\vv{x}_0;\Theta')(1 + (n-1)p(\vv{x}_0;\Theta))}\right)\\
        & ~~~~~~~+ \E_{\Theta}\left(\frac{1}{(1+(n-1)p(\vv{x}_0;\Theta))^{3/2}}\right)\\
        & ~~~~~~~+ \E_{\Theta}((1 - p(\vv{x}_0;\Theta))^n) \Bigg)
    \end{split}
\end{equation}
with $p(\vv{x}_0;\Theta,\Theta') = \P_{\vv{X}}(\vv{P}_{\vv{x}_0}(\Theta) \cap \vv{P}_{\vv{x}_0}(\Theta'))$. 

The bound in (\ref{eq:localMSEens}) 
can be derived directly by a combination of (\ref{eq:thm1.0}), (\ref{eq:thm1.1}), (\ref{eq:thm1.2}), and (\ref{eq:thm1.3}). This completes the proof of Theorem \ref{new.thm1}.

\subsection{Proof of Proposition \ref{prop:localtrieq}}\label{new.SecA.prop1}

Observe that for each $P_i\in P$, it holds that $\mu_P I_{P_i} = \E_{\vv{X}}(\mu | P_i)I_{P_i}$, and similarly, for each $P' \in P'$, it holds that $\mu_{P'} I_{P'_j} = \E_{\vv{X}}(\mu | P'_j)I_{P'_j}$. Thus it follows that 
\[
(\mu - \mu_P) I_{P_{\vv{x}_0}} = (\mu - \E_{\vv{X}}(\mu|P_{\vv{x}_0}))I_{P_{\vv{x}_0}},~~~(\mu - \mu_{P'}) I_{P'_{\vv{x}_0}} = (\mu - \E_{\vv{X}}(\mu|P'_{\vv{x}_0}))I_{P'_{\vv{x}_0}}.
\]
Meanwhile, by the Cauchy--Schwarz inequality, we have
\[
  \E_{\vv{X}}((\mu - \mu_P)(\mu - \mu_{P'}) I_{P_{\vv{x}_0}}I_{P'_{\vv{x}_0}}) \leq \sqrt{\E_{\vv{X}}((\mu - \mu_P)^2 I_{P_{\vv{x}_0}})\E_{\vv{X}}((\mu - \mu_{P'})^2 I_{P'_{\vv{x}_0}})}.
\]
     
By the definition of the signal-induced local cross-partition covariance function in (\ref{eq:localCov}), it holds that 
\begin{align*}
    \text{Cov}_{\mu}(P,P';\vv{x}_0) &= \frac{\E_{\vv{X}}((\mu - \mu_P)(\mu - \mu_{P'}) I_{P_{\vv{x}_0}\cap P'_{\vv{x}_0}})}{\P_{\vv{X}}(P_{\vv{x}_0})\P_{\vv{X}}(P'_{\vv{x}_0})}  = \frac{\E_{\vv{X}}((\mu - \mu_P)(\mu - \mu_{P'}) I_{P_{\vv{x}_0}}I_{P'_{\vv{x}_0}})}{\P_{\vv{X}}(P_{\vv{x}_0})\P_{\vv{X}}(P'_{\vv{x}_0})}\\
    & \leq \sqrt{\frac{\E_{\vv{X}}((\mu - \mu_P)^2|P_{\vv{x}_0})\E_{\vv{X}}((\mu - \mu_{P'})^2|P_{\vv{x}_0})}{\P_{\vv{X}}(P_{\vv{x}_0})\P_{\vv{X}}(P'_{\vv{x}_0})}}\\
    & = \sqrt{\text{Var}_{\mu}(P;\vv{x}_0)\text{Var}_{\mu}(P';\vv{x}_0)}.
\end{align*}
The equality above holds if and only if $(\mu - \mu_P) I_{P_{\vv{x}_0}} = (\mu - \mu_{P'}) I_{P'_{\vv{x}_0}}$ almost surely under $\P_{\vv{X}}$. We next discuss the sufficient and necessary condition for obtaining such equality. 

On the one hand, since $\P_{\vv{X}}(P_{\vv{x}_0}\cap P'_{\vv{x}_0}) >0$, by Assumption \ref{assm:4} we immediately see that $\E_{\vv{X}}(\mu|P_{\vv{x}_0}) = \E_{\vv{X}}(\mu|P'_{\vv{x}_0})$ because $\mu - \mu_P$ and $\mu - \mu_{P'}$ are identical on $P_{\vv{x}_0}\cap P'_{\vv{x}_0}$. 
On the other hand, the equation also implies that $\mu - \E(\mu|P_{\vv{x}_0}) = 0$ on $P_{\vv{x}_0}\backslash P'_{\vv{x}_0}$ and  $P'_{\vv{x}_0}\backslash P_{\vv{x}_0}$. When $P_{\vv{x}_0}$ and $P'_{\vv{x}_0}$ are indistinguishable, $\mu$ must be a constant on $P_{\vv{x}_0} \cup P_{\vv{x}_0}$.

We now move on to the second part of the proof. Note that by the Cauchy--Schwarz inequality,
\[
  \E_{\vv{X}}(\sigma^2 I_{P_{\vv{x}_0}}I_{P'_{\vv{x}_0}}) \leq \sqrt{\E_{\vv{X}}(\sigma^2 I_{P_{\vv{x}_0}})\E_{\vv{X}}(\sigma^2 I_{P'_{\vv{x}_0}})}.
\]
Then it follows from the definition of the model-error-induced local cross-partition covariance function in (\ref{eq:localCov}) that 
\begin{align*}
    \text{Cov}_{\sigma^2}(P,P';\vv{x}_0) &= \frac{\E_{\vv{X}}(\sigma^2 I_{P_{\vv{x}_0}\cap P'_{\vv{x}_0}})}{\P_{\vv{X}}(P_{\vv{x}_0})\P_{\vv{X}}(P'_{\vv{x}_0})}  = \frac{\E_{\vv{X}}(\sigma^2 I_{P_{\vv{x}_0}}I_{P'_{\vv{x}_0}})}{\P_{\vv{X}}(P_{\vv{x}_0})\P_{\vv{X}}(P'_{\vv{x}_0})}\\
    & \leq \sqrt{\frac{\E_{\vv{X}}(\sigma^2|P_{\vv{x}_0})\E_{\vv{X}}(\sigma^2|P_{\vv{x}_0})}{\P_{\vv{X}}(P_{\vv{x}_0})\P_{\vv{X}}(P'_{\vv{x}_0})}}\\
    & = \sqrt{\text{Var}_{\sigma^2}(P;\vv{x}_0)\text{Var}_{\sigma^2}(P';\vv{x}_0)}.
\end{align*}
The equality above holds if and only if $\sigma^2 I_{P_{\vv{x}_0}} = \sigma^2 I_{P'_{\vv{x}_0}}$ almost surely under $\P_{\vv{X}}$, which is equivalent to the indistinguishability of $P_{\vv{x}_0}$ and $P'_{\vv{x}_0}$. This concludes the proof of Proposition \ref{prop:localtrieq}.

\section{Proofs of Theorems \ref{new.thm.binary}--\ref{new.thm3} and Proposition \ref{prop:globaltrieq}} \label{new.SecA}

\subsection{Proof of Theorem \ref{new.thm.binary}}\label{new.example1.derive}

For a given target point $\vv{x}_0$, the terminal node $\mathbf{t}_{\vv{x}_0} = \mathbf{t}(\vv{x}_0;I_l)$ containing $\vv{x}_0$ can be uniquely determined by $\vv{x}_0$ and a binary CART process, where the process is completely independent of $\vv{x}_0$. The independence of the binary CART process and $\vv{x}_0$ enables us to exchange the order of integration with respect to the binary CART process and the testing data $\vv{X}'$ replacing $\vv{x}_0$.

\subsubsection*{Derivation of the local MSE formula}
We first determine the local MSE for the related random forests estimator $\hat{\mu}_{RF}$ according to our general Theorem \ref{new.thm1}. 
We begin with the squared bias terms in (\ref{eq:localMSEens}).
Observe that in (\ref{eq:localMSEens}), we have 
\begin{equation}\label{eq:biasdecompose}
\begin{split}
&\left(\mu(\vv{x}_0) - \E_{\Theta}\left(\E_{\vv{X}}(\mu|\vv{P}_{\vv{x}_0}(\Theta))\right)\right)^2\\
=& \left(\mu(\vv{x}_0) - \E_{\Theta}\left(\E_{\vv{X}}(\mu|\vv{P}_{\vv{x}_0}(\Theta))\right)\right) \left(\mu(\vv{x}_0) - \E_{\Theta'}\left(\E_{\vv{X}}(\mu|\vv{P}_{\vv{x}_0}(\Theta'))\right)\right) \\
=& \E_{\Theta,\Theta'}\left[\left(\mu(\vv{x}_0) - \E_{\vv{X}}(\mu|\vv{P}_{\vv{x}_0}(\Theta))\right)\left(\mu(\vv{x}_0) - \E_{\vv{X}}(\mu|\vv{P}_{\vv{x}_0}(\Theta'))\right)\right],
\end{split}
\end{equation}
where the exogenous factors $\Theta$ and $\Theta'$ are mutually independent. 
Replacing $\vv{P}(\Theta)$ and $\vv{P}(\Theta')$ by two independent bi-partitions with the population CART driven by two independent binary CART processes $I_l$ and $I'_l$,
we see that the squared bias part in the local MSE formula is given by
\begin{align*}
    &\text{The squared bias of tree ensemble} \\
    =& \E_{I_l,I'_l}\left[(\mu(\vv{x}_0) - \E_{\vv{X}}(\mu | \mathbf{t}(\vv{x}_0;I_l))) (\mu(\vv{x}_0) - \E_{\vv{X}}(\mu | \mathbf{t}(x_0;I'_l))) \right].
\end{align*}

Recall that for points in $\mathbf{t}(\vv{x}_0;I_l)$, coordinates $x_{j}$ with $I_{lj} = 1$ are identical to $x_{0j}$, and the rest can take values $0$ or $1$. Hence, it holds that 
\begin{align*}
    \E_{\vv{X}}(\mu|\mathbf{t}(\vv{x}_0;I_l)) & = \sum_{j:1\leq j\leq s, I_{lj} = 1}  \mu(x_{0j}) + \sum_{j: 1\leq j\leq s, I_{lj} = 0} \beta_j \E(X_j)\\
    & = \sum_{j:1\leq j\leq s, I_{lj} = 1} \mu(x_{0j}) + \sum_{j: 1\leq j\leq s, I_{lj} = 0} \frac{\beta_j}{2}
\end{align*}
and consequently, 
\begin{align*}
    \mu(\vv{x}_0) - \E_{\vv{X}}(\mu | \mathbf{t}(x_0;I_l)) & = \sum_{j: 1\leq j\leq s, I_{lj} = 0} \beta_j \left(x_{0j} - \frac{1}{2}\right).
\end{align*}
It then follows that
\begin{align*}
    &\text{The squared bias of tree ensemble}\\
    & = \E_{I_l,I'_l}\left[\left(\sum_{j: 1\leq j\leq s, I_{lj} = 0}\beta_j\left(x_{0j} - \frac{1}{2}\right)\right)
    \left(\sum_{j': 1\leq j'\leq s, I'_{lj'} = 0}\beta_{j'}\left(x_{0j'} - \frac{1}{2}\right)\right)\right].
\end{align*}
Similarly, for one single tree, we can show that 
\begin{align*}
    &\text{The squared bias of a single tree} \\
     =& \E_{I_l}\left[(\mu(\vv{x}_0) - \E_{\vv{X}}(\mu;\mathbf{t}(\vv{x}_0;I_l)))^2\right]\\
     =& \E_{I_l}\left[\left(\sum_{j: 1\leq j\leq s, I_{lj} = 0} \beta_j^2 \left(x_{0j} - \frac{1}{2}\right)\right)^2
    \right].
\end{align*}

We next deal with the cross-partition covariance and the single-partition variance terms in (\ref{eq:localCov}). 
Note that a coordinate in the intersection of two terminal nodes is split as long as it is split by either $I_l$ or $I_l'$. Given $I_l$ and $I'_l$, we have
\begin{align*}
    &\P_{\vv{X}}(\mathbf{t}(\vv{x}_0;I_l) \cap \mathbf{t}(\vv{x}_0;I'_l)) = 2^{-\sum_{j=1}^d \max\{I_{lj},I'_{lj}\}},\\
    &\P_{\vv{X}}(\mathbf{t}(\vv{x}_0;I_l)) = 2^{-\sum_{j=1}^d I_{lj}} ,~~~\P_{\vv{X}}(\mathbf{t}(\vv{x}_0;I'_l)) = 2^{-\sum_{j=1}^d I'_{lj}} ,
\end{align*}
and thus,
\begin{align*}
    \frac{\P_{\vv{X}}(\mathbf{t}(\vv{x}_0;I_l) \cap \mathbf{t}(\vv{x}_0;I'_l))}{\P_{\vv{X}}(\mathbf{t}(\vv{x}_0;I_l))\P_{\vv{X}}(\mathbf{t}(\vv{x}_0;I'_l))} & = 2^{\sum_{j=1}^d I_{lj} + I'_{lj} - \max\{I_{lj},I'_{lj}\}} \\
    & = 2^{\sum_{j=1}^d \min\{I_{lj},I'_{lj}\}}.
\end{align*}
Since the model error has constant variance, in light of \eqref{eq:localCov}, for two independent bi-partitions $P$ and $P'$ driven by two independent binary CART processes $I_l$ and $I'_l$, the model-error induced cross-partition covariance is given by 
\[
\text{Cov}_{\sigma^2}(P,P') = \sigma_0^2 \frac{\P_{\vv{X}}(\mathbf{t}(\vv{x}_0;I_l) \cap \mathbf{t}(\vv{x}_0;I'_l))}{\P_{\vv{X}}(\mathbf{t}(\vv{x}_0;I_l))\P_{\vv{X}}(\mathbf{t}(\vv{x}_0;I'_l))} = \sigma_0^2 2^{\sum_{j=1}^d \min\{I_{lj},I'_{lj}\}}.
\]
The model-error-induced local partition variance function becomes
\[
\text{Var}_{\sigma^2}(P) = \sigma_0^2 \frac{1}{\P_{\vv{X}}(\mathbf{t}(\vv{x}_0,I_l))} = \sigma_0^2 2^{\sum_{j=1}^d I_{lj}}.
\]

Let us proceed with analyzing the signal-induced cross-partition covariance function. Recall the decomposition formula (\ref{eq:keydecompose}) that the coefficient in (\ref{eq:localCov}) satisfies
\begin{align*}
    &\E_{\vv{X}}\left[(\mu - \E_{\vv{X}}(\mu|\mathbf{t}(\vv{x}_0;I_l)))(\mu - \E_{\vv{X}}(\mu|\mathbf{t}(\vv{x}_0;I'_l)))\big| \mathbf{t}(\vv{x}_0;I_l) \cap \mathbf{t}(\vv{x}_0;I'_l)\right] \\
    =& \text{Var}_{\vv{X}}(\mu | \mathbf{t}(\vv{x}_0;I_l) \cap \mathbf{t}(\vv{x}_0;I'_l)) \\
    + &  (\E_{\vv{X}}(\mu|\mathbf{t}(\vv{x}_0;I_l)\cap \mathbf{t}(\vv{x}_0;I'_l)) - \E_{\vv{X}}(\mu|\mathbf{t}(\vv{x}_0;I_l)))\\
    \times&
    (\E_{\vv{X}}(\mu|\mathbf{t}(\vv{x}_0;I_l)\cap \mathbf{t}(\vv{x}_0;I'_l)) - \E_{\vv{X}}(\mu|\mathbf{t}(\vv{x}_0;I'_l))).
\end{align*}
Observe that
\begin{align*}
    \text{Var}_{\vv{X}}(\mu|\mathbf{t}(\vv{x}_0;I_l)\cap \mathbf{t}(\vv{x}_0;I'_l)) & = \frac{1}{4} \sum_{j:1\leq j\leq s,I_{lj} = I'_{lj} = 0} \beta_j^2 \\
    & = \frac{1}{4} \sum_{j=1}^s \beta_j^2 I\{\max\{I_{lj},I'_{lj}\} = 0\}
\end{align*}
and
\begin{align*}
    &(\E_{\vv{X}}(\mu|\mathbf{t}(\vv{x}_0;I_l)\cap \mathbf{t}(\vv{x}_0;I'_l)) - \E_{\vv{X}}(\mu|\mathbf{t}(\vv{x}_0;I_l)))\\
    &\times
    (\E_{\vv{X}}(\mu|\mathbf{t}(\vv{x}_0;I_l)\cap \mathbf{t}(\vv{x}_0;I'_l)) - \E_{\vv{X}}(\mu|\mathbf{t}(\vv{x}_0;I'_l)))\\
    =&\left(\sum_{j:1\leq j\leq s, I'_{lj} - I_{lj} =1} \beta_j\left(x_{0j} - \frac{1}{2}\right)\right)\left(\sum_{j':1\leq j'\leq s, I_{lj'} - I'_{lj'} =1} \beta_{j'}\left(x_{0j'} - \frac{1}{2}\right)\right).
\end{align*}

Hence, for two independent bi-partitions $P$ and $P'$ driven by $I_l$ and $I'_l$, we can deduce that 
\begin{align*}
    \text{Cov}_{\mu}(P,P) & = \E_{\vv{X}}\left[(\mu - \E_{\vv{X}}(\mu|\mathbf{t}(\vv{x}_0;I_l)))(\mu - \E_{\vv{X}}(\mu|\mathbf{t}(\vv{x}_0;I'_l)))\big| \mathbf{t}(\vv{x}_0;I_l) \cap \mathbf{t}(\vv{x}_0;I'_l)\right]\\
    & \times \frac{\P_{\vv{X}}(\mathbf{t}(\vv{x}_0;I_l) \cap \mathbf{t}(\vv{x}_0;I'_l))}{\P_{\vv{X}}(\mathbf{t}(\vv{x}_0;I_l))\P_{\vv{X}}(\mathbf{t}(\vv{x}_0;I'_l))}\\
    & = \text{Var}_{\vv{X}}(\mu|\mathbf{t}(\vv{x}_0;I_l)\cap \mathbf{t}(\vv{x}_0;I'_l))  \frac{\P_{\vv{X}}(\mathbf{t}(\vv{x}_0;I_l) \cap \mathbf{t}(\vv{x}_0;I'_l))}{\P_{\vv{X}}(\mathbf{t}(\vv{x}_0;I_l))\P_{\vv{X}}(\mathbf{t}(\vv{x}_0;I'_l))}\\
    & + (\E_{\vv{X}}(\mu|\mathbf{t}(\vv{x}_0;I_l)\cap \mathbf{t}(\vv{x}_0;I'_l)) - \E_{\vv{X}}(\mu|\mathbf{t}(\vv{x}_0;I_l)))\\
    &\times
    (\E_{\vv{X}}(\mu|\mathbf{t}(\vv{x}_0;I_l)\cap \mathbf{t}(\vv{x}_0;I'_l)) - \E_{\vv{X}}(\mu|\mathbf{t}(\vv{x}_0;I'_l)))\\
    &\times \frac{\P_{\vv{X}}(\mathbf{t}(\vv{x}_0;I_l) \cap \mathbf{t}(\vv{x}_0;I'_l))}{\P_{\vv{X}}(\mathbf{t}(\vv{x}_0;I_l))\P_{\vv{X}}(\mathbf{t}(\vv{x}_0;I'_l))}\\
    & = \left(\frac{1}{4} \sum_{j=1}^s \beta_j^2 I\{\max\{I_{lj},I'_{lj}\} = 0\} \right) 2^{\sum_{j=1}^d \min\{I_{lj},I'_{lj}\}} \\
    & + \left(\sum_{j:1\leq j\leq s, I'_{lj} - I_{lj} =1} \beta_j\left(x_{0j} - \frac{1}{2}\right)\right) \\
    & \times \left(\sum_{j':1\leq j'\leq s, I_{lj'} - I'_{lj'} =1} \beta_{j'}\left(x_{0j'} - \frac{1}{2}\right)\right)
    2^{\sum_{j=1}^d \min\{I_{lj},I'_{lj}\}}.
\end{align*}
Meanwhile, we also have
\begin{align*}
    \text{Var}_{\vv{X}}(\mu|\mathbf{t}(\vv{x}_0;I_l)) & = \frac{1}{4} \sum_{j:1\leq j\leq s,I_{lj} = 0} \beta_j^2 \\
    & = \frac{1}{4} \sum_{j=1}^s \beta_j^2 I\{I_{lj} = 0\}.
\end{align*}
Thus, from (\ref{eq:localVar}) it holds that
\begin{align*}
    \text{Var}_{\mu}(P; \vv x_0) = \text{Var}_{\vv X}(\mu|\mathbf{t}(\vv{x}_0;I_l))/\P_{\vv{X}}(\mathbf{t}(\vv{x}_0;I_l)) = \left(\frac{1}{4} \sum_{j=1}^s \beta_j^2 I\{I_{lj} = 0\}\right) 2^{\sum_{j=1}^d I_{lj}}.
\end{align*}

Substituting the equations derived above into (\ref{eq:localCov}) and \eqref{eq:localMSEens}, we finally obtain that 
\begin{align*}
    \text{MSE}(\hat{\mu}_{RF};\vv{x}_0) 
    & = \frac{B-1}{B} \E_{I_l,I'_l}\left[\left(\sum_{j: 1\leq j\leq s, I_{lj} = 0} \beta_j\left(\frac{1}{2} - x_{0j}\right)\right)\left(\sum_{j':1\leq j'\leq s, I'_{lj'} = 0} \beta_{j'}\left(\frac{1}{2} - x_{0j'}\right)\right)\right]\\
    & + \frac{1}{B} \E_{I_l}\left[\left(\sum_{j: 1\leq j\leq s,I_{lj} = 0}  \beta_j\left(\frac{1}{2} - x_{0j}\right)\right)^2\right] \\ 
    & + \frac{B-1}{B n}\E_{I_l,I'_l}\left[\left(\sigma^2 + \frac{1}{4} \sum_{j=1}^s \beta_j^2 I\{\max\{I_{lj},I'_{lj}\} = 0\} \right) 2^{\sum_{j=1}^d \min\{I_{lj},I'_{lj}\}}\right] \\
     & + \frac{B-1}{B n} \E_{I_l,I'_l}\left[\left(\sum_{j:1\leq j\leq s, I'_{lj} - I_{lj} =1}\beta_j\left(x_{0j} - \frac{1}{2}\right)\right) \right.\\
     &\left.\times\left(\sum_{j':1\leq j'\leq s, I_{lj'} - I'_{lj'} =1}\beta_{j'}\left(x_{0j'} - \frac{1}{2}\right)\right)2^{\sum_{j=1}^d \min\{I_{lj},I'_{lj}\}}\right] \\
    & +  \frac{1}{Bn}\E_{I_l}\left[\left(\sigma_0^2 + \frac{1}{4} \sum_{j=1}^s \beta_j^2 I\{I_{lj} = 0\} \right) 2^{\sum_{j=1}^d I_{lj}}\right]  \\
    & + \mathcal{R}_{RF}(\vv{x}_0),
\end{align*}
in which the remainder satisfies
\[
\mathcal{R}_{RF}(\vv{x}_0) \lesssim (1 - 2^{-l})^n + \frac{2^l}{n (1+ (n-1) 2^{-l})^{1/2}}.
\]

\subsubsection*{Derivation of (\ref{eq:binaryRF}) and (\ref{eq:binarytree})}
To further derive the global MSE, we replace $\vv{x}_0 $ in $\text{MSE}(\hat{\mu}_{RF};\vv{x}_0)$ with a random vector $\vv{X}'$ independent of $I_l$ and $I'_l$. 
Notice that 
\[
\{j: 1\leq j\leq s, I_{lj} - I_{l'j} = 1\} \cap 
\{j': 1\leq j'\leq s, I_{l'j'} - I_{lj'} = 1\}= \varnothing.
\]
Due to the component independence of $\vv{X}'$, it holds that  
\[
\E_{\vv{X'}|I_l,I_l}\left[\left(\sum_{j:1\leq j\leq s, I'_{lj} - I_{lj} =1} \beta_j\left(X'_{0j} - \frac{1}{2}\right)\right)\left(\sum_{j':1\leq j'\leq s, I_{lj'} - I'_{lj'} =1} \beta_{j'}\left(X'_{0j'} - \frac{1}{2}\right)\right)\right]=0.
\]
Hence, the fifth term in the local MSE formula will vanish after replacing $\vv{x}_0$ with $\vv{X'}$ and integrating out.
Further, we have
\begin{align*}
     &\E_{\vv{X'}|I_l,I'_l}\left[\left(\sum_{j: 1\leq j\leq s, I_{lj} = 0} \beta_j\left(\frac{1}{2} - X'_{0j}\right)\right)\left(\sum_{j':1\leq j'\leq s, I'_{lj'} = 0} \beta_{j'}\left(\frac{1}{2} - X'_{0j'}\right)\right)\right]\\
     =& \sum_{j: 1\leq j\leq s, I_{lj} = I'_{lj} = 0} \beta_j^2\text{Var}(X'_j) = \frac{1}{4} \sum_{j=1}^s I\{\max\{I_{lj},I'_{lj}\} = 0\}.
\end{align*}

Consequently, by exchanging the order of integration with respect to $\vv{X'}$ and binary CART processes in the local MSE formula, we can deduce that 
\begin{align*}
     \text{MSE}(\hat{\mu}_{\text{RF}}) & = \frac{B - 1}{4B} \E\left[\sum_{j=1}^s \beta_j^2 (1 - \max\{I_{lj},I'_{lj}\})\right]
         + \frac{1}{4B} \E\left[\sum_{j=1}^s \beta_j^2(1- I_{lj})\right] \\
        & + \frac{B-1}{B}\E\left[\left(\sigma_0^2 + \frac{1}{4} \sum_{j=1}^s \beta_j^2 (1 - \max\{I_{lj},I'_{lj}\})\right)\frac{2^{\sum_{j=1}^d \min\{I_{lj},I'_{lj}\}}}{n}\right]\\
        & + \frac{1}{B}\E\left[\left(\sigma_0^2 + \frac{1}{4}\sum_{j=1}^s \beta_j^2 (1- I_{lj})\right)\frac{2^{\sum_{j=1}^d I_{lj}}}{n}\right] + \mathcal{R}_{RF},
\end{align*}
where the bound of the remainder after integration remains the same, i.e.,
\[
\mathcal{R}_{RF} \lesssim (1 - 2^{-l})^n + \frac{2^l}{n(1 + (n-1) 2^{-l})^{1/2}}.
\]

\subsubsection*{Derivation of convergence rate in (\ref{eq:binrate})}

In what follows, we will study the converge rate for the squared bias part in (\ref{eq:binarytree}). 
Let us first observe that the variance part in (\ref{eq:binaryRF}) satisfies
\begin{equation}\label{eq:binaryvar}
\begin{split}
    &\frac{B-1}{B}\E\left[\left(\sigma_0^2 + \frac{1}{4} \sum_{j=1}^s \beta_j^2 I\{\max\{I_{lj},I'_{lj}\} = 0\}\right)\frac{2^{\sum_{j=1}^d \min\{I_{lj},I'_{lj}\}}}{n}\right]\\
    +& \frac{1}{B}\E\left[\left(\sigma_0^2 + \frac{1}{4}\sum_{j=1}^s \beta_j^2 I\{I_{lj} = 0 \}\right)\frac{2^{\sum_{j=1}^d I_{lj}}}{n}\right]\\
    \leq& \left(\sigma_0^2 + \frac{1}{4}\sum_{j=1}^s \beta_j^2\right) \frac{2^l}{n}.
    \end{split}
\end{equation}
Meanwhile, denote by $N_l = \sum_{j=1}^s I\{I_{lj} = 0\}$ the number of informative variables not being split by time $l$. It is easy to see that the squared bias part in (\ref{eq:binaryRF}) satisfies
\begin{align*}
    \frac{B - 1}{4B} \sum_{j=1}^s \beta_j^2 \E[I\{\max\{I_{lj},I'_{lj}\} = 0\}] +\frac{1}{4B} \sum_{j=1}^s \beta_j^2 \E[I\{I_{lj} = 0]\} \leq \left(\max_{1\leq j\leq s} \frac{\beta_j^2}{4}\right) \E[N_l].
\end{align*}
Hence, it suffices to analyze the convergence rate for $\E[N_l]$.

Note that the process $\{N_k\}_{0\leq k\leq l}$ is a \textit{homogenous absorbing Markov chain} on the state space $E=\{0,1,\ldots,s\}$ with transition probability matrix
\begin{align*}
    Q = (q_{ij}) = \bordermatrix{
    & 0 & 1& \dots & s-1 & s\cr
   0& 1 &  &        &     &  \cr
   1& 1- q_1& q_1&  &     & \cr
   \vdots& & \ddots& \ddots &&\cr
   s-1& & & &q_{s-1} & \cr
   s & & & & 1- q_s & q_s\cr
    },
\end{align*}
in which 
\[
q_i = \frac{\binom{d - i}{\lceil \gamma d\rceil}}{\binom{d}{\lceil \gamma d\rceil}} = \left(1 - \frac{i}{d}\right)\cdots \left(1 - \frac{i}{d - \lceil\gamma d\rceil + 1}\right)
\]
represents the probability that the next split is on non-informative variables, given that there are $i$ informative variables that have not been split yet. 
It then follows that 
\begin{align*}
    \E(N_{l+1} | N_l = i) & = i q_i + (i-1) (1 -q_i) = i - (1 - q_i).
\end{align*}

Recall that function $W_{\gamma,d}(x)$ in (\ref{eq:Wfunc}) is decreasing as $x$ increases and its derivative satisfies 
\begin{equation}\label{eq:W-Wderiv}
W'_{\gamma,d}(x) = - W_{\gamma,d}(x) \left(\sum_{k=0}^{\lceil \gamma d\rceil -1} \frac{1}{d - j - x}\right).
\end{equation}
According to Taylor's theorem, it holds that for $0 \leq i \leq s$, there exists a $\theta\in(0,1)$ such that
\begin{align*}
    1 - q_i = W_{\gamma,d}(0) - W_{\gamma,d}(i) & = -W'_{\gamma,d}(\theta i) i \\
    & = i W(\theta i) \left(\sum_{k=0}^{\lceil \gamma d\rceil -1} \frac{1}{d - j}\right) \\
    & \geq i W_{\gamma,d}(s) \frac{\lceil \gamma d\rceil}{d} = i W_{\gamma,d}(s)\gamma \frac{3}{4}.
\end{align*}
From this observation, we see that
\begin{align*}
    \E(N_{l+1}|N_l = i) & \leq  i (1 - \frac{3}{4}\gamma W_{\gamma,d}(s)),
\end{align*}
and as such, since $N_0 \equiv s$ we have
\[
    \E(N_l) \leq \left(1 - \frac{3}{4}\gamma W_{\gamma,d}(s)\right)^{l + 1} s.
\]

Combining the results above, we finally obtain that 
\begin{align*}
    \max\{\text{MSE}(\hat{\mu}_{RF}),\text{MSE}(\hat{\mu}_{\text{tree}})\} &\lesssim \left(\max_{1\leq j\leq s} \frac{\beta_j^2}{4}\right)\left(1 - \frac{3}{4}\gamma W(s)\right)^{l+1} s \\
    &+ \left(\sigma_0^2 + \frac{1}{4}\sum_{j=1}^s \beta_j^2\right) \frac{2^l}{n},
\end{align*}
which completes the proof.

\subsubsection*{Derivation of the expected cross-tree covariance formula in Remark \ref{rem:3}}

In view of the definition in (\ref{eq:modindpvar}), the variance function of the partition generated by a single binary CART process is the total number of terminal nodes. Notice that at depth $l$, we have a total of $\sum_{j=1}^d I_{lj}$ splits, which means that the total number of nodes is $2^{\sum_{j=1}^d I_{lj}}$. Hence, it holds that 
\[
\E[\text{Var}(P)] = \E[2^{\sum_{j=1}^d I_{lj}}].
\]

To derive the formula for the cross-tree covariance function, we first make some general claims. For any partitions $P$ and $P'$, given a target point $\vv{x}_0$ we define
\begin{equation}\label{eq:indpcov}
    \text{Cov}(P,P';\vv{x}_0) = \frac{\P_{\vv{X}}(P_{\vv{x}_0}\cap P'_{\vv{x}_0})}{\P_{\vv{X}}(P_{\vv{x}_0})\P_{\vv{X}}(P'_{\vv{x}_0})},
\end{equation}
which is identical to the local cross-partition covariance function in (\ref{eq:localCov}) with $\sigma \equiv 1$. According to the definition in (\ref{eq:modindpcov}), it is clear that
\begin{equation}\label{eq:Exp}
    \text{Cov}(P,P') = \E_{\vv{X}'}\left[\text{Cov}(P,P';\vv{X}')\right],
\end{equation}
where $\vv{X}'$ is an independent copy of $\vv{X}$. Indeed, from (\ref{eq:indpcov}) we have
\begin{align*}
    \text{Cov}(P,P';\vv{x}_0) = \sum_{P_i\in P}\sum_{P'_j \in P'} I\{\vv{x}_0\in P_i\cap P'_j\} \frac{\P_{\vv{X}}(P_i \cap P'_j)}{\P_{\vv{X}}(P_i)\P_{\vv{X}}(P'_j)}.
\end{align*}
Replacing $\vv{x}_0$ with $\vv{X'}$ and then integrating with respect to it, we can deduce that  
\begin{align*}
    \E_{\vv{X}'}\left[\text{Cov}(P,P';\vv{X}')\right] & = \E_{\vv{X}'}\left[ \sum_{P_i\in P}\sum_{P'_j \in P'} I\{\vv{X}'\in P_i\cap P'_j\} \frac{\P_{\vv{X}}(P_i \cap P'_j)}{\P_{\vv{X}}(P_i)\P_{\vv{X}}(P'_j)} \right] \\
    & = \sum_{P_i\in P}\sum_{P'_j \in P'}\frac{\P_{\vv{X}}(P_i \cap P'_j)}{\P_{\vv{X}}(P_i)\P_{\vv{X}}(P'_j)} \P_{\vv{X'}}(\vv{X}'\in P_i\cap P_j) \\
    & = \sum_{P_i\in P}\sum_{P'_j \in P'}\frac{\P_{\vv{X}}(P_i \cap P'_j)^2}{\P_{\vv{X}}(P_i)\P_{\vv{X}}(P'_j)} 
     = \text{Cov}(P,P').
\end{align*}

For any nodes, say, $P_i$ and $P'_j$ from $P$ and $P'$ with non-empty intersection, there exist certain $\vv{x}_0$ and two independent binary CART processes $I_l$ and $I'_l$ such that $P_i = \mathbf{t}(\vv{x}_0;I_l)$ and $P'_j = \mathbf{t}(x_0;I'_l)$. Note that a coordinate in the intersection of two nodes is split as long as it is split by either $I_l$ or $I_l'$. Given $I_l$ and $I'_{l}$, it holds that 
\[
\P_{\vv{X}}(\mathbf{t}(\vv{x}_0;I_l) \cap \mathbf{t}(\vv{x}_0;I'_l)) = 2^{-\sum_{j=1}^d \max\{I_{lj},I'_{lj}\}}
\]
and
\[
\P_{\vv{X}}(\mathbf{t}(\vv{x}_0;I_l)) = 2^{-\sum_{j=1}^d I_{lj}},~~~~\P_{\vv{X}}(\mathbf{t}(\vv{x}_0;I'_l)) = 2^{-\sum_{j=1}^d I'_{lj}}.
\]
Consequently, for the ratio we have 
\begin{align*}
     \text{Cov}(P,P';\vv{x}_0)=\frac{\P_{\vv{X}}(\mathbf{t}(\vv{x}_0;I_l) \cap \mathbf{t}(\vv{x}_0;I'_l))}{\P_{\vv{X}}(\mathbf{t}(\vv{x}_0;I_l)) \P_{\vv{X}}(\mathbf{t}(\vv{x}_0;I'_l))} & = 2^{\sum_{j=1}^d I_{lj} + I'_{lj} - \max\{I_{lj},I'_{lj}\}} \\
     & = 2^{\sum_{j=1}^d \min\{I_{lj},I'_{lj}\}},
\end{align*}
which is constant for any $\vv{x}_0$. Thus, by (\ref{eq:Exp}) we can obtain that 
\[
    \text{Cov}(P,P) = \E_{\vv{X'}}\left[\text{Cov}(P,P';\vv{X'})\right] = 2^{\sum_{j=1}^d \min\{I_{lj},I'_{lj}\}}.
\]
Therefore, the final conclusion follows by integrating with respect to $I_l$ and $I'_{l}$. This completes the proof of Theorem \ref{new.thm.binary}.

\subsection{Proof of Theorem \ref{new.thm.unif}}\label{new.example2.derive}

\subsubsection*{Derivation of the local MSE formula}

We first calculate the local MSE for the related random forests estimator $\hat{\mu}_{RF}$ according to our general Theorem \ref{new.thm1}.
Let $\mathbf{x}_0 = (x_{01}, \ldots, x_{0d})^\top \in [0, 1]^d$ be the target point.
The proof here follows similar lines as in Section \ref{new.example1.derive}, where we first address the squared bias part and then the variance part.

On the one hand, let us recall the representation (\ref{eq:biasdecompose}) of the squared bias part for the tree ensemble. Then for two independent bipartitions driven by two independent uniform CART processes $J_l$ and $J'_l$,  
the squared bias part in the local MSE formula can be written as
\begin{align*}
    &\text{The squared bias of tree ensemble}\\
     =& \E_{J_l,J'_l}\left[(\mu(\vv{x}_0) - \E_{\vv{X}}(\mu | \mathbf{t}(x_0;J_l))) (\mu(\vv{x}_0) - \E_{\vv{X}}(\mu | \mathbf{t}(x_0;J'_l))) \right].
\end{align*}
Given a realization of a uniform CART process $J_l$, in light of the discussion related to  (\ref{eq:terminalinterval-all}), the terminal node containing $\vv{x}_0$ is given by
\[
\mathbf{t}(\vv{x}_0,J_l) = \prod_{i=1}^d \left(\frac{K(x_{0i},J_{li})-1}{2^{J_{li}}}, \frac{K(x_{0i},J_{li})}{2^{J_{li}}}\right],
\]
where $K(x_0,m) = \lceil x_02^m\rceil$. In addition, it holds that 
\begin{align*}
    \E_{\vv{X}}(\mu|\mathbf{t}(\vv{x}_0,J_l)) &= \sum_{i=1}^s \beta_i M(x_{0i},J_{li}),
\end{align*}
where $M(x_{0i},J_{li}) = (2K(x_{0i},J_{li}) - 1)/2^{J_{li}+1}$. Then we can show that 
\begin{align*}
    &\text{The squared bias of tree ensemble} \\
    =&  \E_{J_l,J'_l}\left[\left(\sum_{i=1}^s \beta_i \left(x_{0i} - M(x_{0i},J_{li} )\right)\right)\left(\sum_{i'=1}^s \beta_{i'}\left(x_{0i'} - M(x_{0i'},J'_{li'} )\right)\right)\right].
\end{align*}
Similarly, it also holds that 
\begin{align*}
    \text{The squared bias of one single tree}& = \E_{J_l}\left[\left(\sum_{i=1}^s \beta_i\left(x_{0i} - M(x_{0i},J_{li} )\right)\right)^2
    \right].
\end{align*}

On the other hand, observe that for two terminal nodes $\mathbf{t}(\vv{x}_0;J_l)$ and $\mathbf{t}(\vv{x}_0;J'_l)$ in $P$ and $P'$, the edge length of their intersection is $2^{-\max\{J_{li},J'_{li}\}}$ for any $i =1,\ldots,p$. It follows that 
\begin{align*}
    &\P_{\vv{X}}(\mathbf{t}(\vv{x}_0;J_l) \cap \mathbf{t}(\vv{x}_0;J'_l)) = 2^{-\sum_{i=1}^d \max\{J_{li},J'_{li}\}},\\
    &\P_{\vv{X}}(\mathbf{t}(\vv{x}_0;J_l)) = \P_{\vv{X}}(\mathbf{t}(\vv{x}_0;J'_l)) = 2^{-l},
\end{align*}
and thus,
\begin{align*}
    \frac{\P_{\vv{X}}(\mathbf{t}(\vv{x}_0;I_l) \cap \mathbf{t}(\vv{x}_0;I'_l))}{\P_{\vv{X}}(\mathbf{t}(\vv{x}_0;I_l))\P_{\vv{X}}(\mathbf{t}(\vv{x}_0;I'_l))} & = 2^{\sum_{i=1}^d \min\{J_{li},J'_{li}\}}.
\end{align*}

For the model-error induced cross-partition covariance in \eqref{eq:localCov}, since the model error has a constant variance, we have that 
\[
\text{Cov}_{\sigma^2}(P,P') = \sigma_0^2 \frac{\P_{\vv{X}}(\mathbf{t}(\vv{x}_0;J_l) \cap \mathbf{t}(\vv{x}_0;J'_l))}{\P_{\vv{X}}(\mathbf{t}(\vv{x}_0;J_l))\P_{\vv{X}}(\mathbf{t}(\vv{x}_0;J'_l))} = \sigma_0^2 2^{\sum_{i=1}^d \min\{J_{li},J'_{li}\}},
\]
where $P$ and $P'$ are two independent bi-partitions driven by two independent binary CART processes $J_l$ and $J'_l$. Further,
the model-error-induced local partition variance function becomes
\[
\text{Var}_{\sigma^2}(P) = \sigma_0^2 \frac{1}{\P_{\vv{X}}(\mathbf{t}(\vv{x}_0,J_l))} = \sigma_0^2 2^{\sum_{i=1}^d J_{li}} = \sigma_0^2 2^l.
\]

As for the signal-induced cross-partition covariance function, we recall the decomposition formula (\ref{eq:keydecompose}) and can see that the coefficient in (\ref{eq:localCov}) satisfies
\begin{align*}
    &\E_{\vv{X}}\left[(\mu - \E_{\vv{X}}(\mu|\mathbf{t}(\vv{x}_0;J_l)))(\mu - \E_{\vv{X}}(\mu|\mathbf{t}(\vv{x}_0;J'_l)))\big| \mathbf{t}(\vv{x}_0;J_l) \cap \mathbf{t}(\vv{x}_0;J'_l)\right] \\
    =& \text{Var}_{\vv{X}}(\mu | \mathbf{t}(\vv{x}_0;J_l) \cap \mathbf{t}(\vv{x}_0;J'_l)) \\
    + &  (\E_{\vv{X}}(\mu|\mathbf{t}(\vv{x}_0;J_l)\cap \mathbf{t}(\vv{x}_0;J'_l)) - \E_{\vv{X}}(\mu|\mathbf{t}(\vv{x}_0;J_l)))\\
    \times &
    (\E_{\vv{X}}(\mu|\mathbf{t}(\vv{x}_0;J_l)\cap \mathbf{t}(\vv{x}_0;J'_l)) - \E_{\vv{X}}(\mu|\mathbf{t}(\vv{x}_0;J'_l))).
\end{align*}
Notice that
\begin{align*}
    \text{Var}_{\vv{X}}(\mu|\mathbf{t}(\vv{x}_0;J_l)\cap \mathbf{t}(\vv{x}_0;J'_l)) & = \frac{1}{12} \sum_{i=1}^s \beta_i^2 2^{-2\max\{J_{li},J'_{li}\}}
\end{align*}
and
\begin{align*}
    &(\E_{\vv{X}}(\mu|\mathbf{t}(\vv{x}_0;J_l)\cap \mathbf{t}(\vv{x}_0;J'_l)) - \E_{\vv{X}}(\mu|\mathbf{t}(\vv{x}_0;J_l)))\\
    &\times
    (\E_{\vv{X}}(\mu|\mathbf{t}(\vv{x}_0;J_l)\cap \mathbf{t}(\vv{x}_0;J'_l)) - \E_{\vv{X}}(\mu|\mathbf{t}(\vv{x}_0;J'_l)))\\
    =&\left(\sum_{i:1\leq i\leq s, J'_{li} > J_{li}}\beta_i \left(M(x_{0i},J'_{li}) - M(x_{0i},J_{li})\right)\right)\\
    &\times \left(\sum_{i':1\leq i'\leq s, J_{li'} > J'_{li'}}\beta_{i'}\left(M(x_{0i'},J_{li'}) - M(x_{0i'},J'_{li'})\right)\right).
\end{align*}

Hence, for two independent bi-partitions $P$ and $P'$ driven by $J_l$ and $J'_l$, we can deduce that 
\begin{align*}
    \text{Cov}_{\mu}(P,P) & = \E_{\vv{X}}\left[(\mu - \E_{\vv{X}}(\mu|\mathbf{t}(\vv{x}_0;J_l)))(\mu - \E_{\vv{X}}(\mu|\mathbf{t}(\vv{x}_0;J'_l)))\big| \mathbf{t}(\vv{x}_0;J_l) \cap \mathbf{t}(\vv{x}_0;J'_l)\right]\\
    & \times \frac{\P_{\vv{X}}(\mathbf{t}(\vv{x}_0;J_l) \cap \mathbf{t}(\vv{x}_0;J'_l))}{\P_{\vv{X}}(\mathbf{t}(\vv{x}_0;I_l))\P_{\vv{X}}(\mathbf{t}(\vv{x}_0;J'_l))}\\
    & = \text{Var}_{\vv{X}}(\mu|\mathbf{t}(\vv{x}_0;J_l)\cap \mathbf{t}(\vv{x}_0;J'_l))  \frac{\P_{\vv{X}}(\mathbf{t}(\vv{x}_0;J_l) \cap \mathbf{t}(\vv{x}_0;J'_l))}{\P_{\vv{X}}(\mathbf{t}(\vv{x}_0;J_l))\P_{\vv{X}}(\mathbf{t}(\vv{x}_0;J'_l))}\\
    & + (\E_{\vv{X}}(\mu|\mathbf{t}(\vv{x}_0;J_l)\cap \mathbf{t}(\vv{x}_0;J'_l)) - \E_{\vv{X}}(\mu|\mathbf{t}(\vv{x}_0;J_l)))\\
    &\times
    (\E_{\vv{X}}(\mu|\mathbf{t}(\vv{x}_0;J_l)\cap \mathbf{t}(\vv{x}_0;J'_l)) - \E_{\vv{X}}(\mu|\mathbf{t}(\vv{x}_0;J'_l)))\\
    &\times \frac{\P_{\vv{X}}(\mathbf{t}(\vv{x}_0;J_l) \cap \mathbf{t}(\vv{x}_0;J'_l))}{\P_{\vv{X}}(\mathbf{t}(\vv{x}_0;I_l))\P_{\vv{X}}(\mathbf{t}(\vv{x}_0;J'_l))}\\
    & = \left(\frac{1}{12} \sum_{i=1}^s \beta_i^2 2^{-2\max\{J_{li},J'_{li}\}} \right) 2^{\sum_{i=1}^d \min\{J_{li},J'_{li}\}} \\
    & + \left(\sum_{i:1\leq i\leq s, J'_{li} > J_{li}}\beta_i \left(M(x_{0i},J'_{li}) - M(x_{0i},J_{li})\right)\right)\\
    &\times \left(\sum_{i':1\leq i'\leq s, J_{li'} > J'_{li'}}\beta_{i'}\left(M(x_{0i'},J_{li'}) - M(x_{0i'},J'_{li'})\right)\right)\\
    &\times 2^{\sum_{i=1}^d \min\{J_{li},J'_{li}\}}.
\end{align*}
Meanwhile, since
\begin{align*}
    \text{Var}_{\vv{X}}(\mu|\mathbf{t}(\vv{x}_0,J_l)) & = \frac{1}{12} \sum_{i=1}^s \beta_i^2 2^{-2J_{li}},
\end{align*}
it follows from (\ref{eq:localVar}) that 
\begin{align*}
    \text{Var}_{\mu}(P; \vv x_0) = \text{Var}_{\vv X}(\mu|\mathbf{t}(\vv{x}_0;J_l)) / \P_{\vv{X}}(\mathbf{t}(\vv{x}_0;J_l)) = \left(\frac{1}{12} \sum_{i=1}^s \beta_i^2 2^{-2J_{li}}\right) 2^{l}.
\end{align*}

Substituting the equations derived above into (\ref{eq:localCov}) and \eqref{eq:localMSEens}, we can finally obtain that 
\begin{align*}
    \text{MSE}(\hat{\mu}_{RF}; \vv x_0) &=  \frac{B-1}{B}\E_{J_l,J'_l}\left[\left(\sum_{i=1}^s \beta_i\left(x_{0i} - M(x_{0i},J_{li} )\right)\right)\left(\sum_{i'=1}^s \beta_{i'}\left(x_{0i'} - M(x_{0i'},J'_{li'} )\right)\right)\right]\\
    & + \frac{1}{B}\E_{J_l}\left[\left(\sum_{i=1}^s \beta_i \left(x_{0i} - M(x_{0i},J_{li} )\right)\right)^2\right]\\
    & + \frac{B-1}{Bn}\E_{J_l,J'_l}\left[\left(\sigma_0^2 +\frac{1}{12} \sum_{i=1}^s \beta_i^2 2^{-2\max\{J_{li},J'_{li}\}}\right) 2^{\sum_{i=1}^d \min\{J_{li},J'_{li}\}}\right]\\
    & + \frac{B-1}{Bn}\E_{J_l,J'_l}\Bigg[\left(\sum_{i:1\leq i\leq s, J'_{li} > J_{li}}\beta_i\left(M(x_{0i},J'_{li}) - M(x_{0i},J_{li})\right)\right)\\
    &\times \left(\sum_{i':1\leq i'\leq s, J_{li'} > J'_{li'}} \beta_{i'}\left(M(x_{0i'},J_{li'}) - M(x_{0i'},J'_{li'})\right)\right)2^{\sum_{i=1}^d \min\{J_{li},J'_{li}\}}\Bigg]\\
     & + \frac{1}{B n} \E_{J_l}\left[\left(\sigma_0^2 + \frac{1}{12} \sum_{i=1}^s \beta_i^2 2^{-2J_{li}}\right) 2^l\right] \\
    & + \mathcal{R}_{RF}(\vv{x}_0),
\end{align*}
in which the remainder satisfies
\[
\mathcal{R}_{RF}(\vv{x}_0) \lesssim (1 - 2^{-l})^n + \frac{2^l}{n (1+ (n-1) 2^{-l})^{1/2}}.
\]

\subsubsection*{Derivation of (\ref{eq:unifRF}) and (\ref{eq:uniftree})}
We now proceed with examining the global MSE formula. 
Observe that the uniform CART process is independent of $\vv{x}_0$. Then by replacing $x_{0i}$ with a uniform variable $X_i$, it holds that 
\begin{align*}
   \E_{X_i|J_{li}}(M(X_i, J_{li}) - X_i) &= \E_{X_i|J_{li}}\left[\sum_{k=1}^{2^{J_{li}}} \left(\frac{2k-1}{2^{J_{li}+1}} - X_j\right)I\left\{\frac{k-1}{2^{J_{li}}}< X_j \leq \frac{k}{2^{J_{li}}}\right\}\right]\\
   &=\sum_{k=1}^{2^{J_{li}}} 2^{-J_{li}}\left(\frac{2k-1}{2^{J_{li}+1}} - \frac{2k-1}{2^{J_{li}+1}} \right) = 0,\\
   \E_{X_i|J_{li}}[(M(X_i, J_{li}) - X_i)^2] &= \E_{X_i|J_{li}}\left[\sum_{k=1}^{2^{J_{li}}} \left(\frac{2k-1}{2^{J_{li}+1}} - X_j\right)^2I\left\{\frac{k-1}{2^{J_{li}}}< X_j \leq \frac{k}{2^{J_{li}}}\right\}\right]\\
   &=\sum_{k=1}^{2^{J_{li}}} \frac{1}{2^{J_{li}}} \frac{1}{3}\left(\frac{1}{2^{J_{li} + 1}}\right)^2 = \frac{1}{12} 2^{-2J_{jl}}.
\end{align*}
In addition, we have that 
\[
\{i:~1\leq i\leq s, J_{li} < J_{li}\} \cap  \{i':~1\leq i'\leq s, J_{li'} > J_{li'}\} =\varnothing.
\]
Thanks to the component independence of $\vv{X}'$, we can show that 
\begin{align*}
\E_{\vv{X}'|J_{l},J'_{l}}\Bigg[&\left(\sum_{i:1\leq i\leq s, J'_{li} > J_{li}} \beta_i\left(M(X'_i,J'_{li}) - M(X'_i,J_{li})\right)\right)\\
\times &\left(\sum_{i':1\leq i'\leq s, J_{li'} > J'_{li'}} \beta_{i'}\left(M(X'_{i'},J_{li'}) - M(X'_{i'},J'_{li'})\right)\right)\Bigg]=0.    
\end{align*}

Meanwhile, it follows that 
\begin{align*}
     & \E_{\vv{X}'|J_l,J'_{l}}\left[\left(\sum_{i=1}^s \beta_i\left(X_i - M(X_i,J_{li} )\right)\right)\left(\sum_{i'=1}^s \beta_{i'}\left(X_{i'} - M(X_{i'},J'_{li'} )\right)\right)\right]\\
     &= \frac{1}{12}\sum_{i = 1}^s 2^{-2\max\{J_{li}, J'_{li}\}}.
\end{align*}
Summarizing the results derived above, we can deduce that 
\begin{align*}
    \text{MSE}(\hat{\mu}_{RF})
    & = \frac{1}{12B} \E\left[\sum_{j = 1}^s \beta_i^2 2^{-2J_{li}}\right]
    +  \frac{B-1}{12 B} \E\left[\sum_{i = 1}^s \beta_i^2 2^{-2\max\{J_{li}, J'_{li}\}}\right]\\
    & + \frac{1}{B}\E\left[\left(\sigma_0^2 + \frac{1}{12}\sum_{i = 1}^s  \beta_i^2 2^{-2 J_{li}}\right) \frac{2^l}{n}\right] \\
    & + \frac{B-1}{B}\E\left[\left(\sigma_0^2 +\frac{a^2}{12} \sum_{i=1}^s \beta_i^2 2^{-2\max\{J_{li},J'_{li}\}}\right) \frac{2^{\sum_{i=1}^d \min\{J_{li},J'_{li}\}}}{n}\right]\\
    & + \mathcal{R}_{RF},
\end{align*}
where the remainder is bounded by the same one as in the local version formula
\[
\mathcal{R}_{RF} \lesssim (1 - 2^{-l})^n + \frac{2^l}{n (1+ (n-1) 2^{-l})^{1/2}}.
\]

\subsubsection*{Derivation of the convergence rate }
To investigate the convergence rate, we first notice that the convergence rates of the variance parts in (\ref{eq:unifRF}) and (\ref{eq:uniftree}) are bounded in order $2^{l}/n$. Hence, it suffices to focus on the convergence rate for the squared bias part, which is bounded by
\[
\left(\max_{1\leq i\leq s}\frac{\beta_i^2 }{12}\right) \E\left[\sum_{i=1}^s 2^{-2 J_{li}}\right].
\]

Denote by $H_l = \sum_{i=1}^s 2^{-2 J_{li}}$. We see that $H_l$ is non-increasing and satisfies
\begin{align*}
    H_l - H_{l+1} & = 
    \sum_{i=1}^s 
    \left(2^{-2J_{li}} - 2^{-2J_{l+1,i}}\right)
    I\{\text{the $(l+1)$th split is on $i$}\} \\
    & = \frac{3}{4}
    \sum_{i=1}^s 2^{-2 J_{l,i}} 
    I\{\text{the $(l+1)$th split is on $i$}\}.
\end{align*}
Let us define 
$P_{J_l}(i) = \P(\text{the $(l+1)$th split is on $i$}|J_l)$ 
for $1\leq i\leq s$. It holds that 
\begin{align*}
    \E(H_{l+1}|J_l) & = \sum_{i=1}^s 2^{-2J_{li}} \left(1 - \frac{3}{4}P_{J_l}(i)\right). 
\end{align*}

We next prove that 
\begin{equation}\label{eq:unifbound}
    \min_{1\leq i\leq s} P_{J_l}(i) \geq \gamma \left(1 - \frac{s}{d}\right)\cdots \left(1 - \frac{s}{d - \lceil \gamma d\rceil +1}\right)=\gamma W_{\gamma,d}(s),
\end{equation}
where the RHS above is independent of $J_l$, and $W(s)$ is the same as defined in the last subsection. To this end, denote by $n_{l,i} = \#\{1\leq j\leq s: \beta_j^2 2^{-2J_{l,j}} = \beta_i^2 2^{-2J_{l,i}}\}$ and 
$m_{l,i} = \#\{1\leq j\leq s: \beta_j^2 2^{-2J_{l,j}} < \beta_i^2 2^{-2J_{l,i}}\}$.
It follows from the definition that
\begin{align*}
    P_{J_l}(i) & = \frac{1}{n_{l,i}} \frac{\binom{d-s + m_{l,i} + n_{l,i}}{\lceil \gamma d\rceil} - \binom{d-s +  n_{l,i}}{\lceil \gamma d\rceil}
    }{\binom{d}{\lceil \gamma d\rceil}} \\
    & = \frac{1}{n_{l,i}}\bigg\{\left(1 - \frac{s - m_{l,i} - n_{l,i}}{d}\right)\cdots \left(1 - \frac{s - m_{l,i} - n_{l,i}}{d - \lceil \gamma d\rceil +1}\right)\\
    &~~~~~~~ - \left(1 - \frac{s - m_{l,i}}{d}\right)\cdots \left(1 - \frac{s - m_{l,i}}{d - \lceil \gamma d\rceil +1}\right)\bigg\}.
\end{align*}
With an application of \eqref{eq:W-Wderiv} and  Taylor's theorem, it holds that for $0 \leq x \leq y \leq s$, there exists a $\theta\in(0,1)$ such that
\begin{align*}
    W_{\gamma,d}(x) - W_{\gamma,d}(y) & =  W'_{\gamma,d}(x + \theta y) (x - y) \\
    & = (y - x) W_{\gamma,d}(x + \theta y) \left(\sum_{k=0}^{\lceil \gamma d\rceil -1} \frac{1}{d - j - x - \theta y}\right) \\
    & \geq (y - x) W_{\gamma,d}(s) \frac{\lceil \gamma d\rceil}{d}.
\end{align*}

Substituting $x = s - m_{l,i} - n_{l,i}$ and $y = s - m_{l,i}$ into the inequality above, we have
\begin{align*}
    P_{J_l}(i) & = \frac{1}{n_{l,i}} (W_{\gamma,d}(s - m_{l,i} - n_{l,i}) - W_{\gamma,d}(s - n_{l,i})) \\
    & \geq W_{\gamma,d}(s) \frac{\lceil \gamma d\rceil}{d} \geq \gamma W_{\gamma,d}(s).
\end{align*}
As such, it follows that 
\[
   \E(H_{l+1}) \leq \left( 1 - \frac{3}{4}\gamma W_{\gamma,d}(s)\right) \E(H_l) 
\]
and consequently, 
\[
    \E(H_l) \leq \left( 1 - \frac{3}{4}\gamma W_{\gamma,d}(s)\right)^{l+1} s.
\]
Combining the results above, we finally obtain that 
\begin{align*}
    \max\{\text{MSE}(\hat{\mu}_{RF}),\text{MSE}(\hat{\mu}_{\text{tree}})\} &\lesssim \left(\max_{1\leq j\leq s} \frac{\beta_j^2}{12}\right)\left(1 - \gamma W(s)\right)^{l+1} s \\
    &+ \left(\sigma_0^2 + \frac{1}{12}\sum_{j=1}^s \beta_j^2\right) \frac{2^l}{n},
\end{align*}
which completes the proof.

\subsubsection*{Derivation of the expected cross-tree covariance formula in Remark \ref{rem:6}}

The derivation of the formula in Remark \ref{rem:6} is generally similar to that in Remark \ref{rem:3}. Thus, we only emphasize the differences and omit repeated details here. In view of (\ref{eq:modindpvar}), since there are $\sum_{j=1}^d I_{lj}$ splits at depth $l$, the size of terminal nodes is $2^{\sum_{j=1}^d I_{lj}}$ and hence $\text{Var}(P) = 2^{\sum_{j=1}^d I_{lj}}$.

Meanwhile, for two terminal nodes $\mathbf{t}(\vv{x}_0;J_l)$ and $\mathbf{t}(x_0;J'_l)$ in $P$ and $P'$, the edge length of their intersection is $2^{-\max\{J_{li},J'_{li}\}}$ for any $i =1,\ldots,p$. Then it holds that 
\begin{align*}
     \text{Cov}(P,P';\vv{x}_0)=\frac{\P_{\vv{X}}(\mathbf{t}(\vv{x}_0;J_l) \cap \mathbf{t}(\vv{x}_0;J'_l))}{\P_{\vv{X}}(\mathbf{t}(\vv{x}_0;J_l)) \P_{\vv{X}}(\mathbf{t}(\vv{x}_0;J'_l))} & = 2^{\sum_{i=1}^d J_{li} + J'_{li} - \max\{J_{li},J'_{li}\}} \\
     & = 2^{\sum_{i=1}^d \min\{J_{li},J'_{li}\}},
\end{align*}
which remains constant for any $\vv{x}_0$. Therefore, it follows from (\ref{eq:Exp}) that 
\[
    \text{Cov}(P,P) = \E_{\vv{X'}}\left[\text{Cov}(P,P';\vv{X'})\right] = 2^{\sum_{i=1}^d \min\{J_{li},J'_{li}\}},
\]
which yields the final conclusion. This concludes the proof of Theorem \ref{new.thm.unif}.

\subsection{Proof of Proposition \ref{prop:globaltrieq}}\label{new.SecA.prop2}

Let us recall that
\begin{align*}
    &\E_{\vv{X}}((\mu - \E_{\vv{X}}(\mu|P_i))(\mu - \E_{\vv{X}}(\mu|P'_j))|P_i\cap P'_j) \frac{\P_{\vv{X}}(P_i\cap P'_j)^2}{\P_{\vv{X}}(P_i)\P_{\vv{X}}(P'_j)}\\
    =&\E_{\vv{X}}((\mu - \E_{\vv{X}}(\mu|P_i))(\mu - \E_{\vv{X}}(\mu|P'_j))I_{P_i\cap P'_j}) \frac{\P_{\vv{X}}(P_i\cap P'_j)}{\P_{\vv{X}}(P_i)\P_{\vv{X}}(P'_j)}\\
    \leq& \sqrt{\frac{\E_{\vv{X}}((\mu - \E_{\vv{X}}(\mu|P_i))^2|P_i)
    \E_{\vv{X}}((\mu - \E_{\vv{X}}(\mu|P'_j))^2|P'_j)}{\P_{\vv{X}}(P_i)\P_{\vv{X}}(P'_j)}} \P_{\vv{X}}(P_i\cap P'_j).
\end{align*}
Meanwhile, it holds that 
\begin{align*}
    \text{Var}_{\mu}(P) &= \sum_{P_i \in P} {\rm Var}(\mu|P_i) 
     = \sum_{P_i \in P} \E((\mu - \E_{\vv{X}}(\mu|P_i))^2|P_i)\\
     & = \E_{\vv{X}'}\left\{\sum_{P_i\in P} \sqrt{\frac{\E((\mu - \E_{\vv{X}}(\mu|P_i))^2|P_i)}{\P_{\vv{X}}}(P_i)} I\{\vv{X'} \in P_i\}\right\}^2,
\end{align*}
where $\vv{X'}$ is an independent copy of $\vv{X}$.

Then it follows that
\begin{align*}
    \text{Cov}_{\mu}(P,P') &\leq \sqrt{\frac{\E_{\vv{X}}((\mu - \E_{\vv{X}}(\mu|P_i))^2|P_i)
    \E_{\vv{X}}((\mu - \E_{\vv{X}}(\mu|P'_j))^2|P'_j)}{\P_{\vv{X}}(P_i)\P_{\vv{X}}(P'_j)}} \P_{\vv{X}}(P_i\cap P'_j). \\
    & = \E_{\vv{X'}}\Bigg\{\left(\sum_{P_i\in P}\sqrt{\frac{\E_{\vv{X}}((\mu - \E_{\vv{X}}(\mu|P_i))^2|P_i)}{\P_{\vv{X}}(P_i)}} I\{\vv{X'} \in P_i\}\right) \\
    & ~~~~~~~~~\times \left(\sum_{P'_j\in P'}\sqrt{\frac{\E_{\vv{X}}((\mu - \E_{\vv{X}}(\mu|P'_j))^2|P'_j)}{\P_{\vv{X}}(P'_j)}} I\{\vv{X'} \in P'_j\}\right)\Bigg\}\\
    & \leq \sqrt{\E_{\vv{X'}}\Bigg\{\left(\sum_{P_i\in P}\sqrt{\frac{\E_{\vv{X}}((\mu - \E_{\vv{X}}(\mu|P_i))^2|P_i)}{\P_{\vv{X}}(P_i)}} I\{\vv{X'} \in P_i\}\right)^2\Bigg\}}\\
    & \times\sqrt{\E_{\vv{X'}}\Bigg\{
    \left(\sum_{P'_j\in P'}\sqrt{\frac{\E_{\vv{X}}((\mu - \E_{\vv{X}}(\mu|P'_j))^2|Q')}{\P_{\vv{X}}(P'_j)}} I\{\vv{X'} \in P'_j\}\right)^2\Bigg\}}\\
    & \leq \sqrt{\text{Var}_\mu(P)\text{Var}_\mu(P')},
\end{align*}
which proves the triangle inequality for the signal-induced cross-partition covariance function. 
The equality above holds if and only if for any $P_i\in P$, $P'_j\in P'$ with $P_i\cap P'_j\neq\varnothing$,
$(\mu - \E_{\vv{X}}(\mu|P_i))I_{P_i} = (\mu - \E_{\vv{X}}(\mu|P'_j))I_{P'_j}$. 
According to the discussion in Section \ref{new.SecA.prop1}, we know that this holds if and only if $\mu$ is constant on $P_i\cup P'_j$ or $P_i$ and $P'_j$ are indistinguishable. 

We next show the triangle inequality for the model-error dependent covariance function. 
Since $\P_{\vv{X}}(P_i \cap P'_j) \leq \min\{\P_{\vv{X}}(P_i), \P_{\vv{X}}(P'_j)\}$, we have that
\begin{align*} 
            \text{Corr}_{\sigma^2}[P,P']& = \sum_{P_i \in P} \frac{1}{\P_{\vv{X}}(P_i)}\left(\sum_{P'_j\in P'}\E_{\vv{X}}[\sigma^2|P_i\cap P_j]\frac{\P_{\vv{X}}(P_i \cap P'_j)^2}{\P_{\vv{X}}(P'_j)}\right)\\
            & \leq \sum_{P_i \in P} \frac{1}{\P_{\vv{X}}(P_i)}\left(\sum_{P'_j\in P'}\E_{\vv{X}}[\sigma^2 I_{P_i\cap P_j}]\right) \\
            & = \sum_{P_i \in P} \frac{\E_{\vv{X}}(\sigma^2 I_{P_i})}{\P_{\vv{X}}(P_i)} = \sum_{P_i \in P} \E_{\vv{X}}[\sigma^2|P_i] = \text{Var}_{\sigma^2}(P).
\end{align*}
Thus, we can conclude that the correlation function is always bounded by one; that is, $\text{ Cov}_{\sigma^2}(P,P') \leq \sqrt{\text{Var}_{\sigma^2}(P)\text{Var}_{\sigma^2}(P')}$. The equality above holds if and only if 
\[
\frac{\P_{\vv{X}}(P_i\cap P'_j)}{\P_{\vv{X}}(P'_j)} = \frac{\P_{\vv{X}}(P_i\cap P'_j)}{\P_{\vv{X}}(P_i)}= 1
\]
for any $P_i$ and $P'_j$, which entails that $P$ and $P'$ are indistinguishable. This completes the proof of Proposition \ref{prop:globaltrieq}.

\subsection{Proof of Theorem \ref{new.thm2}} \label{new.SecA.3}

The global MSE can be derived by integrating the local MSE with respect to the testing data $\vv{X'}$ independent of the training data $\vv{Z}$, i.e.,
\[
\text{MSE}(\hat{\mu}_{\ens}) = \E_{\vv{X'}}[\text{MSE}(\hat{\mu}_{\ens};\vv{X'})].
\]
Still, we omit the superscript $n$ from $\vv{P}^n$ with no ambiguity. Recall that given a partition $P$, the corresponding partitioning estimator $\hat{\mu}_{\part}(\cdot;\vv{Z},P)$ is defined by
\begin{align*}
    \hat{\mu}_{\part}(\vv{x}_0;\vv{Z}, P) = \frac{1}{N(P_{\vv{x}_0})}\sum_{i=1}^n Y_i I\{\vv{X}_i \in P_{\vv{x}_0}\}.
\end{align*}
However, this representation is inconvenient in analyzing the global MSE since the target point and the cell in partition are mixed up in $\vv{P}_{\vv{x}_0}$ so it is not easy to separate the testing data and the exogenous factors to exchange the order of integration. Instead, we prefer to work with an \textit{alternative} representation 
\begin{equation}\label{eq:thm2.1}
    \hat{\mu}_{\part}(\vv{x}_0;\vv{Z}, P) = \sum_{P_i \in P} \left(\frac{1}{N(P_i)}\sum_{i=1}^n Y_i I\{\vv{X}_i \in P_i\} \right) I\{\vv{x}_0 \in P_i\}.
\end{equation}
Clearly, the form in (\ref{eq:thm2.1}) above is more convenient in exchanging the order of integration since the information of the target point $\vv{x}_0$ is separated by an indicator function $I\{\vv{x}_0 \in P_0\}$. 

\subsubsection*{Terms in the squared bias part}

Let us first deal with the two terms in the squared bias part.
Observe that
\begin{align*}
  \E_{\vv{X}}(\mu|\vv{P}_{\vv{x}_0}(\Theta))) = \sum_{P_i \in \vv{P}(\Theta)} \E_{\vv{X}}(\mu | P_i) I\{\vv{x}_0 \in P_i\} = \mu_{\vv{P}(\Theta)}(\vv{x}_0).
\end{align*}
Hence, we can show that 
\begin{align*}
    \E_{\vv{X'},\Theta}[(\mu(\vv{X'}) - \E_{\vv{X}}(\mu|\vv{P}_{\vv{X'}}(\Theta)))^2] = \E_{\vv{X},\Theta}[(\mu - \mu_{\vv{P}(\Theta)})^2]
\end{align*}
and
\begin{align*}
    \E_{\vv{X'}}\left[\text{Var}_{\Theta}(\E_{\vv{X}}(\mu| \vv{P}_{\vv{X'}}(\Theta)))\right] = \E_{\vv{X'}}\left[\text{Var}_{\Theta}(\mu_{\vv{P}(\Theta)}(\vv{X'}))\right].
\end{align*}

\subsubsection*{Terms in the variance part}

Next, we focus on the two terms in the variance part.
Since the local cross-partition covariance function is related to cells $P_{\vv{x}_0}$ and $P'_{\vv{x}_0}$, we need to change to another form of the representation of these functions.  
The \textit{new} representation is given by 
\begin{align*}
    \text{Cov}_{\mu}(P,P';\vv{x}_0)
    & = \sum_{\substack{P_i \in P,\\
    P'_j\in P'}} \bigg\{\E_{\vv{X}}\left((\mu - \E_{\vv{X}}(\mu|P_i))(\mu - \E_{\vv{X}}(\mu|P'_j))|P_i \cap P'_j\right) \\
    & ~~~~~~~~~~~~~\times \frac{\P_{\vv{X}}(P_i\cap P'_j)}{\P_{\vv{X}}(P_i)\P_{\vv{X}}(P'_j)}I\{\vv{x}_0\in P_i\cap P'_j\}\bigg\},\\
    \text{Cov}_{\sigma^2}(P,P';\vv{x}_0)
    & = \sum_{\substack{P_i \in P,\\
    P'_j\in P'}} \E_{\vv{X}}(\sigma^2|P_i\cap P'_j) \frac{\P_{\vv{X}}(P_i\cap P'_j)}{\P_{\vv{X}}(P_i)\P_{\vv{X}}(P'_j)}I\{\vv{x}_0\in P_i\cap P'_j\}.
\end{align*}
In particular, when $P = P'$, the representations for the variance functions are given by 
\begin{align*}
    \text{Var}_{\mu}(P;\vv{x}_0)
    & = \sum_{\substack{P_i \in P}} \E_{\vv{X}}\left((\mu - \E_{\vv{X}}(\mu|P_i))^2|P_i \right)\frac{1}{\P_{\vv{X}}(P_i)}I\{\vv{x}_0\in P_i\},\\
    \text{Var}_{\sigma^2}(P;\vv{x}_0)
    & = \sum_{\substack{P_i \in P}} \E_{\vv{X}}(\sigma^2|P_i) \frac{1}{\P_{\vv{X}}(P_i)}I\{\vv{x}_0\in P_i\cap P'_j\}.
\end{align*}

Substituting $\vv{X'}$ into the expression above and taking the expectation, we can deduce that 
\begin{align*}
    &\E_{\vv{X'},\Theta,\Theta'}
    \left\{
    \text{Cov}_{\mu}(\vv{P}(\Theta),\vv{P}(\Theta');\vv{X'})\right\} \\
    =& \E_{\vv{X'},\Theta,\Theta'}
    \left\{\sum_{P_i \in P,
    P'_j\in P'} \Bigg\{\E_{\vv{X}}\left((\mu - \E_{\vv{X}}(\mu|P_i))(\mu - \E_{\vv{X}}(\mu|P'_j))|P_i \cap P'_j\right)\right.\\
    &~~~~~~~~~~~~~~~~~~~~~~~~~~\left.\times \frac{\P_{\vv{X}}(P_i\cap P'_j)}{\P_{\vv{X}}(P_i)\P_{\vv{X}}(P'_j)}I\{\vv{X'}\in P_i\cap P'_j\}\Bigg\}\right\} \\
    =& \E_{\Theta,\Theta'}\left\{\sum_{P_i \in P,
    P'_j\in P'} \Bigg\{\E_{\vv{X}}\left((\mu - \E_{\vv{X}}(\mu|P_i))(\mu - \E_{\vv{X}}(\mu|P'_j))|P_i \cap P'_j\right)\right.\\
    &~~~~~~~~~~~~~~~~~~~~~~~~\left.\times \frac{\P_{\vv{X}}(P_i\cap P'_j)}{\P_{\vv{X}}(P_i)\P_{\vv{X}}(P'_j)}\P_{\vv{X}}(P_i\cap P'_j)\Bigg\}\right\} \\
    =& \E_{\Theta,\Theta'}\left\{\text{Cov}_{\mu}(\vv{P}(\Theta),\vv{P}(\Theta'))\right\}. 
\end{align*}
Applying a similar technique, we can also obtain that 
\begin{align*}
    \E_{\vv{X'},\Theta,\Theta'}
    \left\{
    \text{Cov}_{\sigma^2}(\vv{P}(\Theta),\vv{P}(\Theta');\vv{X'})\right\}
    & = \E_{\Theta,\Theta'}\left\{\text{Cov}_{\sigma^2}(\vv{P}(\Theta),\vv{P}(\Theta'))\right\},\\
    \E_{\vv{X'},\Theta}
    \left\{
    \text{Var}_{\mu}(\vv{P}(\Theta);\vv{X'})\right\}
    & = \E_{\Theta,\Theta'}\left\{\text{Var}_{\mu}(\vv{P}(\Theta))\right\},\\
    \E_{\vv{X'},\Theta}
    \left\{
    \text{Var}_{\sigma^2}(\vv{P}(\Theta);\vv{X'})\right\}
    & = \E_{\Theta,\Theta'}\left\{\text{Var}_{\sigma^2}(\vv{P}(\Theta))\right\}.
\end{align*}

Finally, the remainders can be derived similarly by combining an alternative representation separating $\vv{X'}$ from the cell of partitions and the exchange of integrations. We omit the details to save space. This concludes the proof of Theorem \ref{new.thm2}.

\subsection{Proof of Theorem \ref{new.thm3}} \label{new.SecA.4}

According to the dominated convergence theorem, conditions (\ref{cond:globaluniform}) and (\ref{cond:cellprob}) imply that all the remainder terms in (\ref{eq:MSEens})  tend to zero as $n \to \infty$. Hence, the weak consistency holds \textit{if and only if} the leading terms tend to zero.

It is clear that condition $\E_{\vv{X},\Theta}[(\mu - \mu_{\vv{P}^n(\Theta)})^2] \to 0$ in (\ref{eq:partbiasconv}) entails the convergence of the second term of the squared bias parts in the first line of (\ref{eq:MSEens}) with $B = 1$. Similarly, condition $\E_{\vv{X}}[(\mu - \E_{\Theta}(\mu_{\vv{P}^n(\Theta)}))^2] \to 0$ in (\ref{eq:ensbiasconv}) together with $B\to\infty$ entails the convergence of the first term of the squared bias parts in the first line of (\ref{eq:MSEens}). Thus, it remains to consider the convergence for the variance parts.

Observe that $\mu$ and $\sigma^2$ are upper bounded so that we have 
\begin{align*}
  \text{Cov}_{\mu}(P,P') \leq \|\mu\|^2_{\infty} \text{Cov}(P,P'),~~~\text{Cov}_{\sigma^2}(P,P') \leq \|\sigma^2\|_{\infty} \text{Cov}(P,P').
\end{align*}
Meanwhile, since $\sigma^2$ is lower bounded by $\sigma_0^2$, it holds that 
$$\text{Cov}_{\sigma^2}(P,P') \geq \sigma^2_0 \text{Cov}(P,P').$$
It then follows that
\begin{align*}
    \sigma_0^2 \E_{\Theta}[\text{Var}(\vv{P}(\Theta))] 
    & \leq \E_{\Theta}[\text{Var}_{\mu}(\vv{P}(\Theta)) + \text{Var}_{\sigma^2}(\vv{P}(\Theta))] \\
    & \leq (\|\mu\|^2_\infty + \|\sigma^2\|_\infty) \E_{\Theta}[\text{Var}(\vv{P}(\Theta))],
\end{align*}
since both $\text{Var}_{\mu}$ and $\text{Var}_{\sigma^2}$ are nonnegative. In light of $\text{Var}(\vv{P}(\Theta)) = |\vv{P}(\Theta)|$,
we can immediately conclude that the variance part in the third line of (\ref{eq:MSEens}) converges to zero if and only if condition $n^{-1}\E_{\Theta}[|\vv{P}^n(\Theta)|] \to 0$ in (\ref{eq:partbiasconv}) holds.

If in addition $\E_{\Theta,\Theta'}[\text{Cov}_{\mu}(\vv{P}(\Theta), \vv{P}(\Theta'))] \geq 0$, we also have
\begin{align*}
    \sigma_0^2 \E_{\Theta,\Theta'}[\text{Cov}(\vv{P}(\Theta),\vv{P}(\Theta'))] & \leq \E_{\Theta,\Theta'}[\text{Cov}_{\mu}(\vv{P}(\Theta),\vv{P}(\Theta')) + \text{Cov}_{\sigma^2}(\vv{P}(\Theta),\vv{P}(\Theta'))] \\
    & \leq (\|\mu\|^2_\infty + \|\sigma^2\|_\infty )\E_{\Theta,\Theta'}[\text{Cov}(\vv{P}(\Theta),\vv{P}(\Theta'))].
\end{align*}
The equivalence of the convergence of the variance part for the ensemble estimator in (\ref{eq:MSEens}) and condition $n^{-1} \E_{\Theta,\Theta'}[\text{Cov}(\vv{P}^n(\Theta),\vv{P}^n(\Theta'))] \to 0$ (\ref{eq:ensbiasconv}) immediately follows.
Hence, it suffices to 
show that $\E_{\Theta,\Theta'}[\text{Cov}_{\mu}(\vv{P}(\Theta), \vv{P}(\Theta'))] \geq 0$. 

For any $\theta\in \mathcal{D}^n$, denote by $\vv{P}(\theta) = \{P^\theta_r:r=1,\ldots,R_{\theta}\}$. From the definition of the global cross-partition covariance function in (\ref{eq:globalCov}), it holds that 
\begin{align*}
    \text{Cov}_{\mu}(\vv{P}(\theta),\vv{P}(\theta')) = \sum_{r=1}^{R_\theta}\sum_{s=1}^{R_{\theta'}} \frac{\E_{\vv{X}}\left((\mu - \mu_{\vv{P}(\theta)})(\mu - \mu_{\vv{P}(\theta')})I_{P^\theta_r}I_{P^{\theta'}_s}\right)\P_{\vv{X}}(P^{\theta}_r\cap P^{\theta'}_s)}{\P_{\vv{X}}(P^{\theta}_r)\P_{\vv{X}}(P^{\theta'}_s)}.
\end{align*}
Let us define random vectors
\begin{align*}
    \vv{\alpha}(\vv{X};\theta) &= \left(\frac{(\mu(\vv{X}) - \mu_{\vv{P}(\theta)}(\vv{X}))I_{P^\theta_r}(\vv{X})}{\sqrt{\P_{\vv{X}}(P^\theta_r)}}:r=1,\ldots,R_{\theta}\right)^\top,\\
    \vv{\beta}(\vv{X};\theta) &= \left(\frac{I_{P^\theta_r}(\vv{X})}{\sqrt{\P_{\vv{X}}(P^\theta_r)}}:r=1,\ldots,R_{\theta}\right)^\top.
\end{align*}
Further, we introduce the matrix-valued bivariate functions
\begin{align*}
    \vv{A}(\theta,\theta') & =\E_{\vv{X}}\left(\vv{\alpha}(\vv{X};\theta) \vv{\alpha}(\vv{X};\theta')^\top\right) \\
    & = \left(\frac{\E_{\vv{X}}\left((\mu -\mu_{\vv{P}(\theta)})(\mu - \mu_{\vv{P}(\theta')})I_{P^\theta_r\cap P^{\theta'}_s}\right)}{\sqrt{\P_{\vv{X}}(P^\theta_r)\P_{\vv{X}}(P^{\theta'}_s)}}: r=1,\ldots,R_{\theta};s=1,\ldots,R_{\theta'}\right),\\
    \vv{B}(\theta,\theta') & =\E_{\vv{X}}\left(\vv{\beta}(\vv{X};\theta)\vv{\beta}(\vv{X};\theta')^\top\right) \\
    & = \left(\frac{\P_{\vv{X}}(P^\theta_r\cap P^{\theta'}_s)}{\sqrt{\P_{\vv{X}}(P^\theta_r)\P_{\vv{X}}(P^{\theta'}_s)}}: r=1,\ldots,R_{\theta};s=1,\ldots,R_{\theta'}\right),
\end{align*}
where $\vv{A}(\theta,\theta') = \vv{A}(\theta',\theta)^\top$ and $\vv{B}(\theta,\theta') = \vv{B}(\theta',\theta)^\top$.
Then we see that 
\[
\text{Cov}_{\mu}(\vv{P}(\theta),\vv{P}(\theta')) = \text{tr}(\vv{A}(\theta,\theta')\vv{B}(\theta',\theta)),
\]

Let $\vv{Y}$ be an independent copy of $\vv{X}$. By defining $h(\vv{X},\vv{Y};\theta) = \vv{\alpha}(\vv{X};\theta)^\top\vv{\beta}(\vv{Y};\theta)$, we can deduce that 
\begin{align*}
    \text{Cov}_{\mu}(\vv{P}(\theta),\vv{P}(\theta')) & = \text{tr}(\vv{A}(\theta,\theta')\vv{B}(\theta',\theta))\\
    & = \E_{\vv{X},\vv{Y}}\left\{\text{tr}\left(\vv{\alpha}(\vv{X};\theta)\vv{\alpha}(\vv{X};\theta')^\top \vv{\beta}(\vv{Y};\theta')\vv{\beta}(\vv{Y};\theta)^\top\right)\right\}\\
    & = \E_{\vv{X},\vv{Y}}\left\{(\vv{\alpha}(\vv{X};\theta)^\top \vv{\beta}(\vv{Y};\theta)
    (\vv{\alpha}(\vv{X};\theta')^\top \vv{\beta}(\vv{Y};\theta')\right\}\\
    & = \E_{\vv{X},\vv{Y}}\left\{h(\vv{X},\vv{Y};\theta)h(\vv{X},\vv{Y};\theta')\right\}.
\end{align*}
By Assumption \ref{assm:4}, we see that $h(\vv{X},\vv{Y};\theta)$ is a ternary measurable function. Denote by $H(\vv{X},\vv{Y}) = \E_{\Theta}[h(\vv{X},\vv{Y};\Theta)]$. An application of Fubini's theorem yields that 
\begin{align*}
    \E_{\Theta,\Theta'}[\text{Cov}_{\mu}(\vv{P}(\Theta),\vv{P}(\Theta'))] & = \E_{\vv{X},\vv{Y}}\left\{\E_{\Theta}(h(\vv{X},\vv{Y};\Theta)) \E_{\Theta'}(h(\vv{X},\vv{Y};\Theta'))\right\}\\
    & = \E_{\vv{X},\vv{Y}}\left\{H(\vv{X},\vv{Y})^2\right\} \geq 0,
\end{align*}
which verifies the non-negativeness of $\E_{\Theta,\Theta'}[\text{Cov}_{\mu}(\vv{P}(\Theta),\vv{P}(\Theta'))]$. This completes the proof of Theorem \ref{new.thm3}.

\newpage
\section{Some key lemmas and their proofs} \label{new.SecB}

\subsection{Asymptotic inverse moments involving sample counts over subsets}

\begin{lemma}[Upper bounds for inverse moments involving multinomial distributions]\label{lem:upperbound}
    Let $(N_1, \ldots, N_k) \sim \mathcal{M}(n,(p_1,\ldots,p_K))$. Then for any $a_{ij} \geq 1$, $i=1,\ldots,K$, $j=1,\ldots,m_i$, it holds that  
        \begin{align*}
        \E\left[\prod_{i=1}^K\prod_{j=1}^{m_i}\left(\frac{1}{N_i + a_{ij}}\right)\right] \lesssim \prod_{i=1}^K\prod_{j=1}^{m_i}\left(\frac{1}{np_i + a_{ij}}\right).
    \end{align*}
\end{lemma}

\begin{lemma}[Higher-order expansion of expected inverse power of a binomial variable]\label{lem:singlpower}
    Let $N\sim \mathcal{B}(n,p)$ and $r$ be a positive integer. Then for any constant $a \geq 1$, it holds that
    \begin{align*}
        0 \leq \E\left[\frac{1}{(a + N)^r}\right] - \frac{1}{(a + np)^r} 
        \lesssim \frac{1}{(a + np)^{r+1}}. 
    \end{align*}
    Further, we have 
    \begin{align*}
        \E\left[\frac{1}{(a + N)^r}\right] & =  \frac{1}{(a + np)^r}  + \frac{r(r+1) np(1 - p)}{2(a + np)^{r+2}} + R_4,
    \end{align*}
    in which the remainder satisfies
    \begin{align*}
         R_4 &\lesssim \frac{1}{(a + np)^{r + 3/2}}.
    \end{align*}
\end{lemma}

\begin{lemma}[Higher-order expansion of expected products of inverse powers for a binomial variable]\label{lem:singleproduct}
    Let $N\sim \mathcal{B}(n,p)$, and $r$ and $s$ be two nonnegative integers with $r,s \geq 1$. Then for any constants $a,b\geq 1$, it holds that
    \begin{align*}
        0 &\leq \E\left[\frac{1}{(a + N)^r (b + N)^s }\right] - \frac{1}{(a + np)^r (b + np)^s} \\
        &\lesssim \frac{1}{(a + np)^{r+1}(b + np)^s} + \frac{1}{(a + np)^{r}(b + np)^{s+1}}. 
    \end{align*}
    Further, we have 
    \begin{align*}
        \E\left[\frac{1}{(a + N)^r (b + N)^s }\right] & =  \frac{1}{(a + np)^r (b + np)^s} \\
        & + \frac{r(r+1) np(1 - p)}{2(a + np)^{r+2} (b + np)^{s}} \\
        & +  \frac{s(s+1) np(1 -p)}{2(a + np)^{r} (b + np)^{s+2}} \\
        & + \frac{rs np(1 -p )}{(a + np)^{r+1} (b + np)^{s+1}}\\
        & + R_5,
    \end{align*}
    in which the remainder satisfies
    \begin{align*}
         R_5 &\lesssim \frac{1}{(a + np)^{r + 3/2} (b + np)^{s}} + \frac{1}{(a + np)^{r} (b + np)^{s + 3/2}} \\
        & + \frac{1}{(a + np)^{r + 1} (b + np)^{s + 1/2}} + \frac{1}{(a + np)^{r + 1/2} (b + np)^{s + 1}}.
    \end{align*}
\end{lemma}

\begin{lemma}[Expansion of inverse moments of the product of sample counts without overlapping]\label{lem:doubledisjoint}
Let $A_1$ and $A_2$ be two disjoint, non-empty subset with positive probability $p_1 = \P_{\vv{X}}(A_1)$ and $p_2 = \P_{\vv{X}}(A_2)$. Denote by $N_i = \sum_{i=1}^n I\{\vv{X}_i \in P_i\}$ for $i=1,2$. Then for any constants $a,b\geq 1$, it holds that 
\begin{align*}
    \E\left[\frac{1}{(a + N_1) (b + N_2)}\right] & =  \frac{1}{(a + np_1) (b + np_2)} \\
    & + \frac{np_1(1-p_1)}{(a + np_1)^3(b+np_2)}\\
    & + \frac{np_2(1-p_2)}{(a + np_1)(b+np_2)^3}\\
    & - \frac{np_1p_2}{(a + np_1)^2(b+np_2)^2} + R_6,
\end{align*}
in which the remainder satisfies 
\begin{align*}
    R_6 & \lesssim  \frac{1}{(a + np_1)^{5/2} (b + np_2)} + \frac{1}{(a + np_1)(b + np_2)^{5/2}}
    \\
    & + \frac{1}{(a + np_1)^2 (b + np_2)^{3/2}} + \frac{1}{(a + np_1)^{3/2}(b + np_2)^2}.
\end{align*}
In particular, we have 
\begin{align*}
    0 &\leq \E\left[\frac{1}{(a + N_1) (b + N_2)}\right] -  \frac{1}{(a + np_1) (b + np_2)} \\
    &\lesssim \frac{1}{(a + np_1)(b + np_2)}\left(\frac{1}{a + np_1} + \frac{1}{b + np_2}\right).
\end{align*}
\end{lemma}

\begin{lemma}[Expansion of inverse moments of the product of sample counts with overlapping]\label{lem:6}
    Let $P$ and $P'$ be two  subsets in $\mathcal{X}^d$ with a non-empty intersection $P_0 = P \cap P' \neq\varnothing$ having positive probability $p_0 = \P(\vv{X} \in P_0)$. Define 
    $p = \P(\vv{X} \in A)$ and $p' = \P(\vv{X}\in A)$. For a given sample $\{\vv{X}_i\}_{1\leq i\leq n}$, denote by  
    $N = \sum_{i=1}^n I\{\vv{X}_i \in P\}$ and $N'\sum_{i=1}^n I\{\vv{X}_i \in P'\}$ the sample counts of $P$ and $P'$, respectively.
    Then for any constant $a \geq 1$, it holds that
    \begin{equation*}
        \begin{split}
            \E\left[\frac{1}{(a + N)(a + N')}\right]
            &= \frac{1}{(a + np)(a + np')}\left(1 + \frac{1 - p}{a + np}\right) \left(1 + \frac{1 - p'}{a + np'}\right)\\
            & - \frac{n(p_0 - pp')}{(a + np)^2(a +np')^2} + R_7,
        \end{split}
    \end{equation*}
    where the remainder satisfies
    \begin{align*}
        R_7 & \lesssim  \frac{1}{(a + np)^{5/2} (a +np')} + \frac{1}{(a + np)(a + np')^{5/2}}\\
    &+\frac{1}{(a + np)^{2} (a + np')^{3/2}} + \frac{1}{(a +np)^{3/2} (a +np')^{2}}.
    \end{align*}
    In particular, we have 
\begin{align*}
    0 &\leq \E\left[\frac{1}{(a + N) (a + N')}\right] -  \frac{1}{(a + np) (a + np')} \\
    &\lesssim \frac{1}{(a + np)(a + np')}\left(\frac{1}{a + np} + \frac{1}{b + np'}\right).
\end{align*}
\end{lemma}

The results displayed in Lemmas \ref{lem:upperbound}--\ref{lem:6} above are inspired by those in Lemmas 3 and 7 of \cite{Klusowski2024}, and \textit{extend} the scope of these findings by providing a more refined and comprehensive nonasymptotic framework.  
Specifically, we address the \textit{higher-order} inverse moments for binomial distributions and also cover the cases related to the multinomial distributions.

\subsection{Proof of Lemma \ref{keylem1}} \label{new.SecB.1}

Assume that $N = \sum_{i=1}^n I_i$, where $\{I_j\}$ is a sequence of i.i.d. Bernoulli random variables with a success probability $p$. 
To examine the asymptotic expansion of the inverse moment of $N$, we adopt the \textit{leave-one-out technique} with the following observation
\begin{align*}
    \E\left\{\frac{I\{N \geq 1\}}{N}\right\} & = \E\left\{\frac{N}{N^2}I\{N\geq 1\}\right\} \\
    & = \sum_{i=1}^n \E\left\{\frac{I_i}{N^2}I\{N\geq 1\}\right\} = \sum_{i=1}^n \E\left\{\frac{I_i}{N^2}\right\}\\
    & = n\E\left\{\frac{I_1}{(I_1 + N_{-1})^2}\right\},
\end{align*}
in which $N_{-1} = \sum_{j=2}^n I_j \sim \mathcal{B}((n-1),p)$ and is independent of $I_1$. Hence, it holds that 
\[
  \E\left\{\frac{I\{N \geq 1\}}{N}\right\} = n\E\left\{\frac{I_1}{(I_1 + N_{-1})^2}\right\} = np \E\left\{\frac{1}{(1 + N_{-1})^2}\right\}.
\]
With an application of Lemma \ref{lem:singlpower} with $r=2$, we can obtain that 
\[
0 \leq \E\left\{\frac{1}{(1 + N_{-1})^2}\right\} - \frac{1}{(1 + (n-1)p)^2} \lesssim \frac{1}{(1 + (n-1)p)^3}.
\]

Consequently, we have that 
\begin{align*}
    0 \leq \E\left\{\frac{I\{N \geq 1\}}{N}\right\} - \frac{np}{(1 + (n-1)p)^2} \lesssim \frac{np}{(1 + (n-1)p)^3}.
\end{align*}
Then it immediately follows that
\begin{align*}
    \left|\E\left\{\frac{I\{N \geq 1\}}{N}\right\} - \frac{1}{np}\right| & \lesssim \frac{np}{(1 + (n-1)p)^3} + \left|\frac{1}{np} - \frac{np}{(1 + (n-1)p)^2}\right|\\
    & = \frac{np}{(1 + (n-1)p)^3}
    + \frac{2(1-p)}{(1 + (n-1)p)^2}
    + \frac{(1-p)^2}{np(1 + (n-1)p)^2}\\
    & \lesssim \left(1 + \frac{1}{np}\right) \frac{1}{(1 + (n-1)p)^2},
\end{align*}
which concludes the proof of Lemma \ref{keylem1}.

\subsection{Proof of Lemma \ref{keylem2}} \label{new.SecB.2}

\subsubsection*{Proof of conclusion (1)}
Let us first focus on justifying bound (\ref{lemma2.1}) of Lemma \ref{keylem2}.
Denote by 
\[
N_{-1} = \sum_{i=2}^n I\{\vv{X}_i \in P\}~~~\hbox{and}~~~
N'_{-1} = \sum_{i=2}^n I\{\vv{X}_i \in P'\}.
\]
Then it holds that 
\[
\frac{I\{\vv{X}_1 \in P\cap P'\}}{N N'} = \frac{I\{\vv{X}_1 \in P\cap P'\}}{(1 + N_{-1})(1 + N'_{-1})}.
\]

Recall that $N_0 = \sum_{i=1}^n I\{\vv{X}_i \in P\cap P'\}$.
We exploit the leave-one-out technique to deduce that 
\begin{align*}
\E\left[\frac{N_0}{N N'}\right]
=& \sum_{i=1}^n \E\left[\frac{I\{\vv{X}_i \in P \cap P'\}}{N N'}\right] \\
=& n \E\left[\frac{I\{X_1 \in P\cap P'\}}{(1 + N_{-1})(1 + N'_{-1})}\right]\\
=& np_0 \E\left[\frac{1}{(1 + N_{-1})(1 + N'_{-1})}\right],
\end{align*}
where the last step above is because $\vv X_1$ is independent of $N_{-1}$ and $N'_{-1}$.

An application of Lemma \ref{lem:6} with $a = 1$ leads to 
    \begin{equation}\label{eq:bound6}
        \begin{split}
            \E\left[\frac{1}{(a + N_{-1})(a + N'_{-1})}\right] & = \left\{
    \frac{1}{a + (n-1)p}
    \left(1 + \frac{1 - p}{a + (n-1)p}\right)\right\}\\
    &\times 
    \left\{
    \frac{1}{a + (n-1)p'}
    \left(1 + \frac{1 - p'}{a + (n-1)p'}\right)\right\} + R_{11},
        \end{split}
    \end{equation}
    in which the remainder satisfies 
    \begin{align*}
        R_{11} & \lesssim  \frac{1}{(1 + (n-1)p)^{2} (1 +(n-1)p')} + \frac{1}{(1 + (n-1)p)(1 + (n-1)p')^{2}}.
    \end{align*}

Observe that 
\begin{equation}\label{eq:recur}
\begin{split}
    \frac{1}{1 + (n-1)p}\left\{1 + \frac{1}{1 + (n-1)p}\right\} & = 
    \frac{1}{np} - \frac{(1 - p)^2}{np(1 + (n-1)p)^2}.
 \end{split}
\end{equation}
Then the leading term in (\ref{eq:bound6}) becomes
\begin{align*}
    &\left\{
    \frac{1}{a + (n-1)p}
    \left(1 + \frac{1 - p}{a + (n-1)p}\right)\right\} 
    \left\{
    \frac{1}{a + (n-1)p'}
    \left(1 + \frac{1 - p'}{a + (n-1)p'}\right)\right\}\\
    = &\left\{\frac{1}{np} - \frac{(1 - p)^2}{np(1 + (n-1)p)^2}\right\}\left\{\frac{1}{np'} - \frac{(1 - p')^2}{np(1 + (n-1)p')^2}\right\}\\
    = &\frac{1}{n^2 pp'} + R_{12},
\end{align*}
in which the remainder $R_{12}$ satisfies
\begin{align*}
    R_{12} \lesssim \frac{1}{n^2 p p'}\left(\frac{1}{(1 + (n-1)p)^2} + \frac{1}{(1 + (n-1)p')^2}\right).
\end{align*}

Combining the expressions above, we can obtain that 
\begin{align*}
     \E\left[\frac{N_0}{NN'}\right] & = np_0 \E\left[\frac{1}{(1 + N_{-1})(1 + N'_{-1})}\right] \\
     & = \frac{p_0}{npp'} + np_0(R_{11} + R_{12}),
\end{align*}
where the remainder is controlled by
\begin{align*}
    R_{21} = np_0(R_{11} + R_{12}) \lesssim \frac{p_0}{npp'}\left(\frac{1}{1 + (n-1)p} + \frac{1}{1 + (n-1)p'}\right).
\end{align*}

\subsubsection*{Proof of conclusion (2)}
It is easy to see that
\[
\frac{N_1}{N} = \left(1 - \frac{N_0}{N}\right)I\{N\geq 1\} \ \text{ and } \ \frac{N_1}{N'} = \left(1 - \frac{N_0}{N'}\right)I\{N'\geq 1\}.
\]
The left-hand side (LHS) in (\ref{lemma2.2}) can be decomposed as 
\begin{align*}
     &\E\left[\left(\alpha \frac{N_0}{N} + \beta \frac{N_1}{N}\right) \left(\alpha \frac{N_0}{N'} + \gamma \frac{N_1}{N'}\right) \right]\\
    = &\E\left[\left(\alpha \frac{N_0}{N} + \beta\left(1 - \frac{N_0}{N}\right)\right)\left(\alpha \frac{N_0}{N'} + \gamma\left(1 - \frac{N_0}{N'}\right)\right)I\{N\geq 1,N'\geq 1\}\right] \\
    = & (\alpha - \beta)(\alpha - \gamma) \E\left[\frac{N_0^2}{NN'}\right] \\
    + & \gamma (\alpha - \beta)\E\left[\frac{N_0}{N}\right] 
    +  \beta (\alpha - \gamma)\E\left[\frac{N_0}{N'}\right] 
    \\
    + & \beta\gamma\P(N \geq 1, N' \geq 1)\\
    = :& J_1 + J_2 + J_3 + J_4. 
\end{align*}

Observe that
\[
        \P(N \geq 1, N' \geq 1) = 1- \P(\min\{N, N'\} = 0) 
         \geq 1 - (1 - p)^n - (1 - p')^n.
\]
We immediately see that 
\begin{equation}\label{eq:bd3}
    \begin{split}
        J_4 &:= \beta \gamma\P(N \geq 1, N' \geq 1) = \beta\gamma + R_{23}
    \end{split}
\end{equation}
with $R_{23} \leq |\beta||\gamma|((1 - p)^n + (1 - p')^n)$.

Next, for term $J_2$ we can apply the leave-one-out technique to show that 
\begin{align*}
   \E\left[\frac{N_0}{N}\right]& = \sum_{i=1}^n \E\left[\frac{I\{X_i \in P \cap P'\}}{N}\right] 
     = n p_0 \E\left[\frac{1}{1+ N_{-1}}\right].
\end{align*}
An application of Lemma \ref{lem:singlpower} with $r=1$ and $a=1$ results in 
\begin{equation*}
\begin{split}
        \E\left\{\frac{1}{1 + N_{-1}}\right\} & =  
        \frac{1}{1 + (n-1)p}\left(1 + \frac{1 -p}{1 + (n-1)p}\right) + R_{24}, 
\end{split}
\end{equation*}
where the remainder satisfies $R_{24} \lesssim 1/(1 + (n-1)p)^{5/2}$.
Notice that
\begin{align*}
    \frac{1}{1 + (n-1)p} \left(1 + \frac{1 - p}{1 + (n-1)p}\right)
    = \frac{1}{np} - \frac{(1 - p)^2}{np(1 + (n-1)p)^2}.
\end{align*}
It holds that 
\begin{equation*}
    \begin{split}
        \E\left[\frac{N_0}{N}\right] & = \frac{p_0}{p} + R_{25} ~~~\hbox{with}~ R_{25} \lesssim \frac{p_0}{p(1 + (n-1)p)^{3/2}}.
    \end{split}
\end{equation*}

Similarly, by replacing $N$ and $p$ with $N'$ and $p'$, respectively, the same assertion still holds. Consequently, we have that 
\begin{equation}\label{eq:bd1}
\begin{split}
    J_{2} + J_3 & = \gamma (\alpha - \beta) \E\left[\frac{N_0}{N}\right] + \beta(\alpha - \gamma) \E\left[\frac{N_0}{N'}\right]\\
    & = \gamma (\alpha - \beta)\frac{p_0}{p} +  \beta(\alpha - \gamma)\frac{p_0}{p'} + R_{26},
\end{split}
\end{equation}
where the remainder term satisfies
\begin{align*}
    R_{26} &\lesssim \max\{|\alpha|,|\beta|,|\gamma|\}^2 \left(\frac{1}{(1 + (n-1)p)^{3/2}} + \frac{1}{(1 + (n-1)p')^{3/2}}\right).
\end{align*}

For term $J_1$, we still adopt the leave-one-out technique and deduce that 
\begin{align*}
    \E\left[\frac{N_0^2}{NN'}\right]
    = & \sum_{i,j=1}^n \E\left[\frac{I\{\vv{X}_i,\vv{X}_j \in P \cap P'\}}{NN'}\right]\\
    =&\sum_{i=1}^n \E\left[\frac{I\{\vv{X}_i\in P \cap P'\}}{NN'}\right]\\
    +& \sum_{1\leq i\neq j\leq n} \E\left[\frac{I\{\vv{X}_i,\vv{X}_j \in P \cap P'\}}{NN'}\right]\\
    =& n p_0 \E\left[\frac{1}{(1+N_{-1})(1+ N'_{-1})}\right] \\
    +&n(n-1) p_0^2 \E\left[\frac{1}{(2 +N_{-2})(2+N_{-2})}\right]\\
    =:& J_{11} + J_{12},
\end{align*}
in which $N_{-2} = \sum_{i=3}^n I\{\vv{X}_i \in P\}$ and $N'_{-2} = \sum_{i=3}^n I\{\vv{X}_i \in P'\}$.

According to the discussion in conclusion (1), it holds that 
\begin{equation}\label{eq:boundJ11}
    \begin{split}
        J_{11} &= \frac{p_0}{npp'} + R_{21} ~~~\hbox{with}~R_{21}\lesssim \frac{p_0}{npp'}\left(\frac{1}{1 + (n-1)p} + \frac{1}{1+(n-1)p'}\right). 
    \end{split}
\end{equation}
Meanwhile, note that
\begin{align*}
    & \frac{1}{2 + (n-2) p}\left\{1 + \frac{1 - p}{2 + (n-2) p}\right\} \\
    = & \frac{1}{1 + (n-1) p} - \frac{(1 - p)^2}{(1 + (n-1)p)(2 + (n-2) p)^2}.
\end{align*}
Then we can show that 
\begin{align*}
    &\left\{\frac{1}{2 + (n-2) p}\left\{1 + \frac{1 - p}{2 + (n-2) p}\right\}\right\}\\
    \times&\left\{\frac{1}{2 + (n-2) p'}\left\{1 + \frac{1 - p'}{2 + (n-2) p'}\right\}\right\}\\
   =&\left\{\frac{1}{1 + (n-1) p} - \frac{(1 - p)^2}{(1 + (n-1)p)(2 + (n-2) p)^2}\right\}\\
   \times& \left\{\frac{1}{1 + (n-1) p'} - \frac{(1 - p')^2}{(1 + (n-1)p')(2 + (n-2) p')^2} \right\}\\
   =& \frac{1}{(1 + (n-1)p)(1 + (n-1)p')} + R_{27},
\end{align*}
where the remainder $R_{27}$ satisfies
\begin{align*}
    R_{27} \lesssim \frac{1}{(1 + (n-1)p)(1 + (n-1)p')}\left(\frac{1}{(1+ (n-1)p)^2} + \frac{1}{(1 + (n-1)p')^2}\right).
\end{align*}

Applying Lemma \ref{lem:6}, we can obtain that 
\begin{align*}
    \E\left[\frac{1}{(2 +N_{-2})(2+N_{-2})}\right]
    =& \frac{1}{(1 + (n-1)p)(1 + (n-1)p')} \\
    +& \frac{(n-1)(p_0 - pp')}{(2 + (n-2)p)^2(2 + (n-2)p')^2} + R_{27}.
\end{align*}
To further simplify the leading terms, let us observe that
\begin{align*}
    &\frac{1}{(1 + (n-1)p)(1 + (n-1)p')} \\
    =& \frac{1}{n(n-1)pp'} - \frac{n(p' + p - pp')}{n(n-1)pp'(1 + (n-1)p)(1 + (n-1)p')}\\
    -&\frac{(1 - p)(1 - p')}{n(n-1)pp'(1 + (n-1)p)(1 + (n-1)p')}
\end{align*}
and
\begin{align*}
    & \frac{1}{(1 + (n-1)p)(1 + (n-1)p')}\\
    =& \frac{1}{n^2 p p'} - \frac{1 - p'}{n^2 p p' (1 + (n-1)p')} 
    -\frac{1 - p}{n^2 pp'(1 + (n-1)p)}\\
    +&\frac{(1 - p)(1 -p')}{n^2pp'(1 + (n-1)p)(1 + (n-1)p')}.
\end{align*}

Combining the above results together, it follows that 
\begin{align*}
     \frac{1}{(1 + (n-1)p)(1 + (n-1)p')} 
    =\frac{1}{n(n-1)pp'} - \frac{p' + p - pp'}{n^2(n-1)(pp')^2} + R_{28},
\end{align*}
where the remainder satisfies 
\begin{align*}
    R_{28} &\lesssim \frac{p+p'-pp'}{n^2(n-1)(pp')^2}\left\{\frac{1}{1 + (n-1)p} + \frac{1}{1 + (n-1)p'}\right\}\\
    & + \frac{1}{n(n-1)pp'(1 + (n-1)p)(1 + (n-1)p')}.
\end{align*}

Moreover, it holds that 
\begin{align*}
    &\frac{1}{(2 + (n-2)p)(2 + (n-2)p')} \\
    =& \frac{1}{(n-1)^2 pp'} - \frac{2 - p'}{(n-1)^2 pp'(2+(n-2)p')} \\
    -& \frac{2 - p}{(n-1)^2 pp'(2+(n-2)p)} \\
    + &\frac{(2 -p)(2 -p')}{(n-1)^2 pp' (2 + (n-2)p)(2 + (n-2)p')}
\end{align*}
and 
\begin{align*}
    &\frac{1}{(2 + (n-2)p)(2 + (n-2)p')} \\
    =& \frac{1}{n^2 pp'} - \frac{2(1 - p')}{n^2 pp'(2+(n-2)p')} \\
    -& \frac{2(1 - p)}{n^2 pp'(2+(n-2)p)} \\
    + &\frac{4(1 -p)(1 -p')}{n^2 pp' (2 + (n-2)p)(2 + (n-2)p')}.
\end{align*}
A combination of the results above gives 
\begin{align*}
    \frac{(n-1)(p_0 - pp')}{(2 + (n-2)p)^2(2 + (n-2)p')^2}
    = \frac{p_0 - pp'}{n^2(n-1)(pp')^2} + R_{29},
\end{align*}
where the remainder satisfies
\begin{align*}
    R_{29} & \lesssim \frac{p_0 - pp'}{n^2(n-1)(pp')^2}\left(\frac{1}{2 + (n-2)p} + \frac{1}{2 + (n-2)p'}\right).
\end{align*}

In summary, we now have
\begin{align*}
    \E\left[\frac{1}{(2 + N_{-2})(2 + N'_{-2})}\right] & = \frac{1}{n(n-1)pp'} + \frac{p_0 - p - p'}{n^2(n-1)(pp')^2} + R_{27} + R_{28} + R_{29}
\end{align*}
and thus, 
\begin{equation}\label{eq:boundJ12}
    \begin{split}
        J_{12} & = n(n-1) p_0^2 \E\left[\frac{1}{(2 +N_{-2})(2+N_{-2})}\right]\\
        & = \frac{p_0^2}{pp'} + \frac{p_0^2(p_0 - p - p')}{n(pp')^2} + R_{210},
    \end{split}
\end{equation}
where the remainder satisfies 
\begin{align*}
    R_{210} &= n(n-1)p_0^2(R_{27} + R_{28} + R_{29}) \\
    &\lesssim \frac{p_0^2(p + p' - p_0)}{n(pp')^2}\left\{\frac{1}{1 + (n-1)p} + \frac{1}{1+ (n-1)p'}\right\}\\
    &+ \frac{p_0^2}{pp'(1 + (n-1)p)(1 + (n-1)p')}\\
    & \lesssim \frac{p_0}{npp'}\left\{\frac{1}{1 + (n-1)p} + \frac{1}{1+ (n-1)p'}\right\}.
\end{align*}

Then substituting (\ref{eq:boundJ11}) and (\ref{eq:boundJ12}) into the expression, we can conclude that
\begin{equation}\label{eq:bd2}
\begin{split}
    J_1 = \E\left[\frac{N_0^2}{NN'}\right]
    &=  \frac{p_0^2}{pp'} + \frac{p_0^2(p_0 - p - p')}{n(pp')^2} + \frac{p_0}{npp'}
     + R_{21} + R_{210}\\
    & = \frac{p_0^2}{pp'} + \frac{p_0}{npp'}\left(1 - \frac{p_0(p + p' - p_0)}{pp'}\right) + R_{21} + R_{210}\\
    & = \frac{p_0^2}{pp'} + \frac{p_0}{npp'}\left(\frac{p_1p_2}{pp'}\right) + R_{21} + R_{210},
    \end{split}
\end{equation}
where the remainder satisfies 
\begin{align*}
    R_{21} + R_{210} & \lesssim \frac{p_0}{npp'}\left\{\frac{1}{1 + (n-1)p} + \frac{1}{1+ (n-1)p'}\right\}.
\end{align*}

Finally, combining bounds (\ref{eq:bd3}), (\ref{eq:bd1}), and (\ref{eq:bd2}), we can deduce that 
\begin{align*}
    &\E_{\vv{Z}}\left[\left(\alpha \frac{N_0}{N} + \beta \frac{N_1}{N}\right) \left(\alpha \frac{N_0}{N'} + \gamma \frac{N_1}{N'}\right) \right]\\
   = & (\alpha - \beta) (\alpha - \gamma) \left\{\frac{p_0^2}{pp'} + \frac{p_0}{n p p'}\left(\frac{p_1p_2}{pp'}\right)\right\} \\
   + & \gamma (\alpha -\beta) \frac{p_0}{p} + \beta (\alpha - \gamma) \frac{p_0}{p'} + \beta\gamma + R_{22}\\
    = & \left(\alpha \frac{p_0}{p} + \beta \frac{p_1}{p}\right)
             \left(\alpha \frac{p_0}{p'} + \gamma \frac{p_2}{p'}\right)\\
             + & (\alpha - \beta)(\alpha - \gamma) \left(\frac{p_1p_2}{pp'}\right)\left(\frac{p_0}{n p p'}\right) + R_{22}, 
\end{align*}
where the remainder satisfies
\begin{align*}
    R_{22} \lesssim \max\{|\alpha|,|\beta|,|\gamma|\}^2 &\Bigg\{\frac{p_0}{npp'}\left\{\frac{1}{1 + (n-1)p} + \frac{1}{1+ (n-1)p'}\right\}\\
    &+ \frac{1}{(1 + (n-1)p)^{3/2}} + \frac{1}{(1 + (n-1)p')^{3/2}}\\
    &+(1 - p)^n + (1 - p')^n\Bigg\}.
\end{align*}
This completes the proof of Lemma \ref{keylem2}.

\subsection{Proof of Lemma \ref{lem:upperbound}}
    For each $i = 1,\ldots,K$, since $N_i \sim \mathcal{B}(n,p_i)$, by Lemma 3 in \cite{cattaneo2024inferencemondrianrandomforests} we have
    \begin{align*}
        \E\left[\left(\prod_{j=1}^{m_i} \frac{1}{N_i + a_{ij}}\right)^K\right] \lesssim \left(\prod_{j=1}^{m_i} \frac{1}{np_i + a_{ij}}\right)^{K}.
    \end{align*}
    It then follows from the H\"{o}lder inequality that
    \begin{align*}
        \E\left[\prod_{i=1}^K\prod_{j=1}^{m_i}\left(\frac{1}{N_i + a_{ij}}\right)\right] &\leq \prod_{i=1}^K \left\{\E\left(\prod_{j=1}^{m_i} \frac{1}{N_i + a_{ij}}\right)^K\right\}^{1/K}\\
        &\lesssim \prod_{i=1}^K \prod_{j=1}^{m_i} \frac{1}{np_i + a_{ij}},
    \end{align*}
    which proves the conclusion. This concludes the proof of Lemma \ref{lem:upperbound}.

\subsection{Proof of Lemma \ref{lem:singlpower}}
    Let us define an auxiliary function $J(x) = 1/(a + x)^r$. The derivatives of function $J$ are given by 
    \begin{align*}
        J'(x) = -\frac{r}{(a + x)^{r+1}},~~~
        J''(x) = \frac{r(r+1)}{(a + x)^{r+2}},~~~
        J'''(x) = -\frac{r(r+1)(r+2)}{(a+x)^{r+3}}.
    \end{align*}
    By resorting to Taylor's theorem, we see that 
    \begin{align*}
        J(N) & = J(np) + (N - np) J'(np) + \frac{1}{2}(N - np)^2J''(np)\\
        & + \frac{1}{6} (N - np)^3J'''(\theta N + (1-\theta) np),
    \end{align*}
    where $\theta$ is an unknown random parameter in $(0,1)$. Observe that $-J'''$ is a positive convex function. Hence, it holds that 
    \[
    0 \leq R_4 :=-J'''(\theta N + (1-\theta) np) \leq \frac{r(r+1)(r+2)}{(a + N)^{r+3}} + \frac{r(r+1)(r+2)}{(a + np)^{r+3}}.
    \]
    
    Further, we can show that 
    \begin{align*}
        R_4& := \frac{1}{6} \E\left|(N - np)^3J'''(\theta N + (1-\theta) np)\right| \\
        \lesssim &\sqrt{\E[(N -np)^6]} \times \sqrt{\E\left( \frac{r(r+1)(r+2)}{(a + N)^{r+3}} + \frac{r(r+1)(r+2)}{(a + np)^{r+3}}\right)^2}\\
        \lesssim & \sqrt{(1 + np)^3} \times \frac{1}{(a + np)^{r+3}} \\
         \leq & \frac{1}{(a + np)^{r+ 3/2}}.
    \end{align*}
    The final assertion follows by taking the expectation with respect to $N$ on both sides of the Taylor expansion, which completes the proof of Lemma \ref{lem:singlpower}.

\subsection{Proof of Lemma \ref{lem:singleproduct}}
    We define an auxiliary function
    \[
    F(x) = \frac{1}{(a + x)^r (b + x)^s}.
    \]
    Note that the derivatives of function $F(x)$ are given by 
    \begin{align*}
        F'(x) & = - \frac{r}{(a + x)^{r+1} (b + x)^{s}} -  \frac{s}{(a + x)^{r} (b + x)^{s+1}},\\
        F''(x) & =  \frac{r(r+1)}{(a + x)^{r+2} (b + x)^{s}} +  \frac{s(s+1)}{(a + x)^{r} (b + x)^{s+2}} \\
        & + \frac{2rs}{(a + x)^{r+1} (b + x)^{s+1}},\\
        F'''(x) & = - \frac{r(r+1)(r+2)}{(a + x)^{r+3} (b + x)^{s}} - \frac{3sr(r+1)}{(a + x)^{r+2} (b + x)^{s+1}}\\
        &- \frac{s(s+1)(s+2)}{(a + x)^{r} (b + x)^{s+3}} - \frac{3r s(s+1)}{(a + x)^{r+1} (b + x)^{s+2}}\\
        & =: - (F'''_1(x) + F'''_2(x) + F'''_3(x) + F'''_4(x)).
    \end{align*}
    
    Exploiting Talyor's theorem, we can expand function $F(N)$ around $np$ and obtain 
    \begin{align*}
        F(N) & = F(np) + (N - np) F'(np) \\
        & + \frac{1}{2}(N - np)^2 F''(np) + R_5,
    \end{align*}
    in which the remainder $R_5$ depends on an unknown random parameter $\theta = \theta (N,np) \in(0,1)$ and satisfies
    \begin{align*}
        R_5 & = \frac{1}{6} (N - np)^3 F'''(\theta N + (1 - \theta)np)\\
        & = - \frac{1}{6} (N - np)^3 (F'''_1 + F'''_2 + F'''_3 + F'''_4) (\theta N + (1-\theta) np) \\
        & =: R_{51} + R_{52} + R_{53} + R_{54}.
    \end{align*}
    Hence, it suffices to establish $\theta$-independent bounds for these four terms above. 

    Notice that $F'''_1(x)$ can be written as a product of two positive and convex functions $r(r+1)(r+2)/(a + x)^{r+3}$ and $1/(b + x)^s$. Due to the convexity and positivity, it holds that 
    \begin{align*}
        \frac{1}{(a + \theta N + (1 - \theta) np)^{r + 3}} &\leq \frac{1}{ (a + N)^{r+3}} + \frac{1}{(a + np)^{r+3}},\\
        \frac{1}{(b + \theta N + (1 - \theta) np)^{s}} &\leq \frac{1}{ (b + N)^{s}} + \frac{1}{(b + np)^{s}}
    \end{align*}
    and thus, 
    \begin{align*}
        &F'''_1(\theta N + (1 -\theta) np) \\
        \leq& r(r+1)(r+2)\left(\frac{1}{ (a + N)^{r+3}} + \frac{1}{(a + np)^{r+3}}\right)
        \left(\frac{1}{ (b + N)^{s}} + \frac{1}{(b + np)^{s}}\right),
    \end{align*}
    of which the RHS is independent of $\theta$. 
    
    It then follows from the Cauchy--Schwartz inequality that
    \begin{align*}
        \E|R_{51}| &\lesssim \E\left|(N - np)^3 \left(\frac{1}{ (a + N)^{r+3}} + \frac{1}{(a + np)^{r+3}}\right)
        \left(\frac{1}{ (b + N)^{s}} + \frac{1}{(b + np)^{s}}\right)\right| \\
        & \lesssim \sqrt{\E[(N - np)^6]}\times \sqrt{\E\left[\left(\frac{1}{ (a + N)^{r+3}} + \frac{1}{(a + np)^{r+3}}\right)^2
        \left(\frac{1}{ (b + N)^{s}} + \frac{1}{(b + np)^{s}}\right)^2\right]}\\
        & \lesssim \sqrt{(1 + np)^3} \times \frac{1}{(a + np)^{r+3} (b + np)^s} \\
        & = \frac{1}{(a + np)^{r + 3/2} (b + np)^{s}}.
    \end{align*}
    Thanks to the similarity in structures, we can obtain that 
    \begin{align*}
        \E|R_{53}| & = \E\left|\frac{1}{6}(N - np)^3 F'''_3(\theta N + (1 - \theta) np)\right|\\
        & \lesssim \frac{1}{(a + np)^{r} (b + np)^{s + 3/2}}.
    \end{align*}

    We next focus on the term in $F'''_2$. Observe that $F'''_2$ can also be decomposed into a product of convex and positive functions $r(r+1)/(a + x)^{r+2}$ and $3s/(b + x)^s$. Then we have that 
        \begin{align*}
        &F'''_2(\theta N + (1 -\theta) np) \\
        \leq& 3sr(r+1)\left(\frac{1}{ (a + N)^{r+2}} + \frac{1}{(a + np)^{r+2}}\right)
        \left(\frac{1}{ (b + N)^{s+1}} + \frac{1}{(b + np)^{s+1}}\right),
    \end{align*}
    of which the RHS is independent of $\theta$.
    An application of the Cauchy--Schwartz inequality yields that 
    \begin{align*}
        \E|R_{52}| &\lesssim \E\left|(N - np)^3 \left(\frac{1}{ (a + N)^{r+2}} + \frac{1}{(a + np)^{r+2}}\right)
        \left(\frac{1}{ (b + N)^{s+1}} + \frac{1}{(b + np)^{s+1}}\right)\right| \\
        & \lesssim \sqrt{\E[(N - np)^6]}\times \sqrt{\E\left[\left(\frac{1}{ (a + N)^{r+2}} + \frac{1}{(a + np)^{r+2}}\right)^2
        \left(\frac{1}{ (b + N)^{s+1}} + \frac{1}{(b + np)^{s+1}}\right)^2\right]}\\
        & \lesssim \sqrt{(1 + np)^3} \times \frac{1}{(a + np)^{r+2} (b + np)^{s+1}} \\
        & = \frac{1}{(a + np)^{r + 1} (b + np)^{s + 1/2}}.
    \end{align*}
    
    It follows from the similarity in structures that 
    \begin{align*}
        \E|R_{54}| & = \E\left|\frac{1}{6}(N - np)^3 F'''_4(\theta N + (1 - \theta) np)\right|\\
        & \lesssim \frac{1}{(a + np)^{r + 1/2} (b + np)^{s + 1}}.
    \end{align*}
    Therefore, combining the results above, we can deduce that 
    \begin{align*}
        \E|R_5| & \lesssim \frac{1}{(a + np)^{r + 3/2} (b + np)^{s}} + \frac{1}{(a + np)^{r} (b + np)^{s + 3/2}} \\
        & + \frac{1}{(a + np)^{r + 1} (b + np)^{s + 1/2}} + \frac{1}{(a + np)^{r + 1/2} (b + np)^{s + 1}}.
    \end{align*}
    The final assertion follows by integrating out $N$ in the Taylor expansion, which concludes the proof of Lemma \ref{lem:singleproduct}.

\subsection{Proof of Lemma \ref{lem:doubledisjoint}}

Let us define the auxiliary function as
    \[
      H(x,y) = \frac{1}{(a + x) (b + y)}.
    \]
It is easy to verify that the derivatives of $H$ with respect to $x$ and $y$ are given by 
\begin{align*}
    H_x(x,y) & = -\frac{1}{ (a + x)^2 (b + y)},\\
    H_y(x,y) & = -\frac{1}{ (a + x) (b + y)^2},\\
    H_{xx}(x,y) & = \frac{2}{ (a + x)^3 (b + y)},\\
    H_{yy}(x,y) & = \frac{2}{ (a + x) (b + y)^3},\\
    H_{xy}(x,y) & = H_{yx}(x,y) \\
    & = \frac{1}{(a + x)^2 (b + y)^2},\\
    H_{xxx}(x,y) & = -\frac{6}{ (a + x)^4 (b + y)},\\
    H_{yyy}(x,y) & = -\frac{6}{ (a + x) (b + y)^4},\\
    H_{xxy}(x,y) & = H_{xyx}(x,y) = H_{yxx}(x,y) \\
    & = - \frac{2}{ (a + x)^3 ( b + y)^2},\\
    H_{yyx}(x,y) & = H_{yxy}(x,y) = H_{xyy}(x,y) \\
    & = - \frac{2}{ (a + x)^2 ( b + y)^3}.
\end{align*}

Applying Taylor's theorem, we can expand function $H(N_1,N_2)$ around $(nq_1,nq_2)$ as 
    \begin{align*}
        H(N_1, N_2) & = H(nq_1, nq_2) \\
        & + (N_1 - nq_1) H_x(nq_1,nq_2) + 
        (N_2 - nq_2) H_y(nq_1,nq_2)\\
        & + \frac{1}{2}(N_1 - nq_1)^2 H_{xx}(nq_1,nq_2) + 
        \frac{1}{2}(N_2 - nq_2)^2 H_{yy}(nq_1,nq_2)\\
        & + (N_1 - nq_1)(N_2 - nq_2) H_{xy}(nq_1,nq_2) + R_6,
        \end{align*}
        where the remainder $R_6$ is expressed as
\begin{align*}
        R_6 & := R_{61} + R_{62} + R_{63} + R_{64}\\
        & = \frac{1}{6}(N_1 - nq_1)^3 H_{xxx}(\theta N_1 +(1-\theta) nq_1, \theta N_2 +(1-\theta) nq_2) \\
        & + \frac{1}{6}(N_2 - nq_2)^3 H_{yyy}(\theta N_1 +(1-\theta) nq_1, \theta N_2 +(1-\theta) nq_2)\\
        & + \frac{1}{2}(N_1 - nq_1)^2(N_2 - nq_2) H_{xxy}(\theta N_1 +(1-\theta) nq_1, \theta N_2 +(1-\theta) nq_2)\\
        & + \frac{1}{2}(N_1 - nq_1)(N_2 - nq_2)^2 H_{yyx}(\theta N_1 +(1-\theta) nq_1, \theta N_2 +(1-\theta) nq_2)
    \end{align*}
    with an unknown random parameter $\theta = \theta(N_1,N_2,nq_1,nq_2) \in (0,1)$.
    In what follows, we will establish $\theta$-independent bounds for the four remainder terms above.

    To bound the first term $R_{61}$ involving $H_{xxx}$, a useful observation is that $H_{xxx} = - F(x) G(y)$ is separable, where 
    \[
    F(x) = \frac{1}{(a + x)^4}~~~~\hbox{and}~~~~G(y) = \frac{1}{b + y} 
    \]
    are 
    univariate positive convex functions.
    The convexity and positivity of the functions entail that for any $\theta \in (0,1)$,
    \[
    F(\theta x_1 + (1 -\theta) x_2) G(\theta y_1 + (1 -\theta y_2)) \leq \left(F(x_1) + F(x_2)\right)\left(G(y_1) + G(y_2)\right),
    \]
    in which the RHS is independent of $\theta$. 
    As a result, it holds that 
    \begin{align*}
        \E|R_{61}| & \lesssim \E\left|(N_1 - nq_1)^3\left(\frac{1}{(a + N_1)^4} + \frac{1}{(a + nq_1)^4}\right)\left(\frac{1}{b + N_2} + \frac{1}{b + nq_2}\right)\right| \\
        & \lesssim \sqrt{\E(N_1 - nq_1)^6} \sqrt{\E\left\{\left(\frac{1}{(a + N_1)^4} + \frac{1}{(a + nq_1)^4}\right)^2\left(\frac{1}{b + N_2} + \frac{1}{b + nq_2}\right)^2\right\}}\\
        & \lesssim \sqrt{\E(N_1 - nq_1)^6} \sqrt{\E\left\{\left(\frac{1}{(a + N_1)^8} + \frac{1}{(a + nq_1)^8}\right)\left(\frac{1}{(b + N_2)^2} + \frac{1}{(b + nq_2)^2}\right)\right\}}.
    \end{align*}
    
On the one hand, we have that 
$$\E(N_1 - nq_1)^6 \lesssim (nq_1 + 1)^3.$$ 
On the other hand, it follows from Lemma \ref{lem:upperbound} that 
\begin{align*}
\E\left[\frac{1}{(a + N_1)^8 (b + N_2)^2}\right] &\leq \sqrt{\E\left(\frac{1}{(a + N_1)^{16}}\right)\E\left(\frac{1}{(b + N_2)^4}\right)} \\
& \lesssim \frac{1}{(a + nq_1)^8(b+nq_2)^2} 
\end{align*}
and thus 
\begin{align*}
    &\E\left\{\left(\frac{1}{(a + N_1)^8} + \frac{1}{(a + nq_1)^8}\right)\left(\frac{1}{(b + N_2)^2} + \frac{1}{(b + nq_2)^2}\right)\right\}\\
    \lesssim & \frac{1}{(a + nq_1)^8 (b + nq_2)^2}.
\end{align*}

Combining the results above leads to 
\begin{align*}
    \E|R_{61}| & \lesssim \frac{(nq_1 + 1)^{3/2}}{(a + nq_1)^4 (b + nq_2)} 
     \lesssim \frac{1}{(a + nq_1)^{5/2} (b + nq_2)}.
\end{align*}
Similarly, for term $R_{62}$ we can obtain that 
\begin{align*}
    \E|R_{62}| & \leq C' \frac{(nq_2 + 1)^{3/2}}{(a + nq_1) (b + nq_2)^4} 
    \leq C' \frac{1}{(a + nq_1)(b + nq_2)^{5/2}}.
\end{align*}

Next, we proceed with bounding term $R_{63}$. Similar to the arguments for term $R_{61}$, notice that $ - H_{xxy}(x,y) = F'(x) G'(y)$ can be separated into the product of two univariate positive convex functions, where
\[
F'(x) = \frac{1}{(a + x)^3} \ \text{ and } \ G'(y) = \frac{1}{(b + y)^2}.
\]
Then for any $\theta\in(0,1)$, we can establish a $\theta$-independent bound for
\[
    F'(\theta x_1 + (1 -\theta) x_2) G'(\theta y_1 + (1 -\theta y_2)) \leq \left(F'(x_1) + F'(x_2)\right)\left(G'(y_1) + G'(y_2)\right).
\]
As a result, applying H\"{o}lder's inequality repeatedly, we can deduce that 
    \begin{align*}
        \E|R_{63}| & \lesssim \E\left|(N_1 - nq_1)^2(N_2 -nq_2)\left(\frac{1}{(a + N_1)^3} + \frac{1}{(a + nq_1)^3}\right)\left(\frac{1}{(b + N_2)^2} + \frac{1}{(b + nq_2)^2}\right)\right| \\
        & \lesssim \sqrt{\E(N_1 - nq_1)^4(N_2 - nq_2)^2} \\
        & \times \sqrt{\E\left\{\left(\frac{1}{(a + N_1)^3} + \frac{1}{(a + nq_1)^3}\right)^2\left(\frac{1}{(b + N_2)^2} + \frac{1}{(b + nq_2)^2}\right)^2\right\}}\\
        & \lesssim \sqrt{\left(\E(N_1 - nq_1)^6\right)^{2/3}\left(\E(N_2 - nq_2)^6\right)^{1/3}} \\
        & \times \sqrt{\E\left\{\left(\frac{1}{(a + N_1)^3} + \frac{1}{(a + nq_1)^3}\right)^2\left(\frac{1}{(b + N_2)^2} + \frac{1}{(b + nq_2)^2}\right)^2\right\}}\\
        & \lesssim \frac{(nq_1 + 1)(nq_2 + 1)^{1/2} }{(a + nq_1)^3 (b + nq_2)^{2}} \lesssim \frac{1}{(a + nq_1)^2 (b + nq_2)^{3/2}}.
    \end{align*}
    
Similarly, we can also obtain that 
\[
\E|R_{64}| \lesssim \frac{1}{(a + nq_1)^{3/2}(b + nq_2)^2}.
\]
Consequently, we see that the remainder $R_6$ satisfies 
\begin{align*}
    \E|R_6| & \lesssim  \frac{1}{(a + nq_1)^{5/2} (b + nq_2)} + \frac{1}{(a + nq_1)(b + nq_2)^{5/2}}
    \\
    & + \frac{1}{(a + nq_1)^2 (b + nq_2)^{3/2}} + \frac{1}{(a + nq_1)^{3/2}(b + nq_2)^2}.
\end{align*}
Therefore, we can conclude by integrating out $N$ in the leading terms of the Taylor expansion, which completes the proof of Lemma \ref{lem:doubledisjoint}.

\subsection{Proof of Lemma \ref{lem:6}}
Denote by $P_1 = P\backslash P_0$ and $P_2 = P'\backslash P_0$. The sample counts for subsets $P_0$, $P_1$, and $P_2$ are denoted as $N_0$, $N_1$, and $N_2$, respectively. It is clear that $N_i \sim \mathcal{B}(n,p_i)$, where  the success probability
    $p_i = \P(\vv{X} \in P_i)$ for $i=0,1,2$. In addition, it holds that 
    $(N_0,N_1,N_2) \sim \mathcal{M}(n,(p_0,p_1,p_2))$. 

    A simple observation is that 
    $N = N_0 + N_1$ and $N' = N_0 + N_2$. Let us first consider the conditional expectation of the target inverse moment given $N_0 = n_0$. Recall that $(N_1,N_2)| N_0 = n_0 \sim \mathcal{M}(n',(p'_1,p'_2))$, in which
    \begin{equation}
        n' = n - n_0,~~~~p'_1 = \frac{p_1}{1- p_0},~~~~p'_2 = \frac{p_2}{1 - p_0}.
    \end{equation}
    With an application of Lemma \ref{lem:doubledisjoint}, we can deduce that 
    \begin{equation}\label{eq:total4}
        \begin{split}
        &\E\left[\frac{1}{(a + n_0 + N_1)(a + n_0 + N_2)}| N_0 = n_0\right] \\
        = & \frac{1}{(a + n_0 + n'p'_1) (a + n_0 + n'p'_2)} \\
         + &  \frac{n'p'_1(1-p'_1)}{(a + n_0 +  n'p'_1)^3( a + n_0 +n'p'_2)}\\
     + & \frac{n'p'_2(1-p'_2)}{(a + n_0 + n'p'_1)(a + n_0 + n'p'_2)^3}\\
     - & \frac{n'p'_1p'_2}{(a + n_0 + n'p'_1)^2( a + n_0 + n'p_2)^2} + R_4(n_0)\\
     =: & L_{71}(n_0) + L_{72}(n_0) + L_{73}(n_0) + L_{74}(n_0) + R_7(n_0),
     \end{split}
\end{equation}
in which the remainder $R_7$ satisfies 
\begin{align*}
    R_7(n_0) & \lesssim  \frac{1}{(a + n_0 + n'p'_1)^{5/2} (a + n_0 + n'p'_2)} + \frac{1}{(a + n_0+ n'p'_1)(a + n_0 + n'p'_2)^{5/2}}
    \\
    & + \frac{1}{(a + n_0 + n'p'_1)^2 (a + n_0 + n'p'_2)^{3/2}} + \frac{1}{(a + n_0 + n'p'_1)^{3/2}(a + n_0 + n'p'_2)^2}.
\end{align*}

To further integrate with respect to $N_0$, we need to separate $n_0$ from $n'$ in the present expressions. Notice that
$n_0 + n'p'_1 = np'_1 + n_0(1 - p'_1)$, $n_0 + n'p'_2 = np'_2 + n_0(1 - p'_2)$. In addition, we have that 
\begin{align*}
    a + np'_1 + np_0(1 - p'_1) & = a + \frac{n}{1 -p_0} (p_1 + p_0(1 - p_0 -p_1)) \\
    & = a + n (p_1 + p_0) = a + np
\end{align*}
and $a + np'_2 + np_0(1 -p'_2) = a + np'$. Since $N_0 \sim \mathcal{B}(n,p_0)$, an application of Lemma \ref{lem:singleproduct} with $r = s = 1$ yields that 
\begin{equation}\label{eq:total41}
    \begin{split}
        \E[L_{71}(N_0)] & = \frac{1}{(a + np)(a + np')}
     + \frac{np_0 (1 - p_0)(1 - p'_1)^2}{(a + np)^3(a + np')}\\
    & + \frac{np_0 (1 - p_0)(1 -p'_2)^2}{(a + np))(a + np')^3}
     + \frac{np_0 (1 - p_0)(1 - p'_1)(1 -p'_2)}{(a + np)^2(a + np')^2} + W_{71},
    \end{split}
\end{equation}
where the remainder $W_{41}$ satisfies
\begin{align*}
    W_{71} &\lesssim \frac{1}{(a + np)^{5/2} (a + np')} + \frac{1}{(a + np) (a + np')^{5/2}}\\
    & + \frac{1}{(a + np)^{2} (a + np')^{3/2}} +\frac{1}{(a + np)^{3/2} (a + np')^{2}}.
\end{align*}

Meanwhile, by invoking Lemma \ref{lem:upperbound}, we see that remainder $R_7$ satisfies
\begin{align*}
    \E[R_{7}(N_0)] &\lesssim \frac{1}{(a + np)^{5/2} (a +np')} + \frac{1}{(a + np)(a + np')^{5/2}}\\
    &+\frac{1}{(a + np)^{2} (a + np')^{3/2}} + \frac{1}{(a +np)^{3/2} (a +np')^{2}}.
\end{align*}
Observe that
\begin{align*}
    \frac{n'p'_1}{a + n_0 + n'p'_1} & = \frac{np'_1 - n_0p'_1}{a + np'_1 + n_0(1 -p'_1)} 
     = \frac{np'_1 - n_0p'_1}{a + np'_1 + n_0(1 -p'_1)}\\
    & = \frac{np'_1}{a + np'_1 + n_0(1 - p'_1)} -\frac{p'_1}{1 - p'_1}\left(1 - \frac{a + np'_1}{a + np'_1 + n_0(1 - p'_1)}\right)\\
    &=\frac{1}{1-p'_1}\left(\frac{np'_1 + ap'_1}{a + np'_1 + n_0(1 -p'_1)} - p'_1\right)
\end{align*}
and similarly,
\begin{align*}
    \frac{n'p'_2}{a + n_0 + n'p'_2} & = \frac{1}{1-p'_2}\left(\frac{np'_2 + ap'_2}{a + np'_2 + n_0(1 -p'_2)} - p'_2\right).
\end{align*}
Then it follows that
\begin{align*}
    L_{72}(n_0) & = \frac{n'p'_1(1-p'_1)}{(a + np'_1 + n_0(1 -p'_1))^3( a + np'_2 +n_0(1 -p'_2))}\\
    & = \left(\frac{np'_1 + ap'_1}{a + np'_1 + n_0(1 -p'_1)} - p'_1\right) {(a + np'_1 + n_0(1 -p'_1))^2( a + np'_2 +n_0(1 -p'_2))}\\
    & = - \frac{p'_1}{(a + np'_1 + n_0(1 -p'_1))^2( a + np'_2 +n_0(1 -p'_2))}\\
    & + \frac{np'_1 + ap'_1}{(a + np'_1 + n_0(1 -p'_1))^3( a + np'_2 +n_0(1 -p'_2))}.
\end{align*}

Applying Lemma \ref{lem:singleproduct} with $r = 2 ,s =1$ and $r=3, s=1$, respectively, we can deduce that 
\begin{equation}\label{eq:total42}
    \begin{split}
        \E[L_{72}(N_0)] & = - \frac{p'_1}{(a + np'_1 + n_0(1 - p'_1))^2( a + np'_2 + n_0(1 -p'_2))}\\
    & + \frac{np'_1 + ap'_1}{(a + np'_1 + n_0(1 - p'_1))^3( a + np'_2 + n_0(1 -p'_2))} \\
    & = - \frac{p'_1}{(a + np)^2 (a + np')} + \frac{np'_1 +a p'_1}{(a + np)^3 (a + np')} + W_{72}\\
    & = \frac{np'_1 (1 - p)}{(a + np)^3 (a + np')} + W_{72},
    \end{split}
\end{equation}
where the remainder $W_{72}$ satisfies
\begin{align*}
    W_{72} & \lesssim \frac{1}{(a + np)^3( a + np')} + \frac{1}{(a + np)^2( a + np')^2}.
\end{align*}
Similarly, we can also establish that 
\begin{equation}\label{eq:total43}
    \begin{split}
        \E[L_{73}(N_0)] & = - \frac{p'_2}{(a + np'_1 + n_0(1 - p'_1))( a + np'_2 + n_0(1 -p'_2))^2}\\
    & + \frac{np'_2 + ap'_2}{(a + np'_1 + n_0(1 - p'_1))( a + np'_2 + n_0(1 -p'_2))^3} \\
    & = - \frac{p'_2}{(a + np) (a + np')^2} + \frac{np'_2 +a p'_2}{(a + np)(a + np')^3} + W_{73}\\
    & = \frac{np'_2 (1 - p)}{(a + np) (a + np')^3} + W_{73},
    \end{split}
\end{equation}
where the remainder $W_{73}$ satisfies
\begin{align*}
    W_{73} & \lesssim \frac{1}{(a + np)( a + np')^3} + \frac{1}{(a + np)^2( a + np')^2}.
\end{align*}

Moreover, it holds that 
\begin{align*}
    L_{74}(n_0) & = 
    - \frac{n'p'_1p'_2}{(a + np'_1 +n_0(1 -p'_1))^2 (a + np'_2 +n_0(1 -p'_2))^2} \\
    & = - \frac{p'_2}{1 - p'_1}\left(\frac{np'_1 + ap'_1}{a + np'_1 +n_0(1 -p'_1)} - p'_1\right) \\
    & \times \frac{1}{(a +np'_1 +n_0(1 -p'_1))(a + np'_2 + n_0(1 -p'_2))^2}.
\end{align*}
Hence, we have that 
\begin{equation}\label{eq:total44}
\begin{split}
    \E[L_{74}(N_0)] & = - \frac{p'_2}{1 -p'_1}\left(\frac{np'_1 + ap'_1}{a + np} - p'_1\right) \frac{1}{(a + np)(a +np')^2} + W_{74}\\
    &= - \frac{np_1 p_2}{(1 - p_0)(a + np)^2 (a +np')^2} + W_{74},
    \end{split}
\end{equation}
where the remainder $W_{74}$ satisfies
\begin{align*}
    W_{74} & \lesssim \frac{1}{(a + np)^2 (a + np')^2} + \frac{1}{(a + np) (a + np')^3} . 
\end{align*}

Substituting (\ref{eq:total41})--(\ref{eq:total44}) into (\ref{eq:total4}), we can obtain that 
\begin{align*}
    \E\left[\frac{1}{(a + N) (a + N')}\right] & = \frac{1}{(a + np)(a + np')}
     \\
     & + \frac{np_0 (1 - p_0)(1 - p'_1)^2}{(a + np)^3(a + np')} + \frac{np'_1 (1 - p)}{(a + np)^3 (a + np')} \\
    & + \frac{np_0 (1 - p_0)(1 -p'_2)^2}{(a + np))(a + np')^3} + \frac{np'_2 (1 - p)}{(a + np) (a + np')^2}\\
     &+ \frac{np_0 (1 - p_0)(1 -p'_1)(1 -p'_2)}{(a + np)^2(a + np')^2}
      - \frac{np'_1p'_2(1 - p_0)}{(a + np)^2 (a + np')^2} + \tilde{R}_7,
\end{align*}
where the remainder $\tilde{R}_7$ satisfies
\begin{align*}
    \tilde{R}_7 & \lesssim \frac{1}{(a + np)^{5/2} (a +np')} + \frac{1}{(a + np)(a + np')^{5/2}}\\
    &+\frac{1}{(a + np)^{2} (a + np')^{3/2}} + \frac{1}{(a +np)^{3/2} (a +np')^{2}}.
\end{align*}

Some simple calculations lead to 
\begin{align*}
    &\frac{np_0 (1 - p_0)(1 - p'_1)^2}{(a + np)^3(a + np')} + \frac{np'_1 (1 - p)}{(a + np)^3 (a + np')}\\
    = & \frac{n(1 - p)}{(1 - p_0)(a + np)^3 (a + np')}\left(p_0 (1 - p) + p_1\right) \\
    =& \frac{n p (1 - p)}{(a + np)^3 (a + np')}\\
    = & \frac{1 - p}{(a + np)^2 (a +np')} - \frac{a (1 -p)}{(a + np)^3(a +np')},\\
    &\frac{np_0 (1 - p_0)(1 -p'_2)^2}{(a + np))(a + np')^3} + \frac{np'_2 (1 - p)}{(a + np) (a + np')^2}\\
    =& \frac{np'(1 -p')}{(a +np) (a + np')^3}\\
    =& \frac{1 - p'}{(a +np)(a +np')^2} - \frac{a(1 -p')}{(a +np)(a +np')^3}
\end{align*}
and 
\begin{align*}
    &\frac{np_0 (1 - p_0)(1 -p'_1)(1 -p'_2)}{(a + np)^2(a + np')^2}
      - \frac{np'_1p'_2(1 - p_0)}{(a + np)^2 (a + np')^2}\\
    =  &\frac{n(1 - p_0)}{(a +np)^2 (a +np')^2}(p_0(1 - p'_1)(1 -p'_2) - p'_1p'_2) \\
    = & \frac{n}{(1 - p_0)(a +np)^2 (a +np')^2}(p_0(1 - p)(1 - p') - p_1p_2)\\
    =& \frac{n}{(1 - p_0)(a +np)^2 (a +np')^2}(p_0 - p_0(p+p') + p_0pp' - (p -p_0)(p'-p_0))\\
    =&\frac{n}{(a +np)^2 (a +np')^2}(p_0 - pp').
\end{align*}

Therefore, substituting the simplified expressions above into the asymptotic expansion of the inverse moment, we can conclude that
\begin{align*}
    \E\left[\frac{1}{(a + N) (a + N')}\right] & = \frac{1}{(a + np)(a + np')}
      + \frac{1 -p}{(a + np)^2 (a + np')} \\
     & + \frac{1 -p'}{(a + np) (a + np')^2}  + \frac{n(p_0 - pp')}{(a + np)^2 (a + np')^2 } +\tilde{R}'_7,
\end{align*}
which the remainder $\tilde{R}'_7$ has the same bound as $\tilde{R}_7$. The final conclusion follows from a further step of simplification, which concludes the proof of Lemma \ref{lem:6}.

\section{Additional materials in Section \ref{sec:CART}} \label{new.SecC}

\subsection{Algorithms for CART processes in Section \ref{new.Sec.linear}}
\label{new.SecC.1}

For completeness, we list in Algorithm \ref{alg:binary} the \textit{Binary CART process} and in Algorithm \ref{alg:linear} the \textit{Uniform CART process}.

\begin{algorithm}
\caption{Binary CART process}
\label{alg:binary}
\begin{algorithmic}[1]
\State Initialize $I_0 = (0, \ldots, 0)$ of length $d$.
\For{$k = 1$ to $l$}
    \State Randomly select a subset $\Xi_k$ of size $\lceil \gamma d \rceil$ from $\{1, \ldots, d\}$.
    \State Filter $\Xi_{k}$ to obtain unsplit features $\Xi_{0k} =\{j: I_{k-1,j} = 0\}$. 
    \If{$\Xi_{0k} \neq \varnothing$}
        \State Choose randomly an index $j^*$ such that $j^* = \arg\max_{j \in \Xi_{0k}} \beta_j^2$. 
    \EndIf
    \State Update $I_{k}$ by setting $I_{k,j^*} = 1$ for $j^*$ and $I_{k,j} = I_{k-1,j}$ for the rest, if $\Xi_{0k}\neq\varnothing$; otherwise, set $I_k = I_{k-1}$.
\EndFor
\State \Return $I_l$
\end{algorithmic}
\end{algorithm}

\begin{algorithm}
\caption{Uniform CART process}
\label{alg:linear}
\begin{algorithmic}[1]
\State Initialize $J_0 = (0, \ldots, 0)$ of length $d$.
\For{$k = 1$ to $l$}
    \State Randomly select a subset $\Xi_k$ of size $\lceil \gamma d \rceil$ from $\{1, \ldots, d\}$.
    \State  Randomly select $j^*$ from $\Xi_{1k} = \arg \max_{j \in \Xi_{0k}} \beta_j^2 2^{-2 J_{k-1,j}}$.
    
    \State Update $J_{k}$ by setting $J_{k,j^*} = J_{k-1,j^*} + 1$ and $J_{k,j} = J_{k-1,j}$ for the remaining $j$'s.
\EndFor
\State \Return $J_l$.
\end{algorithmic}
\end{algorithm}

\subsection{Additional figures in Section \ref{new.Sec.gamma1}}
\label{new.SecC.2}

\begin{figure}[htp]
    \centering
    \includegraphics[width = 0.8\textwidth]{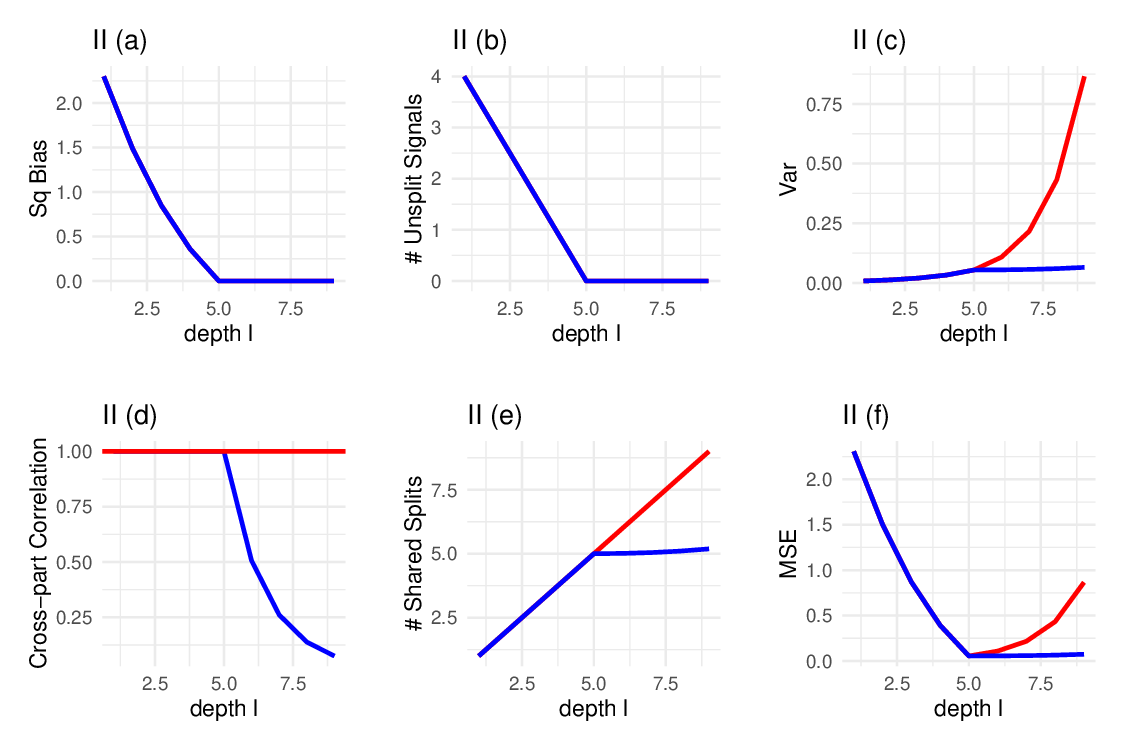}
    \caption{Performance measures (a)--(f) for tree and forest in the binary case under configuration (II) as tree depth $l$ varies when $\gamma = 1$. Red: tree; blue: forest.}
    \label{fig:simu7}
\end{figure}

\begin{figure}[htp]
    \centering
    \includegraphics[width = 0.8\textwidth]{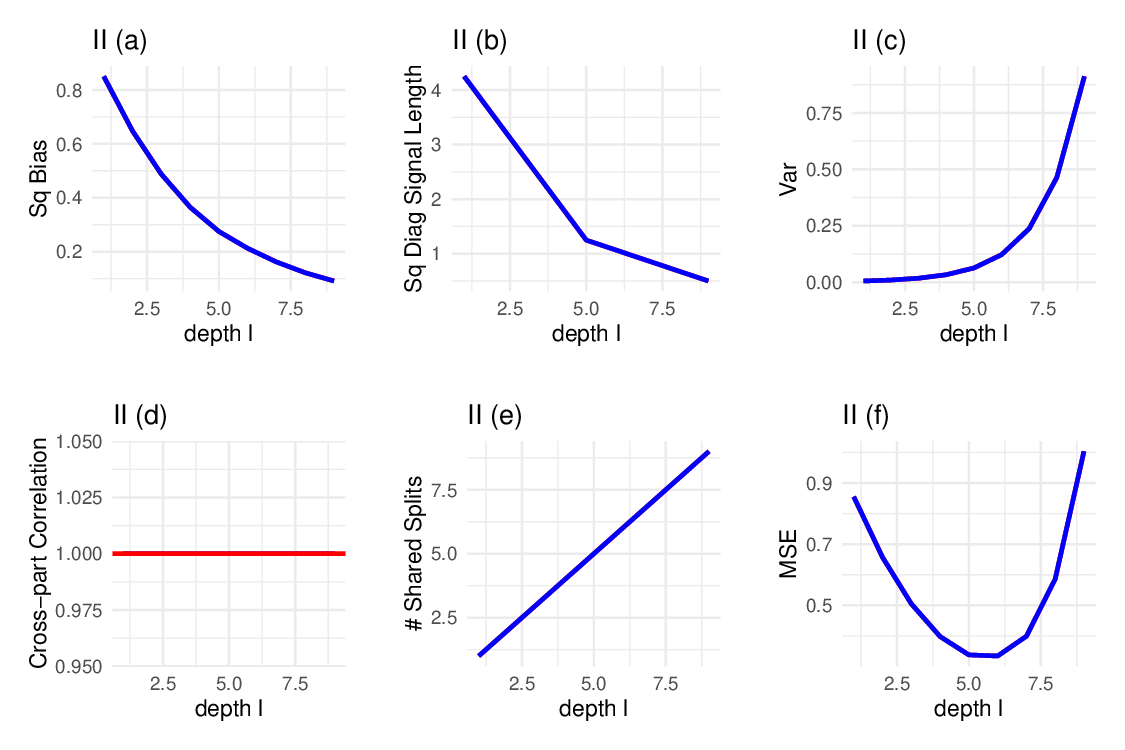}
    \caption{Performance measures (a)--(f) for tree and forest in the continuous case under configuration (II) as tree depth $l$ varies when $\gamma = 1$. Red: tree; blue: forest.}
    \label{fig:simu8}
\end{figure}

Figures \ref{fig:simu7} and \ref{fig:simu8} are additional simulation results from Section \ref{new.Sec.gamma1}.

\subsection{Additional figures and discussions in Section \ref{new.Sec.subsample}}
\label{new.SecC.3}

\begin{figure}[htp]
    \centering
    \includegraphics[width = 0.8\textwidth]{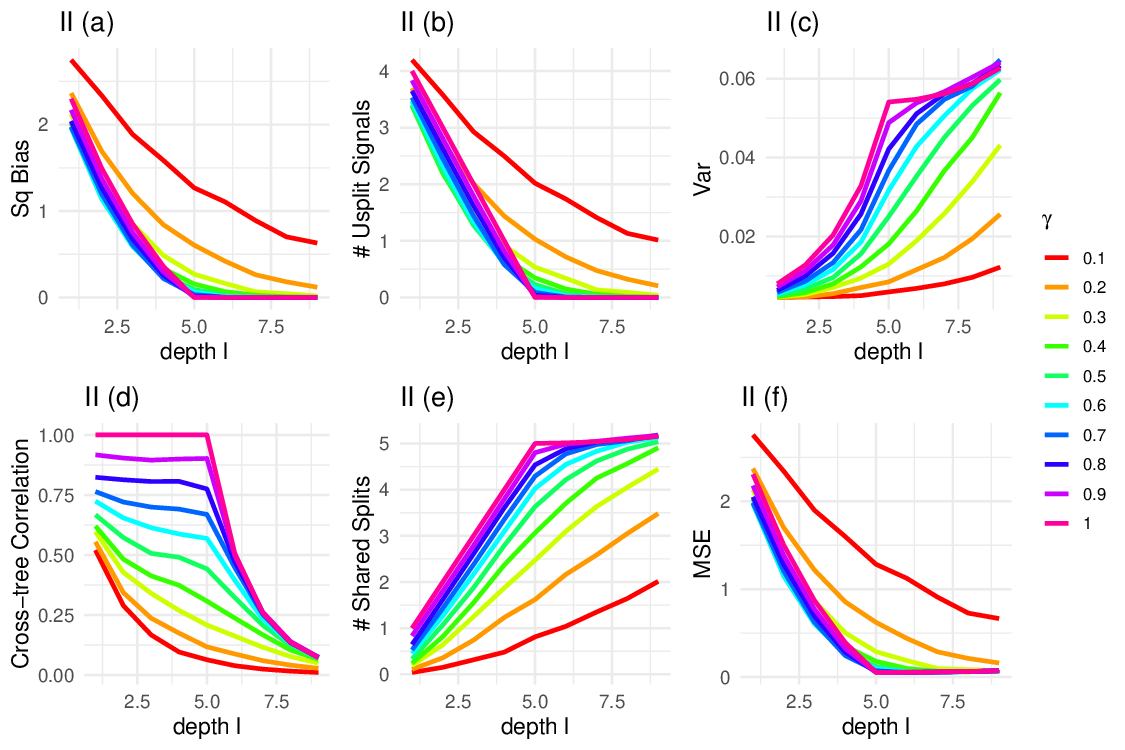}
    \caption{Performance measures (a)--(f) for forests in the binary case when  $\gamma$ and $l$ vary under configuration (II).}
    \label{fig:simu3}
\end{figure}

\begin{figure}[htp]
    \centering
    \includegraphics[width = 0.8\textwidth]{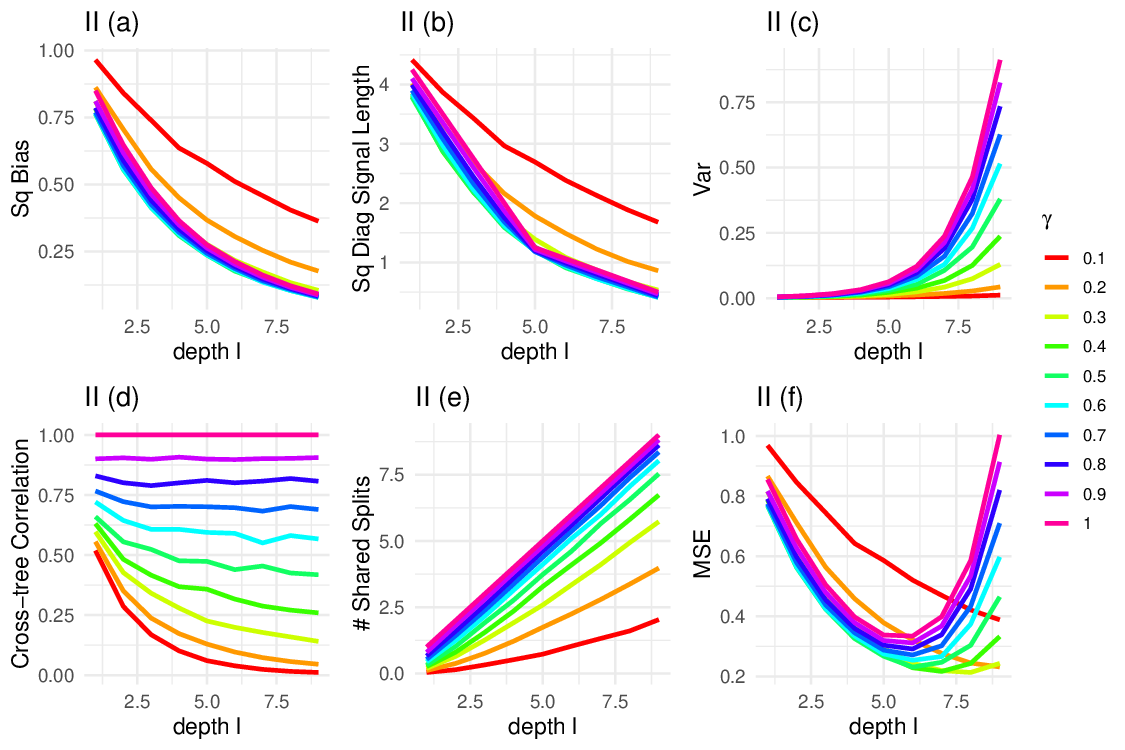}
    \caption{Performance measures (a)--(f) for forests in the continuous case when  $\gamma$ and $l$ vary under configuration (II).}
    \label{fig:simu6}
\end{figure}

Figures \ref{fig:simu3} and \ref{fig:simu6} are additional simulation results from Section \ref{new.Sec.subsample}. To demonstrate feature subsampling does not always increase the squared bias while reducing the variance at the early stage of tree-building, let us consider Unequal Configuration (II).  The tree-building process is purely deterministic when $\gamma = 1$ and $ l\leq s$: features are split deterministically in the decreasing order of impurity decrement (see Figures \ref{fig:simu7} and \ref{fig:simu8} in Section \ref{new.SecC.2}). 
With $\gamma<1$, however, since some important features may not be selected into a subsample set, such deterministic splitting order is disrupted. This may seem harmful to bias reduction because less important variables are split earlier on. However, feature subsampling creates a large number of trees, and ensemble allows (on average) many more informative features to be split earlier on.  This results in a reduction of squared bias by encouraging fewer shared-unsplit-informative variables by two independent CART processes; see (\ref{eq:binaryRF}), (\ref{eq:unifRF}), and Figures \ref{fig:simu3}(a)--(b).  
Meanwhile, Type I exogenous randomness also introduces variability in splits along informative features, allows non-informative features to be involved earlier, and thus helps reduce the cross-tree correlation in Figure \ref{fig:simu3}(d), leading to lower variance in Figure \ref{fig:simu3}(c) and a reduction in MSE in Figure \ref{fig:simu3}(e). 
These observed behaviors in Figure \ref{fig:simu6} resemble those in the binary setting, with similar interpretations.
To summarize, in this example, we see that feature subsampling can introduce beneficial randomness to achieve \textit{both} bias and variance reduction.

\end{document}